\documentclass[acmsmall]{acmart}
\setcopyright{none}
\renewcommand\footnotetextcopyrightpermission[1]{}

\AtBeginDocument{%
  \providecommand\BibTeX{{%
    \normalfont B\kern-0.5em{\scshape i\kern-0.25em b}\kern-0.8em\TeX}}}

\usepackage{algorithm2e}
\usepackage{algpseudocode}
\usepackage{multirow}
\usepackage{makecell}
\usepackage{color}
\usepackage{multicol}
\usepackage{hhline}

\newcommand{\eg}{\emph{e.g.,}\xspace}

\title{Large-Scale Deep Learning Optimizations: A Comprehensive Survey}

\begin{document}

\author{Xiaoxin He}
\email{he.xiaoxin@u.nus.edu}
\affiliation{%
  \institution{National University of Singapore}
  \country{he.xiaoxin@u.nus.edu}
}

\author{Fuzhao Xue}
\email{f.xue@u.nus.edu}
% % \orcid{1234-5678-9012}
\affiliation{%
  \institution{National University of Singapore}
  \country{f.xue@u.nus.edu}
}

\author{Xiaozhe Ren}
\email{renxiaozhe@huawei.com}
\affiliation{%
    \institution{Huawei Noah's Ark Lab}
%   \city{Hong Kong}
    \country{renxiaozhe@huawei.com}
}

\author{Yang You}
\email{youy@comp.nus.edu.sg}
\affiliation{%
    \institution{National University of Singapore}
    \country{youy@comp.nus.edu.sg}
}

\makeatletter
\let\@authorsaddresses\@empty
\makeatother


\begin{abstract}
  Deep learning have achieved promising results on a wide spectrum of AI applications. Larger datasets and models consistently yield better performance. However, we generally spend longer training time on more computation and communication. In this survey, we aim to provide a clear sketch about the optimizations for large-scale deep learning with regard to the model accuracy and model efficiency. We investigate algorithms that are most commonly used for optimizing, elaborate the debatable topic of generalization gap arises in large-batch training, and review the SOTA strategies in addressing the communication overhead and reducing the memory footprints. 
\end{abstract}




\maketitle
\pagestyle{plain} % removes running headers
\thispagestyle{empty}

\section{Introduction}

Nowadays, deep learning (DL) have achieved promising results on a wide spectrum of AI application domains ranging from computer vision (\eg image classification~\cite{image_classification, image_classification_densely_connected_convolutional_networks,lou2021sparse}, object detection and segmentation~\cite{Fast_R_CNN,Faster_R_CNN_Towards_Real_Time_Object_Detection_with_Region_Proposal_Networks,Mask_R_CNN,Fully_convolutional_networks_for_semantic_segmentation}), natural language processing (\eg language modeling~\cite{bert,xue2021go} and machine translation~\cite{Attention_is_All_you_Need, machine_translation}), information retrieval (\eg recommendation system~\cite{Neural_Collaborative_Filtering}) and many others.   The scale is the main driver behind the rise of DL~\cite{KrizhevskySH12, image_classification, SimonyanZ14a, KrizhevskySH17,SzegedyLJSRAEVR15, bert}. Larger datasets and neural networks consistently yield better performance across all tasks that generally require more computation and longer training time. Therefore,
recent years have witnessed a surge of interests from both academia and industry in scaling up DL with distributed training on a large cluster of devices such as TPUs and GPUs with higher computation capability and memory limit.  Data parallelism has become a dominant practice for distributed training.  It distributes a large batch to multiple devices, where each device holds an identical model replica, computes the gradient of a local batch and finally gathers the gradients at each iteration for synchronous parameter update. With recent optimization techniques, it is now able to train very large batches on thousands of GPU devices. However, training at such scales requires overcoming both algorithmic and systems-related challenges. One of the main challenges is the degradation of model accuracy with large batch size beyond a certain point (e.g., 32k). Naively increasing the batch size typically results in  degradation of generalization performance and reduces computational benefits. Additionally, we can not always improve the training speed by just using more processors as the communication cost is a non-negligible  overhead. Intuitively multiple processors collaboratively training one task can reduce the overall training time, but the corresponding communication cost between processors is heavy and limits the model scalibility. Worse still, models with tens of billions to trillions of parameters clearly do not fit into memory of a single device, and simply adding more devices will not help scale the training. This limitation prevents DL researchers from exploring more advanced model architectures.
Existing works investigate and develop optimization techniques to overcome these problems so as to accelerate training large-scale deep neural networks (DNNs). We categorise these works into two categories, one endeavors to maintain/improve the model accuracy in the large-scale setting and the other emphasises on the model efficiency, designing algorithms that are less hungry for communication and memory. Importantly, they are not mutually exclusive but can be used collaboratively to further speed up the training.

\subsection{Related Surveys}
\citet{A_Survey_on_Deep_Learning_Algorithms_Techniques_and_Applications} give an overview of DL from different perspectives, including history, challenges, opportunities, algorithms, frameworks, applications, and parallel and distributed computing techniques. ~\citet{wang2020survey} provide a quick survey on large-scale distributed deep learning systems, which concisely
introduces parallelisms, parameter server architectures, synchronization
schemes, related applications, and platforms. While some other surveys focus on a certain scope in deep learning: communication-efficiency in large-scale parallelism systems ~\cite{survey_Approximate_Communication_Techniques_for_Reducing_Communication_Bottlenecks_in_Large_Scale_Parallel_Systems,survey_Communication_Efficient_Distributed_DL},  parallelization strategies~\cite{Demystifying_Parallel_and_Distributed_Deep_Learning_An_In_depth_Concurrency_Analysis} and numerical optimization algorithms~\cite{survey_Optimization_for_deep_learning_theory_and_algorithms,Optimization_Methods_for_Large_Scale_Machine_Learning,6796060,overview_of_gradient_descent_optimization_algorithms}.
Large-scale DL represents a distinctive setting in which the accuracy, computation, communication and memory are closely connected and mutually restricted. However, existing surveys either merely concern part of them or do not address optimizations in the context of large-scale DL. Different from the two surveys~\cite{wang2020survey, A_Survey_on_Deep_Learning_Algorithms_Techniques_and_Applications} that are mostly related to ours, we focus more on the design of the algorithm rather than the system architecture.  The novelty of this paper is its emphasises on both model accuracy and model efficiency, which captures critical aspects of  large-scale deep learning training by presenting a review of the state-of-the-art (SOTA) optimization techniques and illustrating the trade-off in between.

%  but with the identical pursuit of shorter training time. 
%  \fuzhao{The point we argued is not strong enough. It seems \cite{survey_Communication_Efficient_Distributed_DL} also discusses the large-scale DL from a system perspective, so what is the difference between ours and theirs, and why ours is the better one and why our work is meaningful? } 
%  Large-scale DL represents a distinctive setting in which the design of optimizers, communication and memory are closely connected and mutually restricted. And thus we are intended to provide a comprehensive understanding about optimizations at a system level.

\subsection{Structure of the Survey}
\begin{figure}[t]
    \centering
    \includegraphics[scale=0.5]{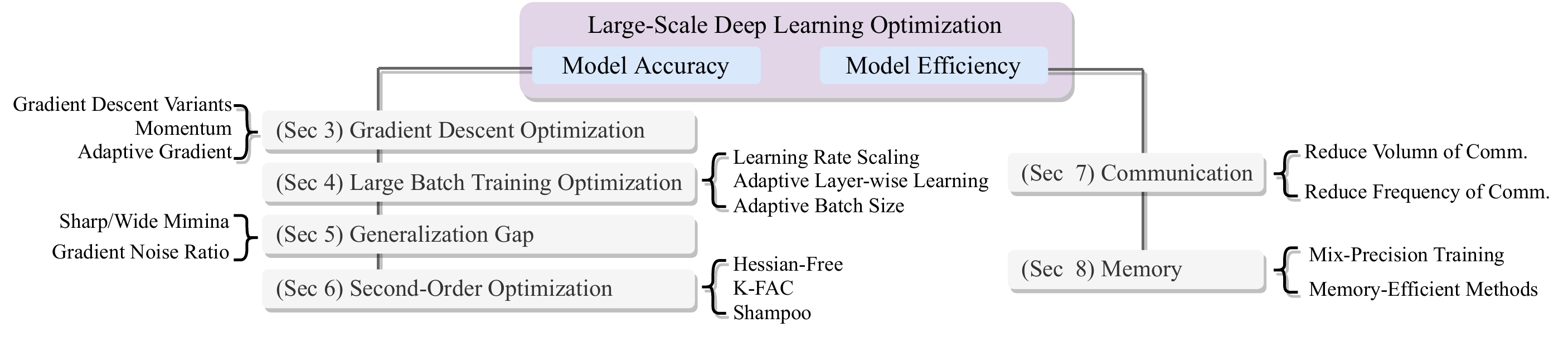}
    \caption{The Overall Structure of this Survey}
    \label{fig: overview}
\end{figure}
The overall structure of this survey is presented in Figure~\ref{fig: overview}.
Section ~\ref{sec: Preliminaries} presents the formulation of a typical neural network optimization problem for supervised learning. We roughly divide the large-scale DL optimization  into two components: model accuracy and model efficiency. 
% \fuzhao{we should remove supervised learning here, SGD and Adam can also be used at self-supervised learning and even unsupervised learning.}. 
Section ~\ref{sec: Gradient Descent Optimization Algorithms} introduces the gradient descent optimization family, including gradient descent variants, momentum SGD and adaptive gradient algorithms. As large batch training with data parallelism has increasing popularity in DL and meanwhile introduces challenges, Section ~\ref{sec: Large Batch Training} discusses problems in this setting, 
and reviews main SOTA optimization strategies to improve the situation. Section ~\ref{section: Generalization Gap} dives into the generalization gap --- a debating topic in large batch training. Section ~\ref{sec: Second Order Optimization} introduces second-order optimizations. Then we turn our attention to model efficiency. Section ~\ref{sec: Communication} investigates the communication bottleneck and Section ~\ref{sec: Memory} focuses on memory-efficient techniques. Finally, Section ~\ref{sec: Conclusion} concludes this article.

\section{Preliminaries}\label{sec: Preliminaries}
Following many machine learning applications~\cite{Optimization_Methods_for_Large_Scale_Machine_Learning, Second_order_Optimization_for_Non_convex_Machine_Learning_an_Empirical_Study, The_Tradeoffs_of_Large_Scale_Learning}, we consider a space of input-output pairs $(x, y) \in X \times Y$ has a probability distribution $P(x, y)$. The conditional distribution $P(y|x)$ represents the true relationship between inputs and outputs. The discrepancy between the predicted output $\hat y$ and the real output $y$ is measured by a smooth but possibly non-convex loss function $\mathscr{l}(\hat y,y)$. The objective is to minimize the \textit{expected risk}
\begin{equation}
    E(f) = \int \mathscr{l}(f(x),y)P(x,y) = \mathbb{E}[\mathscr{l}(f(x),y)],
\end{equation}
that is,
\begin{equation}
    f^*(x)=\mathop{argmin}_{\hat{y}}\mathbb{E}[\mathscr{l}(\hat{y},y)|x].
\end{equation}
Since $\mathbb{P}$ is an unknown distribution, in practice, one seeks the solution of a problem that involves an estimate of the \textit{empirical risk}~\cite{Understanding_Machine_Learning_From_Theory_to_Algorithms}
\begin{equation}\label{Eq: empirical risk}
    E_n(f)=\frac{1}{n}\sum_{i=1}^n\mathscr{l}(f(x_i),y_i)=\mathbb{E}_n[\mathscr{l}(f(x),y].
\end{equation}
The goal of solving Eq.[\ref{Eq: empirical risk}] is to obtain a solution with small generalization error, i.e., high predictive accuracy on unseen data. 

\section{Gradient Descent Optimization Algorithms}\label{sec: Gradient Descent Optimization Algorithms}
Training a DNN is an optimization process, i.e., finding the parameters in the network that minimize the loss function. Gradient descent and its variant algorithms are the most popular algorithms to optimize neural networks (NNs)~\cite{overview_of_gradient_descent_optimization_algorithms}. In order to control the oscillation of gradient descent methods, the idea of using momentum is introduced. Moreover, adapting the learning rate w.r.t. the gradient of the previous
stages is found beneficial to avoid the fluctuation. In this section, we briefly sort out the mainstream optimization algorithms, consisting of gradient descent variants (Section~\ref{subsec: Gradient Descent Variants}),  momentum (Section~\ref{subsec: Momentum}) and adaptive gradient algorithms (Section~\ref{subsec: Adaptive Gradient Algorithms}). 
% We also summarise additional optimization techniques used in practice in Section~\ref{subsec: Additional Optimization Techniques}.

\subsection{Gradient Descent Variants}\label{subsec: Gradient Descent Variants}
Gradient descent and its variants dominate the optimization algorithms of DL. The gradient descent (GD) methods aim to minimize the empirical risk of a model by repeatedly computing the gradient of a loss function on a single training sample, or a (full) batch of samples, and continuously updating the model parameters accordingly by following the gradient of the objective function in the opposite direction.
There are three variants in gradient descent which differ in the number of samples used for each step (updating model parameters), resulting in different accuracy and learning time. 

\subsubsection{Batch Gradient Descent}
Batch gradient descent, a.k.a. vanilla gradient descent, minimizes the loss function $L(x)$ with the following form:
\begin{equation}
L(x)=\frac{1}{|S|}\sum_{s\in S}l(x,s).    
\end{equation}
Here  $x$ is the weights of a network, $S$ is a labeled training set, $|S|$ is the number of samples in the training set, and $l(x,s)$ is the loss computed from sample $s\in S$ and and its label $y$. Typically $l$ is the sum of a classification loss (e.g., cross-entropy) and a regularization loss on $x$. And then update the weights
\begin{equation}
    x= x-\eta \nabla L(x),
\end{equation}
where $\eta$ is a learning rate (LR) which controls how large of a step to take in the opposite direction of the gradient. As we need to go through the whole training set to calculate the gradient for one update of weights, batch gradient descent can be very slow, especially for large datasets (which is very common in DL tasks). Batch gradient descent also does not allow updating model online, i.e., with new examples on-the-fly.

\subsubsection{Stochastic Gradient Descent}\label{subsec: SGD}
Unlike batch gradient descent (GD), which calculates the gradients using the all training samples, Stochastic Gradient Descent (SGD) performs one weights update for each training sample
\begin{equation}
    x = x -\eta \nabla l(x,s).
\end{equation}
Therefore, SGD addresses the computational bottleneck of batch gradient descent. It is significantly faster than batch gradient descent and can be used online. The drawback is that the gradient computed from just one sample is not representative enough for the whole training set. Consequently, the variance of gradients leads to a fierce fluctuation in the loss function.

\subsubsection{Mini-batch Stochastic Gradient Descent}\label{subsec: mini-batch SGD}
Mini-batch SGD takes both advantages of batch GD and SGD by performing weights update for each mini-batch $B$
\begin{equation}
    x = x - \frac{\eta}{|B|}\nabla\sum_{s\in B}l(x,s).
\end{equation}
In batch GD, the gradients are computed over the entire dataset, providing an accurate estimation of the true gradient. It takes lots of time and memory to do that. But the real handicap is the batch gradient trajectory lands in a bad spot. While in SGD, parameters are updated by adding the gradient computed on a single sample of the dataset, which is very noisy and may go off in a direction far from the batch gradient. However, the noisiness is exactly what we want in non-convex optimization, because it helps to escape from saddle points or local minima~\cite{GeHJY15}. The disadvantage is its terribly inefficiency of looping over the entire dataset many times to find a good solution. The mini-batch methodology is a compromise that injects enough noise to each gradient update, while achieving a relative speedy convergence. Mini-batch SGD is found to be very effective in the case of large-scale learning~\cite{The_Tradeoffs_of_Large_Scale_Learning}. 

% Compared with full batch GD, the advantages of mini-batch SGD are: 1) The gradient is quickly computed. There is no need to go through the whole training set to update the weights. 2) Datasets can be redundant. Many data are similar, a subset of the data will give almost the same gradient than using all data samples. When compared to SGD, the benefits are also two-fold: 1) It helps to reduce the variance of gradients which leads to a stable convergence. 2) It enables the distributed synchronized SGD by assigning mini-batch to different machines.  Mini-batch gradient descent seeks to find a balance between the robustness of SGD and the efficiency of batch GD. It is the most common implementation of gradient descent used in the field of DL.

\subsection{Momentum} \label{subsec: Momentum}
SGD has trouble navigating a long and narrow valley in the loss function surface, i.e., the direction of the gradient is almost perpendicular to the long axis of the valley. In such a situation, the system oscillates back and forth in the direction of the short axis, and only moves very slowly along the long axis of the valley. 

\subsubsection{Momentum SGD}
Momentum~\cite{journals/nn/Qian99} is a strategy that helps to counteract the oscillation along the short axis meanwhile accumulate contributions along the long axis. In other words, the momentum strengthens for dimensions whose gradients point in the same directions and dampens updates for dimensions whose gradients change directions.  This allows Momentum to minimize the training loss in fewer steps than full batch gradient descent~\cite{park2019specaugment}. Specifically, momentum SGD adds update in previous step to the current update, and determines the next update $v_t$ as a linear combination of the gradient and the previous update $v_{t-1}$:
\begin{equation}\label{eq: momentum sgd}
\begin{split}
v_t &= \beta v_{t-1} + \eta \nabla L(x)\\
x &= x -  v_t.   
\end{split}
\end{equation}

\subsubsection{Nesterov Accelerated Gradient}
In Eq.[\ref{eq: momentum sgd}], we know that we are going to move by at least $\beta v_{t-1}$ and a bit more by $\eta\nabla L(x)$. And in Nesterov Accelerated Gradient (NAG)~\cite{nesterov1983method}, it looks ahead by calculating the gradient at the partially updated value of $(x-\beta v_{t-1})$ instead of using the current value:
\begin{equation}
\begin{split}
v_t &= \beta v_{t-1} + \eta \nabla L(x -  \beta  v_{t-1})\\
x &= x-  v_t    .
\end{split}
\end{equation}
Such "look before you leap" prevents us from going too fast and results in increased
responsiveness. While the optimization path taken by classical momentum SGD exhibits large oscillations along the high-curvature vertical direction, NAG is able to avoid these oscillations almost entirely~\cite{On_the_importance_of_initialization_and_momentum_in_DL}.

\subsection{Adaptive Gradient Algorithms}\label{subsec: Adaptive Gradient Algorithms}
The methods mentioned above apply a same LR to all dimensions of the parameters. Since each dimension of parameters relates to the loss function in different ways, a per-dimension LR is more advantageous due to the more accurate and precise control on the step size. 
% Therefore, many efforts have been made to accelerate optimization by applying adaptive LR.
% For instance, Adagrad~\cite{AdaGrad} and its variants e.g., RMSProp, Adadelta~\cite{Adadelta}, and Adam~\cite{kingma2017adam} have been used in many applications.
Therefore, a variety of adaptive gradient-based methods have been proposed where gradients are divided by the component-wise accumulation of previous gradients. For example, AdaGrad~\cite{AdaGrad} uses the sum of the squares of all past gradients, whereas Adadelta~\cite{Adadelta}, RMSProp~\cite{RMSProp} and Adam~\cite{kingma2017adam} use an exponentially decaying average.

\subsubsection{AdaGrad}\label{subsection: AdaGrad}
% AdaGrad~\cite{AdaGrad} is an informative gradient-based gradients method which dynamically incorporates knowledge of the geometry of the data observed in earlier iterations. 
Previously, we perform updates for all parameters using a same LR, regardless of their frequency and magnitude. This may lead to a failure in capturing the knowledge of infrequently occurring updates (which are highly informative and discriminative). AdaGrad~\cite{AdaGrad} alleviates this problem by performing larger updates for infrequent parameters and smaller updates for frequent parameters, which enables it to do well with sparse gradients
\begin{equation}\label{eq: adagrad}
        x_{t,i} = x_{t-1,i}-\frac{\eta}{\sqrt{G_{t,ii}+\epsilon }}\nabla L(x_{t,i}).
% x_t = x_{t-1}-\frac{\eta}{\sqrt{G_t + \epsilon}}\odot\nabla L(x_t)
\end{equation}
$G_t = \sum_{\tau=1}^{T}g_{\tau}g_{\tau}^{T}$ is a diagonal matrix where each diagonal element $G_{t,ii}$ is the sum of the squares of all past gradient w.r.t. $x_i$ up to time step $t$. And $\epsilon$ is a smoothing term to avoid division by zero.
A vectorized implementation has the following form:
\begin{equation}
    x_t = x_{t-1}-\frac{\eta}{\sqrt{G_t + \epsilon}}\odot\nabla L(x_t),
\end{equation}
where  $\odot$ is an element-wise matrix-vector multiplication. 
In AdaGrad, each dimension has its own dynamic LR rate which is inversely dependent to the gradient magnitude, thus larger gradients have smaller LRs and small gradients have larger LRs. This is very beneficial for training DNNs since the scale of gradients in each layer is often different by several orders of magnitude. In addition, this accumulation of gradients can be regarded as a kind of simulated annealing which reduces the LRs along the course of training.  Most implementation set the LR $\eta$ to a default value of 0.01, eliminating the need of manual tuning.
However, AdaGrad holds a main drawback with the accumulation of squares of all past gradients, which keeps growing during the course of training. As the LRs radically shrink and vanish, the algorithm no longer gain additional knowledge.

\subsubsection{Adadelta}
Adadelta~\cite{Adadelta} is an extension of AdaGrad~\cite{AdaGrad} that seeks to tackle its monotonically decreasing LRs. Instead of accumulating the sum of all previous squared gradients, Adadelta uses an exponentially decaying average instead. The running average $E[g^2]_t$ at time step $t$ depends on the previous average and the current gradient
\begin{equation}
    E[g^2]_t = \rho E[g^2]_{t-1} + (1-\rho) g_t^2,
\end{equation}
where $\rho$ is a decay constant similar to that used in the momentum method. As the denominator is the root mean squared (RMS) error criterion of the gradient, we can replace it with the criterion short-hand:
\begin{equation}\label{eq: adadelta}
    \begin{split}
        RMS[g]_t &= \sqrt{E[g^2]_t+\epsilon}\\
        \Delta x_t &= -\frac{\eta}{RMS[g]_t}g_t.
    \end{split}
\end{equation}
Noticing the mismatch of units in Eq.[\ref{eq: adadelta}], i.e., the units of the update $\Delta x$ do not match the units of the parameters $x$ which it applies to
\begin{equation}
    units\,of\, \Delta x \propto units\, of\, g \propto \frac{\partial f}{\partial x} \propto \frac{1}{units \, of\, x},
\end{equation}
\citet{Adadelta} rearranges second order method (i.e., Newton’s method)
\begin{equation}
    \Delta x = \frac{\frac{\partial f}{\partial x}}{\frac{\partial^2 f}{\partial x^2}} \Rightarrow \frac{1}{\frac{\partial^2 f}{\partial x^2}} = \frac{\Delta x}{\frac{\partial f}{\partial x}}.
\end{equation}
Since $\Delta x_t$ for the current time step in unknown,  assuming the curvature is locally smooth, $\Delta x_t$ can be approximated by computing the exponentially decaying RMS of previous $\Delta x$
\begin{equation}
    \Delta x_t = - \frac{RMS[\Delta x]_{t-1}}{RMS[g]_t}g_t.
\end{equation}

\subsubsection{RMSProp}\label{subsection: RMSProp}
RMSprop~\cite{RMSProp} was developed independently around the same time with Adadelta to solve the problem of AdaGrad's drastically decreasing gradients. AdaGrad treats all past gradients equally, which is counter to our intuition that fresh gradient is more informative than the elder one. RMSProp redefines $v_t$ by decaying the past gradients at an exponential rate 
\begin{equation}
    \begin{split}
        v_t = 0.9 v_{t-1} + 0.1 g_t^2\\ 
x_t = x_{t-1}-\frac{\eta}{\sqrt{v_t+\epsilon}}g_t.
    \end{split}
\end{equation}

\subsubsection{Adam}\label{subsection: Adam}
Adam~\cite{kingma2017adam} is one of the most popular optimizers for training DNNs nowadays. It computes individual LRs for different parameters base on the estimates of first and second moments of the gradients. In particular, Adam stores an exponentially moving average of past gradients ($m_t$) and squared gradients ($v_t$). The former is an estimate of the first momentum (the mean) and the latter is an estimate of the second momentum (the uncentered variance) of the gradients
\begin{equation}
    \begin{split}
        m_t = \beta_1m_{t-1}+(1-\beta_1)g_t\\
v_t = \beta_2v_{t-1}+(1-\beta_2)g_t^2.
    \end{split}
\end{equation}
 $\beta_1$ and $\beta_2$ are hyper-parameters controlling the decaying rates of theses moving averages. Since the moving averages are initialed as 0’s, the estimates of first and second moments are biased towards zero, especially in the beginning of training. Adam utilizes correction terms to counteract the initialization bias
\begin{equation}
    \begin{split}
\hat m_t = \frac{m_t}{1-\beta_1^t}\\
\hat v_t = \frac{v_t}{1-\beta_2^t}       .
    \end{split}
\end{equation}
Then Adam applies the update rule
\begin{equation}
    x_t=x_{t-1}-\frac{\eta}{\sqrt{\hat v_t+\epsilon}}\hat m_t.
\end{equation}

% \xiaozhe{Hi, Xiaoxin, I think in this paragraph, we should cite this paper \cite{reddi2019convergence}, which addressed the most important problem of Adam optimizer, it may fail to converge to optimal solution even in convex setting, or even diverge in DL training. this paper is the best paper of ICLR 2018.}

Adam is found to be robust and well-suited to a wide range of non-convex optimization problems in the field of DL. There are several variants of Adam. AdaMax~\cite{kingma2017adam} is an extension to Adam that generalizes the approach to the infinite norm (max) and may result in a more effective optimization on some problems.  Nesterov-accelerated Adaptive Moment Estimation (NAdam)~\cite{NAdam} incorporates NAG into Adam. It shows better convergence speed in some cases.   While these algorithms have been successfully employed in several practical applications, they may fail to converge to optimal solution even in convex setting, or even diverge in DL training. \citet{reddi2019convergence} pinpoint the exponential moving average of past squared gradients as a reason for such failures. Recall that the introduction of the exponential average was well-motivated to tackle the key flaw of the Adagrad algorithm: it should prevent the LRs to become infinitesimally small as training progresses by limiting the reliance of the update on essentially only the past few gradients. However, this short-term memory of the gradients can indeed cause significant convergence issues in other scenarios. To resolve this issue, the authors propose new variants of Adam --- AMSGrad, which relies on long-term memory of past gradients. AMSGrad uses the maximum of past squared gradients rather than the exponential average to update the parameters.
\citet{RAdam} argue that the root cause of the bad convergence problem suffered by Adam is that the adaptive LR has undesirably large variance in the early stage of model training, due to the limited amount of training samples being used. Thus, to reduce such variance, it is better to use smaller LRs in the first few epochs of training. The authors propose Rectified Adam (RAdam) to rectify the variance of the adaptive LR.

Choosing an optimizer is a crucial step when training DNNs since it is woven with the training speed and the final predictive performance.
Despite the fact that adaptive optimization methods, including AdaGrad, RMSProp, AdaDelat and Adam, are becoming increasingly popular, to date, how to choose an optimal one  is still theoretically elusive and intractable. Instead practitioners rely on empirical studies~\cite{The_Marginal_Value_of_Adaptive_Gradient_Methods_in_Machine_Learning} and bench-marking~\cite{DeepOBS}. 
\citet{The_Marginal_Value_of_Adaptive_Gradient_Methods_in_Machine_Learning} observed that the solutions found by adaptive methods generalize worse (often significantly worse) than SGD, even when these solutions have better training performance. However, \citet{On_Empirical_Comparisons_of_Optimizers_for_Deep_Learning} suggest that popular adaptive gradient methods never under-perform momentum or gradient descent. They point out the comparisons among optimizers are sensitive to the hyper-parameter tuning protocols. 

\section{Large Batch Training}\label{sec: Large Batch Training}
\RestyleAlgo{ruled}
\SetKwInput{KwInit}{Init}
\begin{algorithm}[t]
\caption{Distributed Synchronous SGD on
Node k.}\label{alg: distributed synchronous SGD}
\KwIn{Dataset $X$,minibatch size $b$ per node, the number of nodes $N$, optimization function \textit{SGD}, init parameters $w={w[0],\cdots,w[M]}$}
\For{$t=0,1,\cdots$}{
$G_t^k\leftarrow0$\;
\For{$i=1,\cdots,B$}{
Sample data $x$ from $X$\;
$G_t^k \leftarrow G_t^k+\frac{1}{Nb}\nabla f(x;w_t)$
}
All-Reduce $G_t^k: G_t \leftarrow \sum_{k=1}^N G_t^k$\;
$w_{t+1}\leftarrow \textit{SGD}(w_t,G_t)$
}
\end{algorithm}
Large DNNs and large datasets have fueled the development of deep learning~\cite{KrizhevskySH12, image_classification, SimonyanZ14a, KrizhevskySH17,SzegedyLJSRAEVR15, bert}. However, training large models on massive datasets is compute-intensive.  For instance, training the SOTA DL models like BERT and ResNet-50 takes 3 days on 16 TPUv3 chips and 29 hours on 8 Tesla P100 gpus respectively~\cite{bert, image_classification}.
An intuitive way to accelerate training is to add more computational power (e.g., more GPU nodes) and use data parallel
% \xiaozhe{should we change to "data-parallel training", although SGD can represent a group of optimizer, but for most AI practitioners, SGD is narrowly means that exactly SGD optimizer, not including Adam.} 
(see Alg.\ref{alg: distributed synchronous SGD}).  
Considering communication (i.e., synchronizing the updates at each iteration) is an issue, each GPU must be utilized as much as possible to amortize the communication cost. Therefore, large batch should be used to distribute more data to each GPU. The nontrivial growth of batch size often results in test performance degradation, as observed in~\cite{one_weird_trick, generalization_gap_and_sharp_minima, Efficient_mini-batch_training_for_stochastic_optimization, Train_longer_generalize_better}. We describe the training difficulties introduced by large batch in Section~\ref{subsec: Large Batch Training Difficulties}, a recipe for large batch training (i.e., linear LR scaling with a warmup strategy) in Section~\ref{Learning Rate Scaling for Large Batch}, other supplementary strategies such as adaptive layer-wise learning in Section~\ref{subsec: Adaptive Layerwise Learning} and adaptive batch size in Section~\ref{subsec: Adaptive Batch Size}, and finally discuss the extent to which we can scale up the batch size in Section~\ref{subsec: Efficient Scaling}.

% Typically there are two ways to distribute training: data parallel and model parallel. Our discussion will mainly focus on the former as it is more widely adopted in distributed learning. 
% In data parallel training, each computational unit (e.g., GPU) receives a chunk of training data, calculates the gradients on the given data synchronously, communicates for a global gradient and finally updates the weights, and repeats until the convergence of model. 

\subsection{Large Batch Training Difficulties}\label{subsec: Large Batch Training Difficulties}
Although large batches are preferable to increase the parallelism by distributing the workload to multiple nodes, they may slow down convergence rate in practice~\cite{Sample_size_selection_in_optimization_methods_for_machine_learning}. 
% For general convex objective functions, the convergence of SGD is $\mathscr{O}(1/\sqrt{T})$; for mini-batch SGD with minibatch
% size $b$, the convergence is $\mathscr{O}(1/\sqrt{bT}+1/T)$~\cite{Optimal_Distributed_Online_Prediction_Using_Mini_Batches}. Since the total number of examples examined is $bT$ while there is only a $\sqrt{b}$ times improvement, the convergence speed degrades with increasing mini-batch size. 
Empirically, an increase in mini-batch size after a certain point (e.g. 1024) without a careful optimization scheme typically decreases the rate of convergence. The test accuracy of the converged solution becomes significantly lower than the baseline~\cite{Training_ImageNet_in_1_Hour, generalization_gap_and_sharp_minima, Train_longer_generalize_better, Efficient_mini-batch_training_for_stochastic_optimization}. In addition to a degradation of the test performance, \citet{abs-1804-07612} provide evidence that increasing the batch size also results in a progressively smaller range of LRs that allows stable training.

\citet{generalization_gap_and_sharp_minima} find a drop in generalization (often denoted as \textit{generalization gap}) to be as high as 5\% even for smaller networks, and correlate the generalization gap with the sharpness of the loss landscape. They argue that large-batch methods tend to converge to sharp minimizers of the training and testing functions, whereas small-batch methods consistently converge to flat minimizers. \citet{Train_longer_generalize_better} deny the existence of inherent generalization gap and suggest that training longer will help the algorithm to generalize better and keep the accuracy higher.  \citet{Training_ImageNet_in_1_Hour} admit that large batches cause optimization difficulties, but when these are addressed the trained networks exhibit good generalization. They tried to bridge the generalization gap with heuristics of LR scaling~\cite{Training_ImageNet_in_1_Hour} with a warpup strategy. However, empirical study ~\cite{two_regime} shows that LR scaling heuristics with the batch size do not hold across all problems or across all batch sizes. Later \citet{LARS}
proposed Layer-wise Adaptive Rate Scaling (LARS) to solve the large batch optimization difficulties. Several recent works successfully scaled the batch size to large
values using adaptive learning rates without degrading the performance.
\subsection{Learning Rate Scaling for Large Batch}\label{Learning Rate Scaling for Large Batch}
A nice property of large batch is its lower variance of the gradient. This is because when we take the gradient over more examples, the variance is obviously lower. Consequently, large batch allows us to take a larger step per iteration.
% Since increasing the batch size while keeping the training epochs unchanged means fewer iterations (parameter updates), higher learning rate should be used to allow larger step as a compensate.
Followings are two commonly used LR heuristics: linear scaling and sqrt scaling, to guide us to adapt the LR for large batches.

\subsubsection{Linear Scaling}\label{subsection: Linear Scaling}
\cite{one_weird_trick, Training_ImageNet_in_1_Hour, Optimization_Methods_for_Large_Scale_Machine_Learning} suggest linearly scaling up LR with batch size, i.e., when the mini-batch size is multiplied by $k$, multiply the LR by $k$. Intuitively, after $k$ iterations of mini-batch SGD , we have
\begin{equation}
    x_{t+k} = x_t - \eta\frac{1}{|S|} \sum_{i<k}\sum_{s\in S}\nabla l(x_{t+i},s),
\end{equation}
while after one iteration of large mini-batch $\bigcup_{j} B_j$ of size $|S|=k|B|$ we have
\begin{equation}
        \hat x_{t+1} = x_t - \hat \eta\frac{1}{k|B|} \sum_{j<k}\sum_{s \in B_j}\nabla l(x_t,s).
\end{equation}
If we assume $\nabla l(x_{t+i})\approx \nabla l(x_{t})$ for $i<k$, then the adjustment $\hat \eta = k\eta$ would yield $\hat x_{t+1} \approx x_{t+k}$. 
Noted that this assumption holds with the premises:
(1) $k$ cannot be infinite. That is, we cannot scale up the batch size without limits; (2) $t$ cannot be too small. Because at the beginning of training, the gradients change rapidly, and thus the difference between  $\nabla l(x_t)$ and $\nabla l(x_{t+i})$ is no longer negligible. Using LR warmup and linear scaling, \citet{Training_ImageNet_in_1_Hour} trained Resnet-50 with batch B=8K without loss in accuracy.
% Linear scaling has been successfully  adopted  in cases ~\cite{three_factor_influencing_minima_in_SGD}~\cite{one_weird_trick}~\cite{AdaBatch}~\cite{increase_batch_size}.

\subsubsection{Sqrt Scaling}\label{subsection: Sqrt Scaling}
Another scaling strategy is sqrt scaling, i.e., when the mini-batch size is multiplied by $k$, multiply the LR by $\sqrt{k}$. In SGD, the co-variance matrix of the parameters update $\Delta x$ is ~\cite{Train_longer_generalize_better}
\begin{equation}
    cov(\Delta x, \Delta x) \thickapprox \frac{\eta^2}{|B|}(\frac{1}{N}\sum_{n=1}^N g_ng_n^T).
\end{equation}
A simple way to keep this co-variance constant when we change the batch size is to choose $\eta \propto \sqrt{|B|}$. \citet{Train_longer_generalize_better} find that by using "Ghost Batch Normalization" and sqrt scaling, the generalization gap can be significantly decreased. However, the largest batch size used was 4,096, which does not rule out an effect appearing at still larger batch sizes, as suggested by the work of \citet{Training_ImageNet_in_1_Hour}. Moreover, establishing this invariant co-variance remains poorly justified, and often sqrt scaling is found to degrade model quality in practice, see ~\cite{one_weird_trick, Training_ImageNet_in_1_Hour, three_factor_influencing_minima_in_SGD}.

\subsubsection{Warmup}
After adjusting the LR with these strategies, the main obstacle for scaling up batch size is the instability of training with high LR, especially in the initial epochs when the gradients change dramatically. This issue can be alleviated by a properly designed \textit{warmup} strategy by using less aggressive LRs in the initial epochs.

\textbf{Constant warmup.}
\citet{DBLP:conf/uemcom/VermaQF17} use a low "safe" constant LR for the first few epochs of training and after that return to the target LR $\hat \eta = k\eta$. \citet{Training_ImageNet_in_1_Hour} find constant warmup particularly helpful for prototyping object detection and segmentation methods~\cite{Fast_R_CNN, Faster_R_CNN_Towards_Real_Time_Object_Detection_with_Region_Proposal_Networks, Mask_R_CNN}, but not sufficient enough to solve the large batch optimization problem. In particular, a transition out of the low LR warmup phase can cause the training error to spike. This motivates them to use a more moderate warmup stragegy --- gradual warmup.

\textbf{Gradual warmup.}
Unlike constant warmup, gradual warmup avoid a sudden increase of LR by gradually arising the LR from a small to a large value. We denote the LR of the $t$-th iteration as $lr(t)$ and the maximum LR during training as $lr_{max}$. Given a predefined time frame $T_{warmup}$, the LR scheduler for the $t$-th iterations is defined as
\begin{equation}
    lr(t) = \frac{t}{T_{warmup}}lr_{max},\quad t\leq T_{warmup}.
\end{equation}
After this warmup stage, the LR will be set by classical LR schedulers (e.g., cosine decay). A LR warmup stage is proved to be beneficial when training NNs with extremely large batch size~\cite{LAMB, Training_ImageNet_in_1_Hour}. \citet{RAdam} claim that the benefit of the warmup stage comes from reducing the variance for the adaptive LR in the Adam optimizer. They further propose Rectified Adam (RAdam) by introducing a term to rectify the variance of the adaptive LR. Additionally, \citet{On_Layer_Normalization_in_the_Transformer_Architecture} find the LR warm-up stage also helps quite a lot for other optimizers. 

\subsection{Adaptive Layerwise Learning}\label{subsec: Adaptive Layerwise Learning}
Linear/Sqrt LR scaling with warmup mitigates the vulnerability to the fluctuation of gradients in the initial epoch by taking less aggressive steps, starting from a small LR which is safe enough for all layers and gradually increasing it to the target value.  \citet{Train_longer_generalize_better} use less aggressive sqrt scaling with "Ghost Batch Normalization" to train Alexnet with $B=8K$, but still the accuracy ($53.93\%$) was much worse than baseline ($57.10\%$).  \citet{Training_ImageNet_in_1_Hour} use LR warmup and linear scaling to train Resnet-50 with batch B=8K without loss in accuracy. While these works demonstrate the feasibility of these strategies for reducing the wall time for training large DNNs, they are not general enough if we want further enlarge the batch size. For instance, \citet{LARS} applied linear scaling and warmup scheme to train Alexnet with batch normalization on Imagenet, and observed  a $2.2\%$ drop when $B=8K$ in the test accuracy. \citet{LARS} explain their method to solve this problem:
\textit{To analyze the training stability with large LRs we measured the ratio between the norm of the layer weights and norm of gradients update. We observed that if this ratio is too high, the training may become unstable. On the other hand, if the ratio is too small, then weights don’t change fast enough.} This ratio works like a hint about how to adapt the LR for each layer. In this section, we will first introduce a general adaptive layerwise strategy motivated by this ratio, followed by two specific algorithms, LARS~\cite{LARS} and LAMB~\cite{LAMB}.

\subsubsection{General Layerwise Strategy}

Suppose we use an iterative base algorithm $\mathscr{A}$ (e.g., SGD or Adam) in the small batch setting with the following layerwise update rule
\begin{equation}
    x_{t+1} = x_t +\eta_t u_t,
\end{equation}
where $u_t$ is the update made by $\mathscr{A}$ at time step $t$. \citet{LAMB} propose the following two changes to the update for large batch settings:
\begin{enumerate}
\item The update is normalized to unit $l_2$-norm. This is ensured by modifying the update to the form $u_t/\lVert u_t\rVert$. Such a normalization is done layer-wise, i.e., the update for each layer is ensured to be unit $l_2$-norm.
\item The LR is scaled by $ \phi(\lVert x_t\rVert)$ for some function $\phi: \mathbb{R}^+ \rightarrow \mathbb{R}^+$. Similar to the normalization, such a scaling is done layer-wise.
\end{enumerate}
Suppose the base algorithm $\mathscr{A}$ is SGD, then the modification results in the following update rule
\begin{equation}
    x_{t+1}^{(i)} = x_{t}^{(i)} -\eta_t \frac{\lVert \phi(x_{t}^{(i)}) \rVert}{\lVert g_t^{(i)}\rVert}g_t^{(i)}
\end{equation}
for all layers $i\in [h]$.
% and where $x_t^{(i)}$ and $g_t^{(i)}$ are the parameters and the gradients of the $i$th layer at time step $t$.
The normalization modification $g_t^{(i)}/\lVert g_t^{(i)}\rVert$ is similar to one typically used in normalized gradient descent except that it is done layer-wise. 
Normalization of this form provides robustness to exploding/vanishing gradients (where the gradient can be arbitrarily large/small) by  essentially ignoring the size of the gradient but preserving the  direction. As for the scaling step,  the scaling term involving $\phi$ ensures that the norm of the update is of the same order as that of the parameter. When the parameters are small, we take a small step and vice versa.

There are two notable differences between this general strategy and other adaptive algorithms such as Adam or RMSProp: (1) it uses a separate LR for each layer and not for each weight. (2) the magnitude of the update is controlled w.r.t the weight norm for better control of training speed. 
Both LARS~\cite{LARS} and LAMB~\cite{LAMB} are based on this general strategy, using momentum and Adam optimizer as the base algorithm respectively.
% Followings are methods scheduled to enable large batch training without losing model convergence.
% \subsection{Parallelism}
% There are three common strategies to distribute deep learning training across multiple processors: data parallel, model parallel, and hybrid parallel.

% \textbf{Data Parallel.}
% The data parallel approach distributes a mini-batch of training data evenly and replicates the model parameter across different processors. In each step, each processor executes the forward and backward propagation on a different subset of the mini-batch data, and uses averaged gradients across processors to update the model locally.

% \textbf{Model Parallel.}
% Increasing deep learning model size (layers and parameters) can result in better accuracy, however there is a limit to the maximum model size with respect to the capacity of the host memory of a single processor. The model parallel approach splits the model vertically, partitioning the computation and parameters in each layer across multiple devices, requiring significant communication between each layer. 

% \textbf{Hybrid Parallel.} 
% Model parallel and data parallel are not mutually exclusive but can be combined together. Hybrid parallel either applies different parallel approaches on different layers, or partitions processors into groups where model parallel is exploited within group while data parallel is used across groups.

% \textbf{Pipeline Parallel.}
% PP splits a model horizontally across layers running each partition on a different device and use micro-batching to hide the pipeline bubble. 

\subsubsection{LARS}
The first instantiation of the general strategy is the LARS algorithm ~\cite{LARS}, which is obtained by using momentum optimizer as the base algorithm $\mathscr{A}$ in the framework. LARS stands for Layer-wise Adaptive Rate Scaling, which was proposed for large batch learning for ResNet on ImageNet. Specifically, a local LR $\lambda^l$ is defined for each layer $l$
\begin{equation}
\lambda ^l = \eta \frac{{\lVert x\rVert}_2^2}{{\lVert \nabla L(x)\rVert}_2^2}.
\end{equation}
The hyper-parameter $\eta<1$  describes the extent to which we can trust the layer to update its weights during each epoch. At the beginning of training, the numerator ${\lVert x\rVert}_2^2$ above is relatively small. In contrast, the denominator ${\lVert \nabla L(x)\rVert}_2^2$  is probably large since when everything is wrong, the loss and gradients are large. Any steps we take are likely to be small. In this way we naturally warm up as the weights increase. As we approach 0 loss, the gradients become smaller and the local LR increases again, encouraging jumping out of the local minima to prevent over-fitting. The parameter update is
\begin{equation}
    \Delta x_t^l = \gamma * \lambda^l* \nabla L(x_t^l),
\end{equation}
where $\lambda$ is the global LR. In this way, each layer can learn at its own pace accurately. The training for SGD with LARS are summarized in the Algorithm \ref{alg: LARS}. 
\SetKwComment{Comment}{//}{}
\RestyleAlgo{ruled}
\SetKwInput{KwInit}{Init}
\begin{algorithm}[t]
\caption{{SGD with LARS. \\Example with weight decay, momentum and polynomial LR decay.}}\label{alg: LARS}
\KwIn{base LR $\gamma_0$, momentum $m1$, weight decay $\beta$, "trust" coefficient $\eta$, number of steps $T$}
\KwInit{$t = 0$; $v = 0$. Init weight $w_{0}^l$ for each layer $l$}
\While{$t < T$ for each layer $l$}{
$g_t^l \leftarrow \nabla L(w_t^l)$ \Comment*[r]{obtain a stochastic gradient for the current mini-batch}
$\gamma_t \leftarrow \gamma_0 * (1-\frac{t}{T})^2$\Comment*[r]{compute the global LR}
$\lambda^l\leftarrow \frac{\lVert w_t^l\rVert}{\lVert g_t^l\rVert+\beta \lVert w_t^l\rVert}$\Comment*[r]{compute the local LR}
$v_{t+1}^l \leftarrow mv_t^l + \gamma_t * \lambda^l * (g_t^l+\beta w_t^l)$\Comment*[r]{update the momentum}
$w_{t+1}^l \leftarrow w_t^l - v_{t+1}^l$ \Comment*[r]{update the weights}
}
\end{algorithm}

% Since we can now be more confident of each step, the cautionary warm-up often used in LR schedules is no longer necessary (as proposed by LARS) and we can scale to much bigger batch sizes without diverging. However, in the paper, the author only presents the experimental effect of warm-up + LARS in ImageNet data set, which is less persuasive about the widespread effectiveness of this method.
Several works successfully scaled the batch size to large values using LARS without degrading the performance, thereby, finishing ResNet-50 training on ImageNet in a few minutes~\cite{LARS, Image_Classification_at_Supercomputer_Scale, DBLP:journals/corr/abs-1903-12650}. LARS also applies to tasks such as self-supervised image representation learning and contrastive learning of visual representations~\cite{GrillSATRBDPGAP20, ChenK0H20}.

\subsubsection{LAMB}

LAMB is the second instantiation of the general strategy, which is obtained by using Adam as the base algorithm $\mathscr{A}$. The pseudo-code is provided in Algorithm~\ref{Alg: LAMB}. The adaptivity of LAMB is two-fold: (1) per dimension normalization w.r.t the square root of the second moment used in Adam and (2) layer-wise normalization obtained due to layer-wise adaptivity. 
By using LAMB, \citet{LAMB} scale the batch size of BERT pre-training
to 64K without losing accuracy, thereby, reducing the BERT training time from 3 days to around 76 minutes. LAMB is also the first large batch adaptive solver that can achieve the SOTA accuracy on ImageNet training with RESNET-50. LAMB has also been adopted by many other work ~\cite{ALBERT}. 

Despite of the popularity of LARS and LAMB, their utility as a "large batch optimizer" is challenged by ~\cite{opp_LARS_LAMB}, which argues that they are more indirect regularizers than optimizers. By sophisticated tuning, traditional, generic algorithms (e.g., Momentum or Adam) achieve strong results across batch size. They appeal to researchers that the superiority of one particular optimizer over others should be claimed with extreme caution since the fair comparisons between optimizers crucially depend on the effort spent tuning hyperparameters for each optimizer.

\begin{algorithm}[t]
\caption{{LAMB}}\label{Alg: LAMB}
\KwIn{$x_1\in \mathbb{R}^d$, LR $\{\eta_t\}_{t=1}^{T}$, parameters $0<\eta_1, \eta_2<1$, scaling function $\phi$, $\epsilon>0$}
\KwInit{Set $m_0=0,\,v_0=0$}
\For{$t=1$ to T}{
$g_t = \nabla L(x_t)$\Comment*[r]{obtain a stochastic gradient for the current mini-batch}
$m_t = \beta_1 m_{t-1}+(1-\beta_1)g_t$\;
$v_t = \beta_2 v_{t-1}+(1-\beta_2)g_t^2$\;
$m_t = m_t/(1-\beta_1^t)$\;
$v_t = v_t/(1-\beta_2^t)$\;
$r_t = \frac{m_t}{\sqrt{v_t}+\epsilon}$\;
$x_{t+1}^{(i)} = x_{t}^{(i)} - \eta_t \frac{\phi(\lVert x_{t}^{(i)} \rVert)}{\lVert r_{t}^{(i)} + {\lambda x_{t}^{(i)}}\rVert}(r_{t}^{(i)}+ {\lambda x_{t}^{(i)}})$
}
\end{algorithm}

\subsection{Adaptive Batch Size}\label{subsec: Adaptive Batch Size}
% The use of gradual LR warmup and linear scaling enables batch sizes of 8192 for ImageNet CNN training in a large, distributed GPU cluster setting~\cite{Training_ImageNet_in_1_Hour}. And the use of a adaptive layer-wise scaling allows even higher batch sizes of 32,768~\cite{LARS}.
% Both results use a fixed batch size throughout training whereas ~\cite{AdaBatch}~\cite{increase_batch_size} suggests changing the batch sizes during training. They show empirically that increasing the batch size and decaying the LR are quantitatively equivalent. 
It is a common practice to decay the LR during training. When one decays the LR, one simultaneously decays the "noise scale", i.e., the scale of random fluctuations in the SGD dynamics~\cite{A_Bayesian_Perspective_on_Generalization_and_Stochastic_Gradient_Descent}
\begin{equation}
\begin{split}
        g &= \frac{\epsilon}{1-m}(\frac{N}{B}-1)\\
        &\thickapprox \frac{\epsilon N}{(1-m)B}.
\end{split}
\end{equation}
When we decay the LR, the "noise scale" falls, enabling us to converge to the minima of the loss function.  We can achieve the same reduction in noise scale at constant LR by increasing the batch size. \citet{increase_batch_size} and \citet{AdaBatch} empirically demonstrated the equivalence between decaying LR and increasing the batch size. Instead of decaying the LR by a factor of $\alpha$, they increase the batch size by $\alpha$ during training. This strategy reaches equivalent test accuracy after the same number of training epochs, but with fewer parameter updates, leading to greater parallelism and shorter training times. 
Crucially, such strategy is complementary to existing training schedules requiring no hyper-parameter tuning.
% To reduce the number of parameter updates required to train a model, they proposed to scale batch size $B \propto \frac{\epsilon}{1-m}$ by synchronously increasing the LR $\epsilon$ and momentum coefficient $m$. 

% The initial noisy optimization phase allows us to explore a larger fraction of the parameter space without becoming trapped in local minima. Once we have located a promising region of parameter space, we reduce the noise to fine-tune the parameters.

\subsection{Efficient Scaling}\label{subsec: Efficient Scaling}
Increasing the batch size is one of the most appealing ways to accelerate NN training on data parallel hardware. Ideally, parallel mini-batch SGD can achieve a linear speed-up of the training time w.r.t. the number of workers compared with SGD over a single worker. However, such linear scalability in practice is significantly limited by the growing demand for gradient communication as more workers are involved. Moreover, when batch very large, the stochastic gradients become very close to true gradients, so increasing the batch does not give much additional gradient information comparing to smaller batches.

A series of work has conducted comprehensive experiments on the relationship between batch size and training time for NNs~\cite{two_regime, smith2020generalization, zhang2019algorithmic}. ~\citet{two_regime} experimentally measure the effects of data parallelism training across different families of NNs, training algorithms and data sets, finding no evidence that larger batch sizes degrade out-of-sample performance. 
They observed three distinct scaling regimes in the relationship between batch size and training time: a "\textit{perfect scaling}" regime where doubling the batch size halves the number of training steps required to reach a target out-of-sample error, followed by a regime of "\textit{diminishing returns}", and finally a "\textit{maximal data parallelism}" regime where further increasing the batch size does not reduce training time, even assuming idealized hardware. They also provide experimental evidence that the critical batch size depends on the model architecture, the dataset and regulation technology. 
% Take-aways are: (1) Momentum SGD performs similarly to plain SGD in the regime of small batch sizes but helps in the large-batch regime. (2) Some models allow training to scale to much larger batch sizes than others (3)

\section{Generalization Gap}\label{section: Generalization Gap}
 Optimization in general is an extremely difficult task, especially for training NNs. With non-convex and high-dimensional functions, it is possible to have many local minima and saddle points. Optimization methods, such as SGD, generally converge to different regions of parameter space, highly dependent on the design of network architecture, the choice of optimizer, variable initialization, and a variety of other considerations~\cite{two_regime}. The term generalization refers to how well a hypothesis applies even to new examples that it hasn't seen in the training set. As mentioned in Section~\ref{subsec: Large Batch Training Difficulties}, it is observed that while yielding similar values of training functions, models trained with large-batch methods perform worse on test data compared to small-batch methods~\cite{generalization_gap_and_sharp_minima, Train_longer_generalize_better, two_regime, abs-1804-07612}. Such persistent degradation in generalization performance is referred to as the \textit{generalization gap}. Identifying the origin of this gap and finding ways to close it is of significant practical importance whereas remains an open problem.

This section is structured as follows. Section~\ref{subsec: Sharp and Flat (Wide) Minima} introduces the concept of sharp and flat (wide) minima;  Section~\ref{subsec: Generalization Gap and Sharp Minima} addresses the relationship between sharpness/flatness of local minima and their generalization ability; Section~\ref{subsec: Gradient Noise Ratio} provides explanation for the so-called generalization gap and Section~\ref{subsec:  Train longer, Generalize Better} provides a somewhat opposing account.

\subsection{Sharp and Flat (Wide) Minima}\label{subsec: Sharp and Flat (Wide) Minima}
\begin{figure}[t]
    \centering
    \includegraphics[scale=0.25]{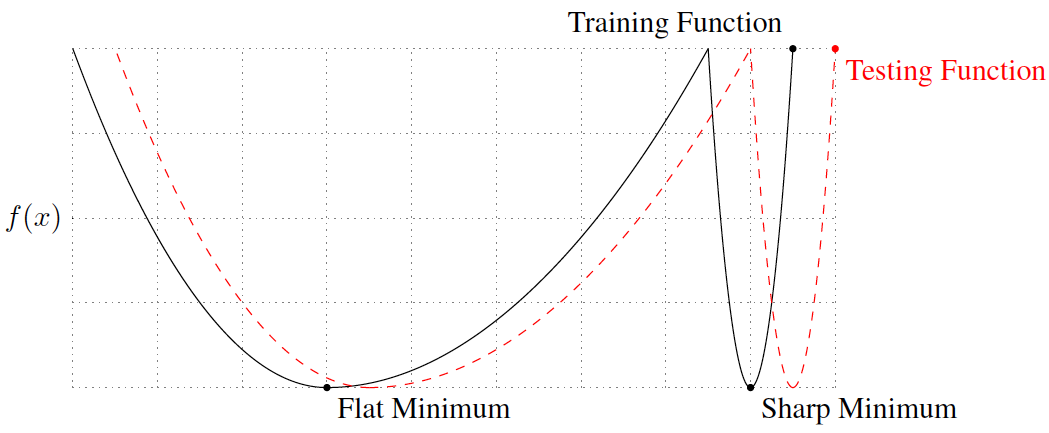}
    \caption{A Conceptual Sketch of Flat and Sharp Minima, src: ~\cite{generalization_gap_and_sharp_minima}}
    \label{fig: Flat and Sharp Minima}
\end{figure}

When training a DL model, we are seeking for a solution that minimizes a loss function on a given training set. This solution lies in a very high dimensional space (thousands, millions or even billions of parameters to learn) called parameter space. The landscape of parameter space is showed empirically crucial to generalize well.  That being said, the wider the solution’s local geometry, the better the generalization~\cite{entropy_SGD, generalization_gap_and_sharp_minima, Visualizing_the_Loss_Landscape_of_Neural_Nets}. Figure ~\ref{fig: Flat and Sharp Minima} provides an intuitive explanation. There is generally a shift of the loss function in the parameter space, flat minima is more robust to the perturbation of parameter than the sharp one and thus generalizes better.

There are various definitions for "sharpness/flatness" of the landscape. \citet{hochreiter1997flat} define "flatness" as a large connected region in weight space where the error remains approximately constant. \citet{generalization_gap_and_sharp_minima} characterize "flatness" by the magnitude of the eigenvalues of Hessian, and propose a computational feasible $\epsilon$-sharpness measure. \citet{Sharp_Minima_Can_Generalize}  show that flat minima in practical DL hypothesis spaces can be turned into sharp minima via re-parameterization without affecting the generalization gap. \citet{entropy_SGD} exploit the local geometric properties of the objective function and use "local entropy" as a measure of "flatness", which is invariant to the simple re-parametrization in ~\cite{Sharp_Minima_Can_Generalize}. \citet{foret2021sharpnessaware} capture the "sharpness" at parameter $w$ by measuring how quickly the training loss can be increased by moving from $w$ to a nearby parameter value.

Empirically, optimizers like SGD, Adam, etc. implicitly converge towards wide valleys solutions. But there is no guarantee that this will always be the case.  This has motivated the creation of algorithms that will actively look for flat minima such as Entropy SGD~\cite{entropy_SGD}, Sharpness-Aware Minimization (SAM)~\cite{foret2021sharpnessaware} and many others.
\subsection{Generalization Gap and Sharp Minima}\label{subsec: Generalization Gap and Sharp Minima}

With regards to large batch training, Keskar et al.~\cite{generalization_gap_and_sharp_minima} observed that naively increasing the batch size typically results in degradation of generalization
performance and reduces computational benefit. They speculate that \textit{“the lack of generalization ability is due to the fact that large-batch methods tend to converge to sharp minima of the  training functions”}. Specifically, large-batch methods are more vulnerable to sharp minima whose training function increases rapidly in a relatively small neighborhood (see Fig.~\ref{fig: Flat and Sharp Minima}). 
Such kind of high susceptibility to training functions hampers the trained model from fitting on the test data. By contrast, small-batch methods can resist the attraction of these minima and converge to a flat minima whose training function varies slowly in a relatively large neighborhood. They contribute the success of small batch methods to their noisy gradients in the computation step. On one hand, the noise expels the iterations from the trap of sharp minima. On the other hand, the noise encourages them to move towards and retain in a flatter loss landscape. However, as a larger batch size depicts a more accurate gradient, their noise is no longer sufficient enough to eject the iterations from the basin of sharp minima. 

It is widely thought that small-batch SGD produces “flat” minima that generalize well, while large batches converge to “sharp” minima with poor generalization~\cite{hochreiter1997flat, generalization_gap_and_sharp_minima, entropy_SGD}. However, there are some disputes about the effects of batch size on model's generalization ability.  \citet{Train_longer_generalize_better} deny the existence of inherent generalization gap and show empirically that the "generalization gap" stems from the relatively small number of updates rather than the batch size, and can be completely eliminated by adapting the number of weight updates. \citet{Training_ImageNet_in_1_Hour} hold the view that optimization difficulty is the main issue with large mini-batches, rather than the poor generalization (at least on ImageNet). Specifically, using linear scaling and warmup strategy, they show no loss of accuracy when training with large mini-batch sizes up to 8,192 images on the ImageNet dataset. 

\subsection{Gradient Noise Ratio}\label{subsec: Gradient Noise Ratio}
As mentioned above, how batch size affects sharpness and generalization is controversial. \citet{A_Bayesian_Perspective_on_Generalization_and_Stochastic_Gradient_Descent} show that the test accuracy peaks at an optimal batch size, if one holds the other hyper-parameters constant. They believe that the arise of peak is not controlled by the batch size itself,  but the underlying scale of random fluctuations in the SGD dynamics. 

Consider a simple model of SGD; 
% the loss function $l(x)$ is a sum over the losses of individual examples in our dataset $l(x)=\frac{1}{N}\sum_{i=1}^{N}l_i(x)$; mini-batch size is $B$; LR is $\eta$;
the estimated gradient step is $\eta \nabla_{SGD}(x)=\frac{\eta}{|B|}\sum_{i \in B}\nabla l_i(x)$, which can be restated as the true gradient and a gradient noise term
\begin{equation}\label{eq: noisy term}
    \eta \nabla_{SGD}(x)
    =\underbrace{\eta \nabla l(x)}_{gradient}+\underbrace{\frac{\eta}{|B|}\sum_{i \in B}(\nabla l_i(x)-\nabla l(x))}_{noise\,term}.
\end{equation}
\citet{A_Bayesian_Perspective_on_Generalization_and_Stochastic_Gradient_Descent} analogy between SGD and stochastic differential equations (SDEs) to describe the noise in the SGD dynamics. In particular, they depicted Eq.[\ref{eq: noisy term}] as the discrete update of a stochastic differential equation (SDE) and derive an analytical expression for the stochastic "noise scale"  $g= \eta (\frac{N}{B}-1)\approx \eta \frac{N}{B}$, which controls the scale of random fluctuations in the SGD dynamics. 
 Noise drives SGD away from sharp minima, and therefore there is an optimal batch size which maximizes the test accuracy. This optimal batch size is proportional to the LR and training set size $B_{opt}\propto \eta N$. Therefore, they attribute the so-called "generalization gap" observed in ~\cite{Sharp_Minima_Can_Generalize} as a consequence of scaling batch size above this optimal batch size. Similarly,  \citet{three_factor_influencing_minima_in_SGD} derive a "stochastic noise" using a different SDE. They verify experimentally that the ratio of LR to batch size,  $\eta / B$, influences the width of the minima found by SGD, and that higher values of the ratio lead to wider minima and often better generalization. Despite the slightly difference in the form of "stochastic noise", both~\cite{A_Bayesian_Perspective_on_Generalization_and_Stochastic_Gradient_Descent, three_factor_influencing_minima_in_SGD} 
 indicate that gradient noise can be beneficial, especially in non-convex optimization. Also they theoretically explain the empirical finding in ~\cite{Train_longer_generalize_better, Training_ImageNet_in_1_Hour} that rescaling the LR with the square root of the batch size and train for more epochs, or linearly with batch size, can reach the same generalization with a large batch size.

\subsection{Train longer, Generalize Better}\label{subsec: Train longer, Generalize Better}
Another observation in ~\cite{generalization_gap_and_sharp_minima} is that large batch methods are more likely to be attracted to minima close to the initial point, whereas small batch methods are more explorative and always locate minima that are farther away, with a ratio of $\lVert x_S^{*}-x_0 \rVert_2/\lVert x_{L}^{*}-x_0\rVert$  in the range of 3 to 10.
\citet{Train_longer_generalize_better} further find that the weight distance from initialization point increases logarithmically with the number of training iterations (weight updates), $\lVert w_t - w_0 \rVert \sim log\,t $. They therefore deny the existence of inherent generalization gap and believe that "generalization gap" stems from the relatively small number of updates rather than the batch size. Specifically, they "stretched" the time-frame of the optimization process, where each time period of $e$ epochs in the original regime will be transformed to $\frac{B_L}{B_S}e$ epochs according to the mini-batch size used. However, such modification anneals the speedup effect of large batch training.

\section{Second Order Optimization}\label{sec: Second Order Optimization}
% rewrite
Optimizations in DL, both theoretically and empirically, are presently dominated by first-order gradient methods~\cite{Finding_Approximate_Local_Minima_for_Nonconvex_Optimization_in_Linear_Time, agarwal2016second,bollapragada2016exact,Accelerated_Methods_for_NonConvex_Optimization,Trust_Region_Methods,Second_order_Optimization_for_Non_convex_Machine_Learning_an_Empirical_Study}. Second-order optimization methods that involve second derivatives and/or second order statistics of the data, are far less prevalent despite strong theoretical properties, due to their prohibitive computation, memory and communication costs.  In this section, we setup second-order optimization basics in Section~\ref{subsec: Second-Order Optimization Basics}, start from the classical Newton's method in Section~\ref{subsec: Newton's Method}, and turn to some up-to-date algorithms such as the Hessian-Free Method (in Section~\ref{subsec: Hessian-Free Method}), K-FAC (in Section~\ref{subsec: K-FAC}) and Shampoo~\cite{shampoo} (in Section~\ref{subsec: shampoo}). 

\subsection{Second-Order Optimization Basics}\label{subsec: Second-Order Optimization Basics}
Basically, many problems in machine learning can be simply described as minimizing the loss function over variables $x \in \mathbb{R}^d$
\begin{equation}
    \mathop{min}\limits_{x\in {\mathbb{R}}^{d}} F(x).
\end{equation}
When training the weights of a NN, we are trying to get as far down the error surface as possible. In most cases, we often use SGD to update the parameter vector to solve this optimization problem
\begin{equation}
\begin{split}
    x_{t+1} &= x_t - \eta_t g_t.
\end{split}
\end{equation}
% Here $g_t$ is a gradient more often a stochastic gradient of the loss function at the current iterate $t$ and $\eta_t$ is the step size. 
Another very popular family of algorithms  used in practice are the adaptive optimization algorithms (e.g., AdaGrad~\cite{AdaGrad}, Adadelta~\cite{Adadelta}, RMSProp~\cite{RMSProp}, Adam~\cite{kingma2017adam}, etc.). These are basically algorithms that update for each individual entry in the parameter vector. Each entry has its own step size which is an adaptive update using past gradients.
\begin{equation}
    x_{t+1}^{(i)}=x_t^{(i)}-\eta_{t}^{(i)}g_t^{(i)} \quad i=1,...,d
\end{equation}
And there are also momentum variants of these methods that have slightly different update rules.
Potentially more powerful family of algorithms are known as the preconditioned algorithms which use some matrices called preconditioners to transform the gradient before taking a step.
% \begin{equation}
%     w_{t+1}=w_t-B_t^{-1}g_t
% \end{equation}
In general, the idea of second-order optimization is to model the objective function $f$ by the local approximation
\begin{equation}
    f(x+\delta)\approx M(x)\equiv f(x)+ \nabla f(x)^T\delta+ \frac{1}{2}\delta^TB(x)\delta.
\end{equation}
Here, $B$ is a symmetric preconditioner and $\delta$ is the change in parameters. In Newton's method, $B=H$, or $B=H+\lambda I$. Fully optimizing $ M(x_t)$ w.r.t. $\delta$ gives
\begin{equation}\label{eq: second_order_basic}
    \delta^* = \mathop{argmin}_\delta  M(x_t)=-B^{-1}\nabla f,
\end{equation}
then apply the update
\begin{equation}\label{eq: preconditioning}
    x_{t+1} = x_t +\delta^*.
\end{equation}
This family includes algorithms mentioned above such as AdaGrad~\cite{AdaGrad}, Adam~\cite{kingma2017adam}, where the preconditioners are diagonal. But they can be more powerful if the metrics are not diagonal but full preconditioners, for exapmle, full AdaGrad, Natural Gradient~\cite{natural_gradient} and also classical algorithms like Newton's method, Quasi-Newton methods~\cite{L-BFGS, quasi_newton} and so on.
It is well-known in optimization that preconditioning often leads to faster convergence or better "condition number" in many different scenarios. But it comes with obvious caveats: supposing the number of parameter is $n$, we need (1) at least quadratic space $\Omega (n^2)$ in the dimension to store the preconditioner. (2) $n^3$ time to invert the preconditioner to apply to the gradient vector. Generally speaking, these second-order methods are not very practical where the cost of computation and memory is formidable in the DL settings. Alternatively practitioners use the diagonal approximation or using the SGD again.

Recently, there has been considerable advancement in the development of second-order methods, seeking a balance between between full matrics and the diagonal case. These methods usually approach preconditioners of the gradient in a modular way, which is as powerful (or nearly powerful) as the full matrix case, but can be used in practical like the diagonal case in terms of storage and run-time. Inspired by the idea of the natural gradient method~\cite{Adaptive_Method_of_Realizing_Natural_Gradient_Learning_for_Multilayer_Perceptrons}, \citet{K-FAC} use a Kronecker-factored approximation to the Fisher matrix as its preconditioning matrix that can be applied to multi-layer perceptrons (MLPs), which was subsequently extended to other architectures, such as convolutional neural networks (CNNs) ~\cite{KFAC_CNN} and recurrent neural networks (RNNs)~\cite{distributed_K_FAC}. Kronecker-factored preconditioners based on the structure of the Hessian and quasi-Newton methods have also been developed ~\cite{quasi_newton, Kronecker_factored_Quasi_Newton_Methods_for_CNNs}. 

% \subsection{Recipe for Preconditioning}
% The recipe for preconditioning from some previous work is actually pretty simple:
% \begin{equation}\label{eq: precondition_h}
%     H_t = \mathop{argmin}\limits_{H \succ 0}\{H^{-1} \bullet \overline{G}_t\ + \Phi(H)\}
% \end{equation}
% where
% \begin{equation}
%     \overline{G}_t = \sum\limits_{s=1}^{t}g_s{g_s}^T.
% \end{equation}
% The preconditioner at time $H_t$ is the minimizer of overall positive definite metrics of some objectives which make a balance between the dot product of the inverse of the preconditioner and the accumulative co-variance matrix of the gradients observed so far, and a potential function of the preconditioner. We will just focus on one choice of the potential function which is the trace of the preconditioner and in this case the minimization problem can be solved in a coarse form. We derive the algorithm known as AdaGrad~\cite{AdaGrad} which simply pick the preconditioner as the root of the cumulative co-variance matrix.
% \begin{equation}
%     \Phi (H) = Tr(H) \Rightarrow H_t = \overline{G}_t^{\frac{1}{2}}
% \end{equation}

% However, computing the root of such a huge matrix is impractical and thus some computational constraints should be placed on this minimization problem. Instead of searching the overall positive definite matrices (i.e, $H \succ 0$), we can actually search over those can be used at large scale in practice. More explicitly, this kind of constraint could be a low-rank matrix, a very sparse matrix or the combination of the two, or a sketched version of the preconditioner.

\subsection{Newton's Method}\label{subsec: Newton's Method}
Recall that in GD method, the gradient of a function is defined as the vector of partial derivatives.
\begin{equation}
    g_t \triangleq \nabla f(x)= <\frac{\partial f}{x_1},\frac{\partial f}{x_2},\cdots,\frac{\partial f}{x_n}>
\end{equation}
It means we are assuming that the error surface of the NNs locally looks and behaves like a circle. And we are ignoring all curvatures of the surface, which may lead our training to progress very slowly. To rectify this, we can use information from the second derivative of a function. The idea of Newton's method is to apply a linear transformation that turns ellipses into circles. If we apply that transformation to the gradient vector, it will be as if we were going downhill in a circular error surface. Formally, Newton's method use the Hessian matrix as preconditioner
\begin{equation}
    H=\left[ \begin{array}{cccc}
         \frac{\partial^2 f}{\partial x_1^2}&\frac{\partial^2 f}{\partial x_1\partial x_2}&   \cdots & \frac{\partial^2 f}{\partial x_1\partial x_n}\\
         \frac{\partial^2 f}{\partial x_2\partial x_1}& \frac{\partial^2 f}{\partial x_2^2}&\cdots&\frac{\partial^2 f}{\partial x_2\partial x_n}\\
         \vdots&\vdots&\ddots&\vdots\\
         \frac{\partial^2 f}{\partial x_n\partial x_1}&
         \frac{\partial^2 f}{\partial x_n\partial x_2}&
         \cdots&
         \frac{\partial^2 f}{\partial x_n^2}
    \end{array} \right], \quad H_{ij}=\frac{\partial^2f}{\partial x_i\partial x_j}.
\end{equation}
The Hessian is a function of the parameters and we need to take its inverse and multiply the gradient by that. Then we need to go some distance in that direction
\begin{equation}\label{eq: newton's method}
    \Delta x = - \eta H(x)^{-1}\nabla f(x).
\end{equation}
 If it is a truly quadratic surface and we choose the LR correctly, we will arrive at the minima of the surface in a single step. However, that single step involves something complicated which is inverting that Hessian matrix. Assuming that we 
only have a million parameters in our NN, the Hessian matrix will have a trillion terms which is completely infeasible to invert.

\textbf{Curvature Matrices.}
Each element in the curvature matrix specifies how the gradient in one direction changes as we move in some other direction. The off-diagonal terms in a curvature matrix correspond to "twists" in the error surface. A twist means that when you travel in one direction, the gradient in another direction changes. If we have a nice circular bulb, all those off-diagonal terms are zero. As we travel in one direction, the gradient in other directions doesn't change. But when we have an elliptical error surface, as we travel in one direction, the gradient in another direction changes.  This is actually what is going wrong with GD. As GD updates one of the weights, at the same time it is updating all the other weights, causing a change in the gradient for
the first weight. That means when we update it we may actually make things worse. The gradient may have actually reversed sign due to the changes in all the other weights. And so the more weights we get, the more cautious about changing each one of them we need to be, because the simultaneous changes in all the other weights can change the gradient of a weight.

\textbf{How to avoid inverting a huge matrix.}
The intensive computation of the curvature has limited the applications of second-order optimization methods in DL settings. To address this problem, there are various ideas in the literature. One very popular line of work looks at diagonal approximations, e.g.,  Adagrad~\cite{AdaGrad}, RMSProp\cite{RMSProp}, Adadelta~\cite{Adadelta} and many others~\cite{Diagonal_Gauss_Newton, SGD_QN}.  But these diagonal terms consist only a tiny fraction of the interactions, so we are ignoring most of the terms (nearly all of them) in the curvature matrix. And the experimental evidence indicates that there is limited or almost no improvement in practice when compared to well-tuned SGD with or without momentum (see ~\cite{Adadelta, Diagonal_Gauss_Newton}).  The benefits of these diagonal approaches seem to lie mainly in the ease of choosing the LR, but may not provide any fundamental benefits beyond that.  Another thing we could do is to approximate the curvature matrix with much lower rank matrix but capturing its main aspects. Limited-memory BFGS (L-BFGS)~\cite{L-BFGS} is the most well-known example. Again there is limited/non-existent empirical success for NN optimization. 

More recently, many focus are put on the trade-off between the full matrix and the diagonal case, seeking a balance between these two extreme case. Some researchers find something in between that is as powerful (or nearly powerful) as the full matrix case, but can be used in practical like the diagonal case in terms of storage and run-time. Some recent approaches for approximating a full-matrix preconditioner are K-FAC~\cite{K-FAC} and Shampoo~\cite{shampoo}. Others incorporate automatically the Hessian operator, such as Hessian-Free method~\cite{Hessian_free} and trust-region~\cite{Trust_Region_Methods, Second_order_Optimization_for_Non_convex_Machine_Learning_an_Empirical_Study}.

\subsection{Hessian-Free Method}\label{subsec: Hessian-Free Method}
The Hessian-Free method ~\cite{Hessian_free} is a quasi-Newton method that uses no low-rank approximations. Named "free" because it never explicitly computes the preconditioner $B$ but instead does approximate minimization of quadratic model $M(\delta)$ (see Eq.[\ref{eq: second_order_basic}]).  The Hessian-Free (HF) method is motivated by two observations. The first one being that it is relatively easy to compute the matrix-vector product $Hv$ for an arbitrary vectors $v$, e.g., use finite differences to approximate the limit.
\begin{equation}\label{eq: HF}
    Hv=\mathop{lim}\limits_{\epsilon \rightarrow 0}\frac{\nabla f(x+\epsilon v)-f(x)}{\epsilon}
\end{equation}
% $Hv$ is computed for the exact value of $H$, there is no low-rank or
% diagonal approximation here!
The second motivating observation is that linear conjugate gradient (CG) minimizes positive definite quadratic cost functions using only matrix-vector products, which is relatively easy to obtained (as shown in Eq.[\ref{eq: HF}]). Conjugate gradient is a very clever method that instead of trying to go straight to the minimum like in Newton's method, it tries to minimize in one  direction at a time. It starts off by taking the direction of GD and goes to the minimum in that direction that might involve re-evaluating the gradient or re-evaluating the error a few times to find the minimum in that direction. Once it is done, CG method now finds another direction and goes to the minimum in that second direction. The clever thing about the technique is that it chooses the second direction in such a way that doesn't mess up the minimization it already did in the first direction, which is called a conjugate direction. "Conjugate" means that as we go in the new direction we do not change the gradients in the previous directions. What CG achieves is that it gets to the global minimum of an $n$-dimensional quadratic surface in only $n$ steps. More importantly, in many less than $n$ steps on a typical quadratic surface, it will have reduced the error very close to the minimum value. And that's why we use it. As doing the full $n$ steps that would be as expensive as inverting the whole matrix, we are going to do many less than $n$ steps and  get quite close to the minimum. Pseudo-code for a simple variant of damped HF optimization is provided in Algorithm~\ref{alg: Damped Hessian-Free Optimization}.
\RestyleAlgo{ruled}
\SetKwInput{KwInit}{Init}
\begin{algorithm}[t]
\caption{Damped Hessian-Free Optimization}\label{alg: Damped Hessian-Free Optimization}
\For{$n=1$ to max-epochs}{
compute gradient $g_t=\nabla f(x_t)$\;
choose/adapt $\eta_t$ according to some heuristic\;
define the function $B_t(v)=Hv+\eta_t v$\;
$p_t = CGMinimize(B_t,-g_t)$\;
$x_{t+1}=x_t+p_t$
}
\end{algorithm}

However, common variants of HF don’t work particular well for NNs. Enhancements techniques (e.g., the Gauss-Newton approximation to the Hessian, early CG stopping, damping, etc.) are provided in ~\cite{Hessian_free}. More recently research has revealed that DNN learning is easier than previously thought using simple methods~\cite{On_the_importance_of_initialization_and_momentum_in_deep_learning}. Carefully tuned momentum methods suffice for dealing
with the curvature issues in deep and recurrent network training objectives without the need for sophisticated second-order methods. Despite SGD with or without momentum still being the most widely used and best method in most situations, the fact that HF uses 100-1000x fewer iterations than SGD supports the idea that a second order method can help a lot in principle, provided that we can make these iterations cheap enough to compute.

\subsection{K-FAC}\label{subsec: K-FAC}
Kronecker-Factored Approximate Curvature (K-FAC) ~\cite{K-FAC} is one of the natural gradient approximation methods where the preconditioner is a high-quality approximation of Fisher information matrix (FIM).  We first give a brief introduction about natural gradient descent and then explained the outline of K-FAC.

\subsubsection{Natural Gradient Descent}
% Natural Gradient Descent (NGD) measures how different two models are using KL-divergence between the conditional probability $P_{y|x}(\theta)$ that they present.
% NGD updates weight by inverse Fisher Information Matrix (FIM).
% \begin{equation}
%     x_{t+1} = x_t-\eta F(x_t)^{-1}\nabla f_i(x_t)
% \end{equation}
% Natural Gradient Descent (NGD) was introduced in \cite{natural_gradient_descent}.

Natural Gradient Descent (NGD)~\cite{natural_gradient} is a second order optimization method based on information geometry. NGD acquires the loss landscape correctly by using FIM as curvature of loss function and converges faster in term of iterations than a simple first-order method. The FIM associated with network's distribution $P_{y|x}(\theta)$ is
\begin{equation}\label{eq:FIM}
    F = E[\nabla log(p(y|x;\theta))\nabla log(p(y|x;\theta))^T] .
\end{equation}
Importantly, one property of $F$ is that it can be interpreted as  the negative expected Hessian of our model’s log likelihood~\cite{New_Insights_and_Perspectives_on_the_Natural_Gradient_Method}
\begin{equation}
    F=-E_{p(y|x;\theta)}[H_{log p(x|\theta)}].
\end{equation}
Knowing this result, we can see the role of $F$ as a measure of curvature of the log likelihood function. Thus the immediate application of $F$ is as drop-in replacement of $H$ in second order optimization methods.  Using KL-divergence to measures how different two models are, the update rule of NGD is
\begin{equation}
    \theta_{t+1} \leftarrow \theta_{t} - \eta_t F^{-1}\nabla f(\theta_t). 
    % x_{t+1} = x_t-\eta F(x_t)^{-1}\nabla f_i(x_t)
\end{equation}
Here the inverse of the FIM is applied to the gradient of loss, and the gradient preconditioned by the FIM is called the natural gradient. For the parameters of size $N$, the size of FIM is $N \times N$, and NNs used in DL tend to have a massive number of parameters (e.g., 60 million parameters in AlexNet for ImageNet classification) so the inverse of the FIM is intractable, and it limits the number of the applications of NGD to DL. In recent years, some works have proposed methods that approximate or avoid inversing the FIM.

% is an efficient method for approximating natural gradient descent in NN ~\cite{K-FAC}. While computing the FIM is complex, K-FAC employs a preconditioning scheme that approximates FIM as Kronecker products of smaller matrices, which are more efficiently invertible.

% \subsubsection{Kronecker Product.} The Kronecker product $A\otimes B$ where $A$ has size $m\times n$ and $B$ has size $p\times q$ is:

% \begin{equation}
%     A\otimes B = \left[\begin{array}{ccc}
%          a_{11}B& \cdots & a_{1n}B \\
%          \vdots&\ddots&\vdots\\
%          a_{m1}B&\vdots&a_{mn}B
%     \end{array}\right]
% \end{equation}
% The Kronecker product has two convenient properties: 
% \begin{equation}
%     \begin{split}
%         {(A\otimes B)}^{-1} &= {A}^{-1}\otimes{B}^{-1}\\
%         {(A\otimes B)}vec(C) &= vec(B^{T}CA)
%     \end{split}
% \end{equation}

\subsubsection{K-FAC Approximation.}
K-FAC approximates the FIM so that the inverse matrix is easy to calculate. Firstly, K-FAC approximates F as $\hat F$, a diagonal block matrix where each block represents one layer in a NN with $L$ layers
\begin{equation}
    \hat F = diag(\hat F_1,...,\hat F_l,...,\hat F_L).
\end{equation}
Next, each diagonal block matrix $F_l$ is approximated as a Kronecker product
\begin{equation}
    \hat F_l \approx A_{l-1} \otimes G_l  .
    % = a_{l-1}{a_{l-1}}^T \otimes g_{l}{g_l}^T
\end{equation}
This is called Kronecker factorization and $G_l$, $A_{l-1}$ are
called Kronecker factors, representing the gradient of the output of the $l$-th layer and the activation of the ($l$-1)-th layer respectively. By using the critical property of the Kronecker product of the matrices $(A\otimes B)^{-1}=A^{-1}\otimes B^{-1}$,
the inverse of $\hat F_l$ can be computed as
\begin{equation}
    {\hat F_l}^{-1} = {A_{l-1}}^{-1} \otimes {G_l}^{-1}.
\end{equation}
% We can then use ${\hat F_l}^{-1}$ to precondition the gradient $\nabla L$, and update the parameters $w_l$ in the $l$ th layer as follows:
% \begin{equation}
%     w_{l}^{(t+1)}=w_l^{(t)}-\eta^{(t)} {\hat F_l}^{-1}\nabla L_l(w_k^{(t)})
% \end{equation}
% To compute the preconditioned gradient ${\hat F_l}^{-1}\nabla L_l(w_k^{(t)})$, we can make use of the well-known identity $(A\otimes B)vec(x) = vec(BXA^T)$ to get
% \begin{equation}
%     {\hat F_l}^{-1}\nabla L_l(w_k^{(t)}) = {G_l}^{-1} \nabla L_l(w_k^{(t)}) {A_{l-1}}^{-1}
% \end{equation}
The final update step of parameters $w_l$ in the $l$-th layer is as follows:
\begin{equation}
    w_{l}^{(t+1)}=w_l^{(t)}-\eta^{(t)}{G_l}^{-1} \nabla L_l(w_l^{(t)}) A_{l-1}^{-1}.
\end{equation}
In most implementations, Tikhonov regularization is used to avoid ill-conditioned matrix inverses with K-FAC by adding a damping parameter $\gamma$ to the diagonal of $\hat F_l$~\cite{feature_decomposition_K_FAC, KFAC_CNN}
\begin{equation}
    (\hat F_l + \gamma I)^{-1} = {({A_{l-1}+ \gamma I})^{-1}} \otimes {({G_l+ \gamma I})^{-1}}.
\end{equation}
A standard K-FAC update step for one layer requires inverting two matrices $(A_{l-1}+\gamma I)$ and $(G_l+\gamma I)$, which can be computed implicitly using an alternative method based on the eigendecompostion of $\hat F_l$~\cite{KFAC_CNN, feature_decomposition_K_FAC}.
\begin{equation}
    \begin{split}
        V_1&=Q_G^TL_i(w_i^{(k)})Q_A\\
        V_2&=V1/(v_G(v_A)^T+\lambda)\\
        (\hat F_l + \gamma I)^{-1}\nabla L_i(w_i^{(k)}) &= Q_GV_2{Q_A}^T
    \end{split}
\end{equation}
In practice, practitioners avoid significant computation and communication by reducing the frequency of computing these factors and and their eigendecompositions, at the cost of introducing staled information. For example, \citet{feature_decomposition_K_FAC} update K-FAC statistics for every 500 iterations for ResNet scaling experiments on 64 GPUs.

\subsubsection{Distributed K-FAC} 
Some studies have used K-FAC and implemented the algorithm in a distributed computing environment~\cite{distributed_K_FAC, feature_decomposition_K_FAC}. With only 35 epochs and a 16K batch size, ResNet50 can be trained to achieve 75\% Top1 accuracy in ImageNet~\cite{distributed_K_FAC}. More recently,  \citet{feature_decomposition_K_FAC} scales up K-FAC for training CNNs. It mainly refers to the calculation scheme of preconditioned gradient in ~\cite{KFAC_CNN} and uses feature decomposition to replace matrix inversion. 
% Integrating techniques of layer-wise distribution, inverse-free second-order gradient evaluation, K-FAC approximation decoupling, and dynamic K-FAC update frequency, ResNet-50 on the ImageNet-1k dataset converges to the 75.9\% MLPerf ResNet-50 baseline with 18–25\% less time than SGD.

\begin{table}[t]
\tiny
    \centering
    \begin{tabular}{|c|c|c|c|}
    \hhline{|=|=|=|=|}
         \small Algorithm & \small Preconditioning&  \small Memory&  \small Computation\\\hhline{|=|=|=|=|}
         \makecell*[c]{{Full Matrix AdaGrad} \\
         \raisebox{-\totalheight}{\includegraphics[scale=0.2]{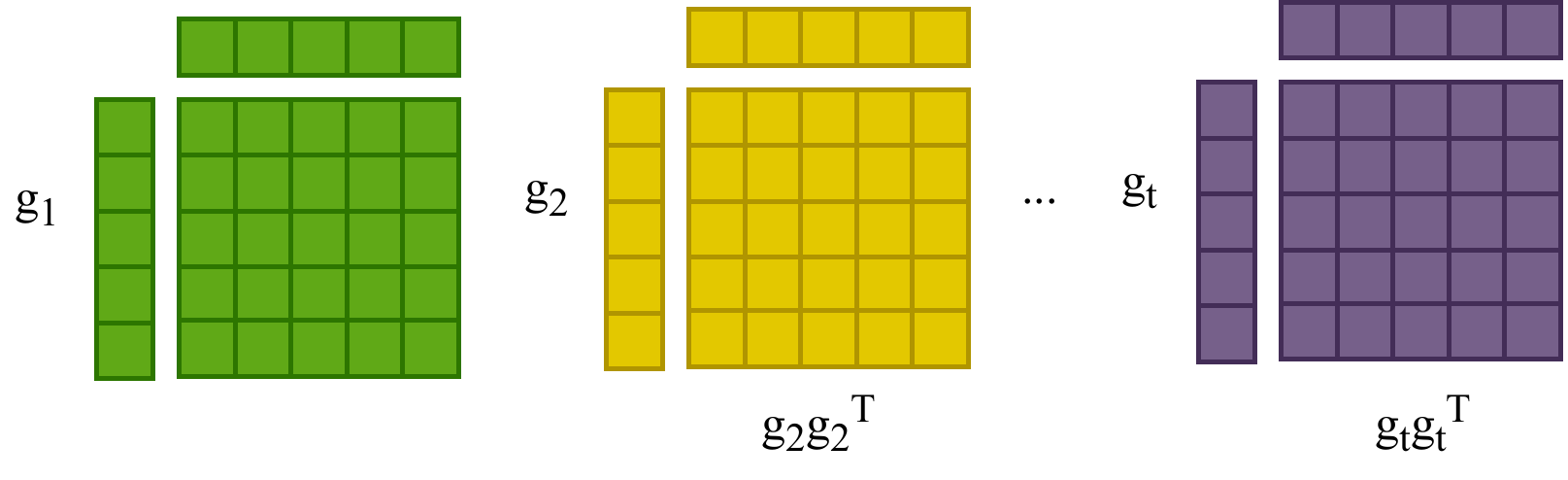}}}
         &\makecell*[c]{
         $H_t = (\sum_{s=1}^t g_sg_s^T)^{\frac{1}{2}}$\\
         $W_{t+1} = W_{t}-\eta_t H_{t}^{-1/2}$}&
         $O((mn)^2)$& 
         $O((mn)^2)$\\\hline
         \makecell*[c]{Shampoo\\
         \raisebox{-\totalheight}{\includegraphics[scale=0.25]{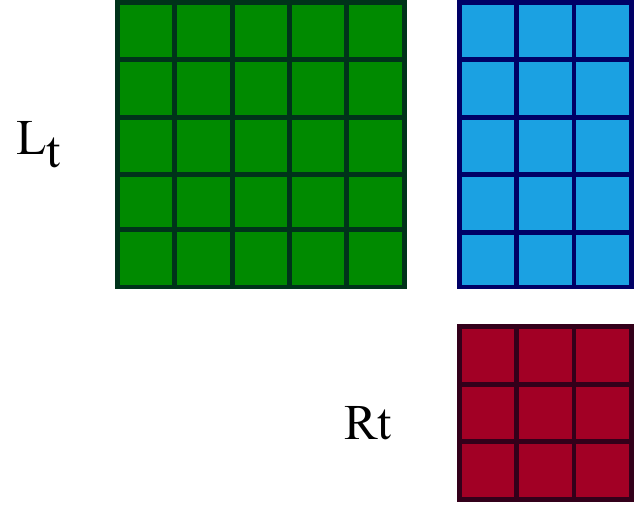}}}&
         \makecell*[c]{
         $L_t = L_{t-1}+g_tg_t^T$\\
         $R_t = R_{t-1}+g_t^Tg_t$\\
         $W_{t+1} = W_{t}-{L_t^{-1}G_tR_t^{-1}}$}&
         $O(m^2+n^2)$&
         $O(m^2+n^2)$\\\hline
         \makecell*[c]{K-FAC\\
         \raisebox{-\totalheight}{\includegraphics[scale=0.15]{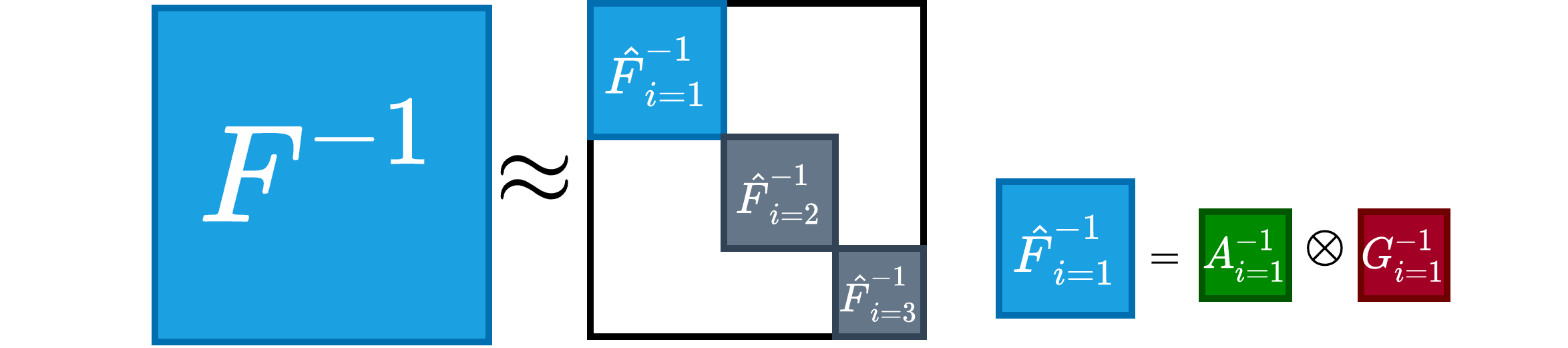}}}&
         \makecell*[c]{$\hat F_i = A_{i-1}\otimes G_i$\\$A_{i-1}=a_{i-1}a_{i-1}^T$\\$G_{i}=g_{i}g_{i}^T$\\
         $ W_{l}^{(t+1)}=W_l^{(t)}-\eta^{(t)}{G_l}^{-1} \nabla L_l(W_l^{(t)}) A_{l-1}^{-1}$}&
         $O(m^2+n^2)$&
         $O(m^3+n^3)$
         \\\hline
         \makecell*[c]{{Diagonal AdaGrad}\\
         \raisebox{-1\totalheight}{\includegraphics[scale=0.1]{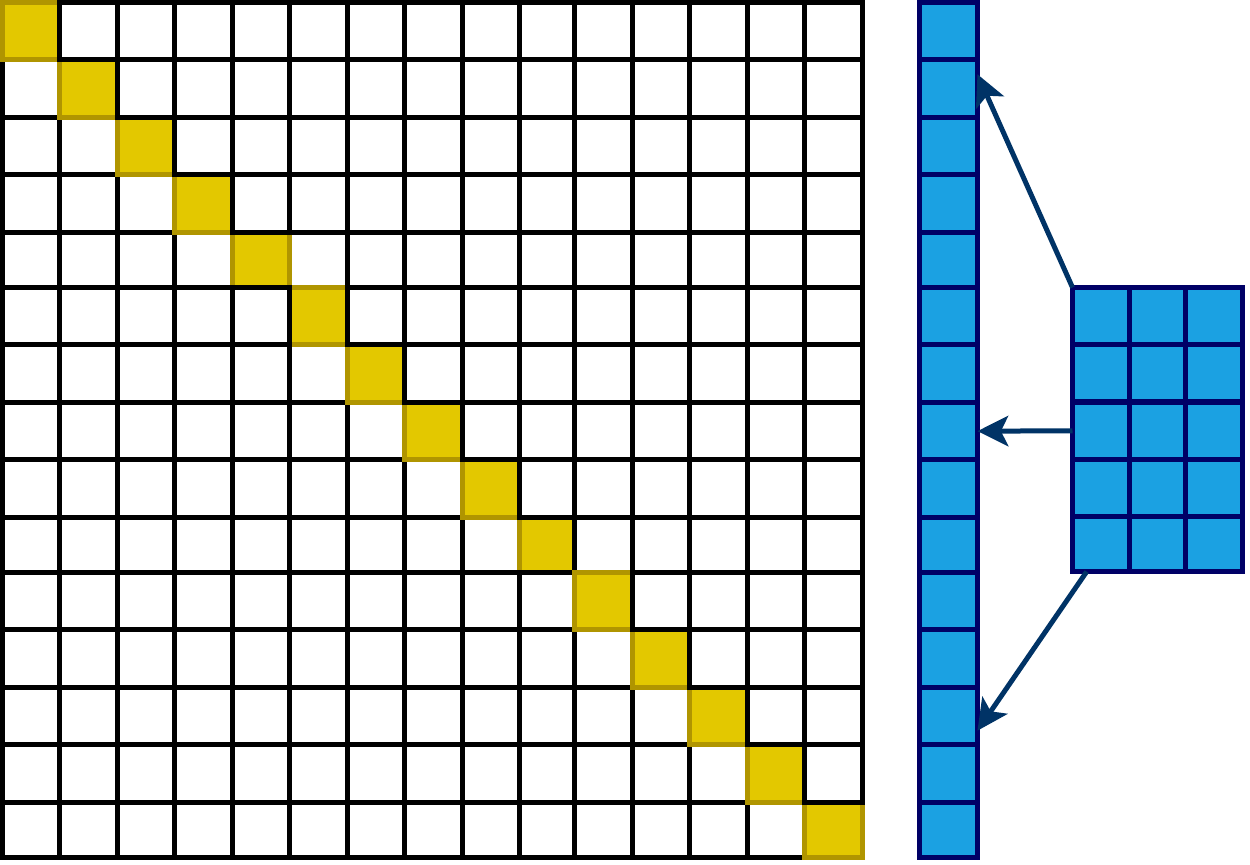}}}&
         \makecell*[c]{$H_{t,ij}=\sum_{s\leq t}g^2_{s,ij}$
         \\$W_{t+1} = W_{t}-\eta_t H_{t}^{-1/2}$}&
         $O(mn)$&
         $O(mn)$
         \\\hline
         \makecell*[c]{SM3\\\raisebox{-\totalheight}{\includegraphics[scale=0.25]{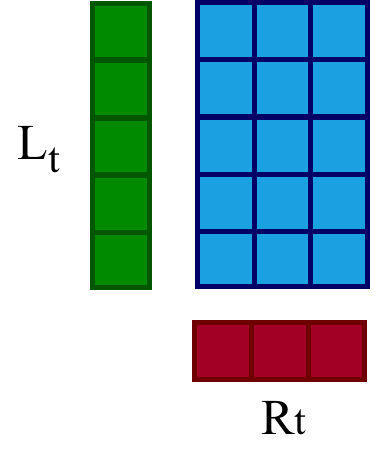}}}&
         \makecell*[c]{ 
         $\widehat{H_{t,ij}} = min(L_{t-1,i},R_{t-1,j}) + g_{t,ij}^2 $\\
         $ L_{t,i}=\mathop{max}\limits_j(\widehat{H_{t,ij}})$\\
         $ R_{t,j}=\mathop{min}\limits_i(\widehat{H_{t,ij}})$\\
         $W_{t+1, i} = W_{t, i} - \eta g_{t, i}{\widehat{H_{t,ij}}}^{-1/2}$} &
         $O(m+n)$&
         $O(mn)$
         \\\hline
    \end{tabular}
    \caption{Summary of Preconditioning Methods. Example of a fully connected layer $[m,n]$.}
    \label{tab: Memory} 
\end{table}
\subsection{Shampoo}\label{subsec: shampoo}

Shampoo ~\cite{shampoo} is another preconditioning algorithm, which is an approximation version of full matrix AdaGrad~\cite{AdaGrad}. It first approximates by treating each layer independently (block diagonal). And it uses small matrices whose Kronecker product approximates the full AdaGrad statistics.
Such two approximations make Shampoo more expressive than the diagonal preconditioning and practical to store and compute at large scale. 
Mathematically, the preconditioner Shampoo looking for can be written as a Kronecker product of two smaller matrices $L$ and $R$
\begin{equation}
    \mathop{argmin}\limits_{H=L\otimes R\atop L,\,R\succ 0}\,\{H^{-1} \bullet \overline{G_t}+Tr(H)\}.
\end{equation}
Though it cannot solve the exact optimization problem, it has a nice limit that relaxes the upper bounds in a matrix sense:
\begin{equation}
    \frac{1}{\sqrt{r}}\underbrace{(\sum\limits_{t=1}^T g_t (g_t)^T)^{\frac{1}{2}}}_{full\, AdaGrad \, precond.}\preceq \underbrace{(\sum\limits_{t=1}^{T}{G_t{G_t}^T})^{\frac{1}{4}}}_{L_t}\otimes\underbrace{{(\sum\limits_{t=1}^{T}{{G_t}^T G_t})^{\frac{1}{4}}}}_{R_t}
\end{equation}
The full AdaGrad peconditioner is given on the left, bounded by a Kronecker product of two smaller matrices. The update statistic of Shampoo is given as follows: 
\begin{equation}
\begin{split}
    L_t = L_{t-1}+G_tG_t^T, \quad R_t = R_{t-1}+G_t^TG_t\\
    W_{t+1} = W_{t}-{L_t^{-1}G_tR_t^{-1}}
\end{split}
\end{equation}

Both Shampoo and K-FAC employ a preconditioning scheme that approximates the FIM. Despite their similarity in construction, they differ in several important ways. The differences are based on choices such as the empirical FIM or FIM, moving average or sum, and the inverse component. Another key difference is that Shampoo construction is agnostic to layer types. K-FAC relies heavily on the structure of the back-propagated gradients in a feed-forward neural network. In contrast, Shampoo is virtually oblivious to the particular model structures and only depends on standard gradient information.
More recently, \citet{scalable_shampoo} extend Shampoo in a number of ways so as to make it applicable to a larger range of deep architectures. 

Despite the fact that first-order methods have been dominant in the recent decade,
recently second order methods, such as K-FAC and Shampoo, show some promise. They mitigate the space and run-time costs of full-matrix second-order algorithms and have been applicable to a larger range of deep architectures (see Table~\ref{tab: Memory}). It is interesting to see whether second order methods can outperform first order ones in the future.

% \begin{table}[t]
%     \centering
%     \begin{tabular}{cccccc}
%     \hline
%          AdaGrad& Shampoo & K-FAC & Diagonal AdaGrad& SM3 \\\hline
%          \raisebox{-\totalheight}{\includegraphics[width=0.2\linewidth]{figs/full_adagrad.pdf}}&\raisebox{-\totalheight}{\includegraphics[width=0.15\linewidth]{figs/shampoo.pdf}}&\raisebox{-\totalheight}{\includegraphics[width=0.2\linewidth]{figs/K_FAC.pdf}}&\raisebox{-\totalheight}{\includegraphics[width=0.15\linewidth]{figs/diag_adagrad.pdf}}&\raisebox{-\totalheight}{\includegraphics[width=0.1\linewidth]{figs/SM3.pdf}}\\
%          \tiny$H_t = (\sum_{s=1}^t g_sg_s^T)^{\frac{1}{2}}$&
%          \tiny\makecell*[c]{$L_t = L_{t-1}+g_tg_t^T$\\$R_t = R_{t-1}+g_t^Tg_t$}&
%          \tiny\makecell*[c]{$G$}&
%          \tiny$H_{t,ij}=\sum_{s\leq t}g^2_{s,ij}$&
%          \tiny \makecell*[c]{ 
%          $\widehat{H_{t,ij}} = min(L_{t-1,i},R_{t-1,j}) + g_{t,ij}^2 $\\
%          $ L_{t,i}=\mathop{max}\limits_j(\widehat{H_{t,ij}})$\\
%          $ R_{t,j}=\mathop{min}\limits_i(\widehat{H_{t,ij}})$}\\
%          &\tiny$W_{t+1} = W_t - \eta L_t^{-\frac{1}{4}}G_tR_t^{-\frac{1}{4}}$&&&\\
%          \hline
%     \end{tabular}
%     \caption{Comparison of Preconditioning}
%     \label{tab:my_label}
% \end{table}

\section{Communication}\label{sec: Communication}

Large-scale distributed training improves the productivity of training deeper and larger models, where data parallelism is adopted so as to take full advantage of the compute capability on multiple workers. SGD is usually selected as the optimization method because of its high computation efficiency and well support
by the DL tool-kits, such as TensorFlow~\cite{tensorflow}, PyTorch~\cite{pytorch} and DeepSpeed~\cite{DeepSpeed}.  In data-parallel SGD, each worker processes a random mini-batch of its training data, and then the local updates are synchronized by making an All-Reduce step or through a centralized parameter server, which aggregates stochastic gradients from all workers, and taking a Broadcast step that transmits the updated parameter vector back to all workers. The process of gradient synchronization is repeated until an appropriate convergence criterion is met.  

Increasing the number of workers and taking advantage of data parallelism help to reduce the computation time  on the same size training data dramatically. However, as the scale of distributed systems grows up, 
gradient and parameter synchronization prolongs the communication time and hinders the perfect scalability~\cite{Communication_Efficient_Distributed_Machine_Learning_with_the_Parameter_Server,TernGrad}. Therefore, the high network communication cost becomes a significant bottleneck of distributed training. There have been many attempts to reduce the communication overhead in data-parallel SGD. One notable method is to let each worker use compressed gradients rather than raw gradients for communication. For example, quantized SGD or sparcified SGD allow each worker to use fewer bits to pass gradients by sacrificing the convergence to a mild extent. Another notable method is to reduce the frequency of communication~\cite{Parallelized_SGD,Distributed_Training_Strategies_for_the_Structured_Perceptron,Parallel_SGD_When_does_averaging_help,Efficient_Decentralized_Deep_Learning_by_Dynamic_Model_Averaging,Use_Local_SGD}. We detail on these two communication-efficient methods in the subsequent sections.

\subsection{Gradient Compression}

\begin{figure}
    \centering
    \includegraphics[scale=0.2]{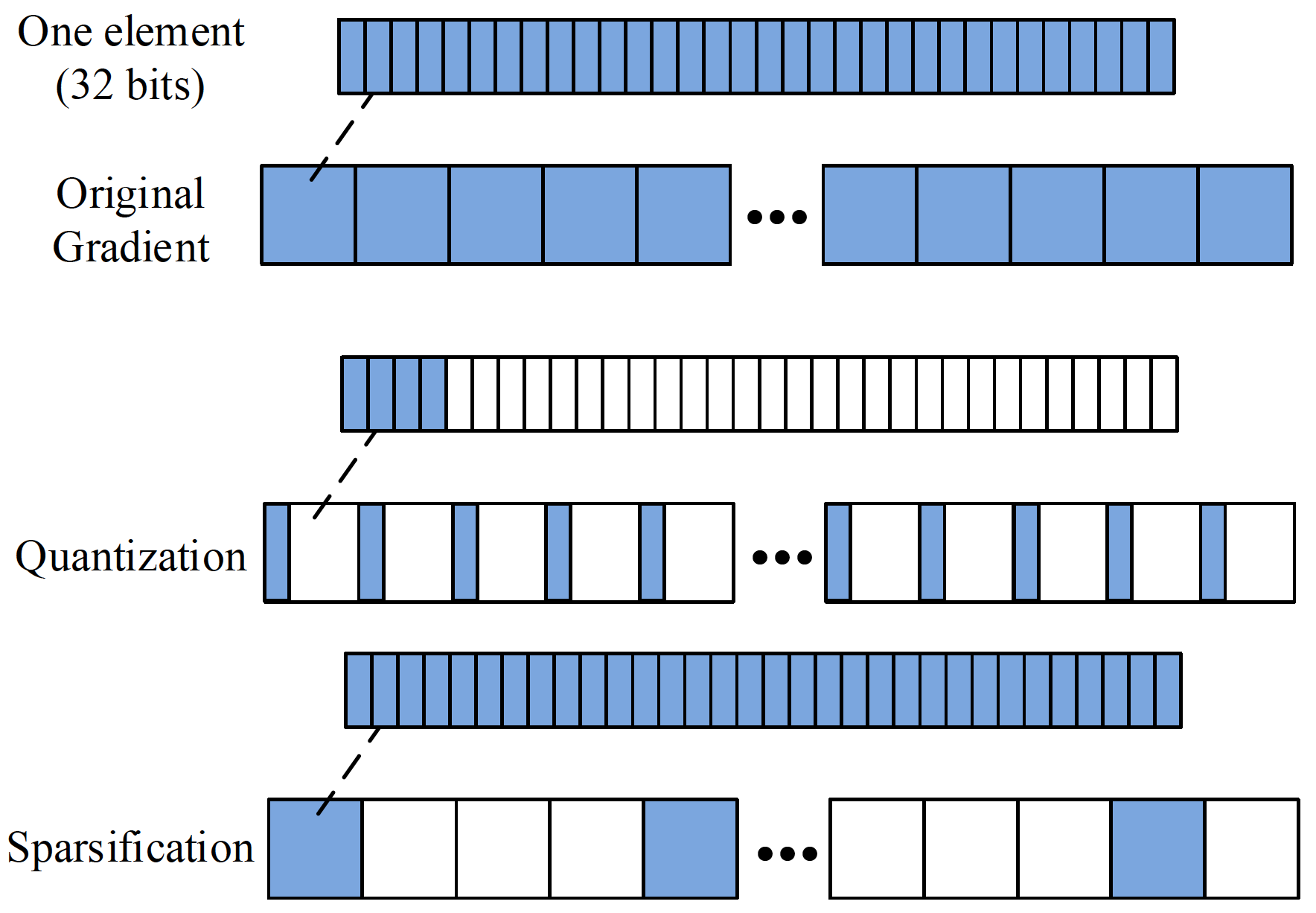}
    \caption{Comparison of Quantization and Sparsification, src:~\cite{survey_Communication_Efficient_Distributed_DL}}
    \label{fig: Comparison of Quantization and Sparsification}
\end{figure}

 \citet{deep_gradient_compression} find that 99.9\% of the gradient exchange in distributed SGD are redundant. One promising solution is gradient compression, e.g., through gradient quantization~\cite{QSGD,TernGrad,NUQSGD,1_bit_SGD,1_bit_Adam,1_bit_LAMB} and/or gradient sparsification~\cite{Sparse_Communication_for_Distributed_Gradient_Descent,deep_gradient_compression,adaptive_gradient_quantization,Scalable_distributed_DNN_training_using_commodity_GPU_cloud_computing}.
 Sparcification means transmitting only those gradients that are important (e.g., gradients with large absolute values), while quantization refers to using fewer bits to represent the original gradient. Their difference is described in Fig.~\ref{fig: Comparison of Quantization and Sparsification}.

\subsubsection{Gradient Quantization}
\begin{figure}[t]
\centering
\begin{minipage}[t]{0.55\textwidth}
\centering
\includegraphics[scale=0.4]{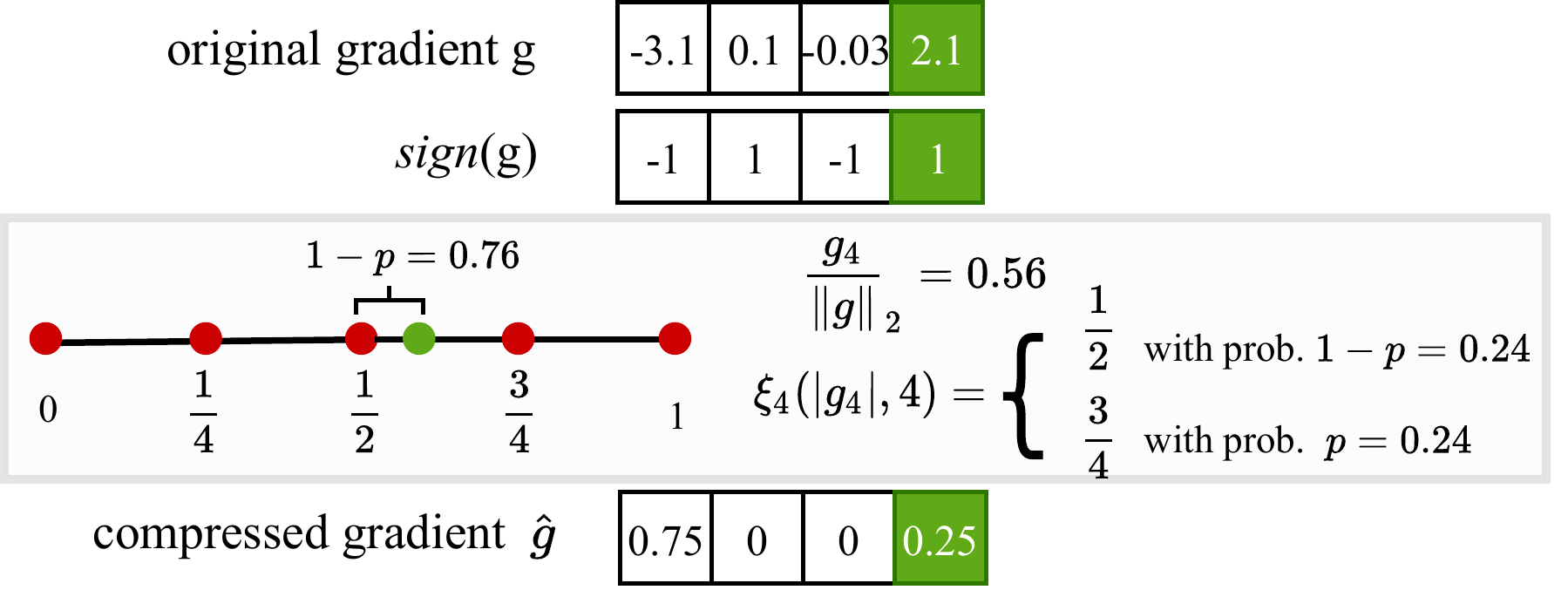}
\caption{QSGD Example with $s=4, l=3$}
\label{fig: QSGD}
\end{minipage}
\begin{minipage}[t]{0.35\textwidth}
\centering
\includegraphics[scale=0.4]{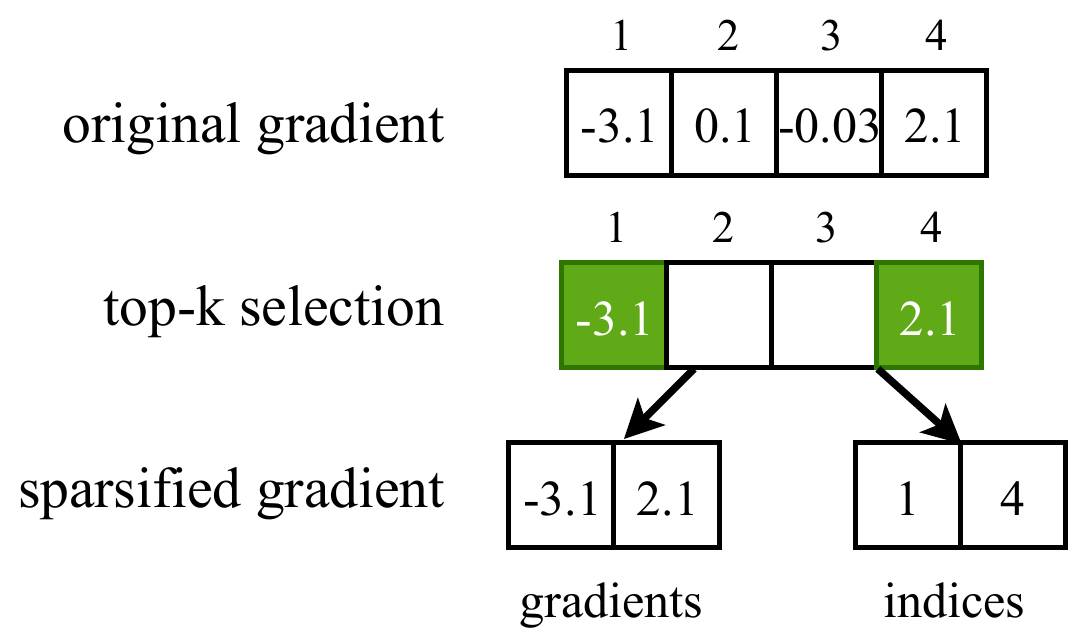}
\caption{Top-k Example}
\label{fig: topk}
\end{minipage}
\end{figure}

Quantizing the gradients to low-precision values can reduce the communication bandwidth. In full-precision data-parallel SGD, each processor broadcasts its locally computed stochastic gradient vector at every iteration, whereas in quantized data-parallel SGD, each processor quantizes its stochastic gradient before broadcasting.  Gradient quantization is usually done by mapping a continuous space of gradient values onto a discrete set. Take the classic Quantized Stochastic Gradient Descent (QSGD)~\cite{QSGD} as an example, the quantization function is denoted with $Q_s(v)$:
\begin{equation}
    Q_s(v_i)=\lVert v \rVert_2 \cdot sign(v_i) \cdot \xi_i(v,s)
\end{equation}
where $\xi_i(v,s)$ are independent random variables. Let $0\leq l < s$ be an integer such that $v_i/\lVert v\rVert_2 \in [l/s,(l+1)/s]$. That is, $[l/s, (l+1)/s]$ is the quantization interval corresponding to $v_i/\lVert v\rVert_2$. Then $\xi_i(v,s)$ is defined as follows
\begin{equation}
    \xi_i(v,s) = \left\{ 
    \begin{array}{cc}
         l/s& with\, probability \, 1-p(\frac{|v_i|}{\lVert v\rVert_2},s)\\
         (l+1)/s& otherwise
    \end{array}
    \right.
\end{equation}
Here, $p(a,s) = as - l$ for any $a\in [0,1]$. For gradients $v$, quantization is used to randomly convert gradient values in each dimension $v_i$ to some discrete value in a predetermined discrete set. After normalized by the Euclidean norm of the gradients ($|v_i|/\lVert v\rVert$), the value of each dimension will fall on a sub-interval $[0,1]$, and we approximate it to one of the endpoints of the sub-interval with a certain probability each time, so that the continuous value space of the original gradient value can be replaced by a set of finite discrete values. Here $\xi_i(v,s)$ is a binary random variable which guarantees that each value is quantized in a way which preserves the value in expectation,$E[\xi_i(v,s)]=|v_i|/\lVert v\rVert$, and introduce minimal variance. An instance of QSGD is provided in Fig.~\ref{fig: QSGD}.
% \begin{figure}[t]
%     \centering
%     \includegraphics[scale=0.25]{figs/generalized stochastic quantization.png}
%     \caption{An illustration of generalized stochastic quantization with 5 levels.}
%     \label{fig: generalized stochastic quantization}
% \end{figure}

This gradient quantization method greatly reduces the amount of communication required by a single node. Instead of passing $n$ 32-bit floating-point gradients, only one 32-bit floating-point gradient with one bit for gradient sign and $log(s)$ bits for $\xi_i(v,s)$ on each dimension are required. In addition, there is another method TernGrad~\cite{TernGrad} developed simultaneously with QSGD. Their underlying idea is essentially similar, where TernGrad can be viewed as a special case of QSGD when $l=1$. TernGrad randomly quantizates gradient $g_t$ to a ternary value vector with value of $\{-1,0,1\}$. Formally, with a random binary vector $b_t$, gradient is ternarized as
\begin{equation}
\begin{split}
    \Tilde{g}_t&=ternarize(g_t)=s_t\cdot sign(g_t)\circ b_t\\
    s_t &\triangleq \lVert g_t\rVert_\infty \triangleq max(abs(g_t)).
\end{split}
\end{equation}
Here, $\circ$ is the Hadamard product. 
% Giving a $g_t$, each element of $b_t$ independently follows the Bernoulli distribution
% \begin{equation}
%     \left \{
%     \begin{array}{cc}
%          P(b_{tk}=1|g_t)=|g_{tk}|/s_t&  \\
%          P(b_{tk}=0|g_t)=1- |g_{tk}|/s_t& 
%     \end{array}
%     \right.
% \end{equation}
% where $g_{tk}$ and $b_{tk}$ is the $k$-th element of $g_t$ and $b_t$, respectively.
TernGrad also adopts techniques such as layer-wise ternarizing and gradient clipping to improve convergence.
% After the gradient is compressed, the variance of the gradient on each node becomes larger, which may make the parameter update inaccurate, thus hindering the convergence of the training model or reducing the prediction accuracy. Therefore, we had better be able to control this factor, which is what the adaptive gradient quantization method tries to solve. Two adaptive gradient quantization methods, ALQ and AMQ, are proposed in this paper. They all belong to the adaptive version of QSGD (so they can also be called AQSGD). The basic idea is to reduce the variance between gradients on a single working node to speed up training by learning and adjusting the parameters of gradient compression in real time. 
\citet{NUQSGD} propose nonuniform quantization levels (NUQSGD) and demonstrate superior empirical results compared to QSGD. \citet{Natural_Compression_for_Distributed_Deep_Learning} propose  natural compression and natural dithering, where the latter is a special case of logarithmic quantization.

Unlike QSGD and its variants which use stochastic rounding which has an unbiased gradient expectation, there are  methods adopt biased ones. 
Methods performing updates only based on the sign of each coordinate of the gradient have gained popularity for training DL models~\cite{1_bit_SGD,1_bit_SGD_convergence,EF_Sign_SGD,1_bit_Adam,1_bit_LAMB}. \citet{1_bit_SGD} proposed  SignSGD, a.k.a. 1-bit SGD, to quantize the gradients aggressively to one bit per value. In this scheme, gradient updates greater than or equal to zero are encoded using the value 1, and those less than zero with the value 0. The reconstruction values are chosen to be the means of the non-negative and negative updates, respectively, in order to minimize the square quantization error. This is done column-wise over the weight matrix. In each data exchange, the two reconstruction values are transmitted along with their respective quantized column.
\citet{1_bit_SGD_convergence} later provided convergence guarantees for a variant of SignSGD. \citet{EF_Sign_SGD} proposed EF-SignSGD, which is an improved version of SignSGD.
% \RestyleAlgo{ruled}
% \SetKwInput{KwInit}{Init}
% \begin{algorithm}[t]
% \caption{EF-SIGNSGD (SIGNSGD with Error-Feedback)}\label{alg: EF-SIGNSGD}
% \KwIn{LR $\gamma$, initial parameters $x_0 \in \mathbb{R}^d$, initial error $e_0=0$}
% \For{$t=0,1,\cdots,T-1$}{
% $g_t :=$stochasticGradient($x_t$)\Comment*[r]{calculate mini-batch stochastic gradient}
% $p_t:=\gamma g_t+e_t$\Comment*[r]{error correction}
% $\Delta_t:=(\lVert p_t\rVert_1/d)sign(p_t)$\Comment*[r]{compression}
% $x_{t+1}:=x_t-\Delta_t$\Comment*[r]{update parameters}
% $e_{t+1}:=p_t-\Delta_t$\Comment*[r]{update residual error}
% }
% \end{algorithm}
More recently, gradient compression with error compensation has been successfully applied to adaptive optimizer such as Adam (1-bit Adam ~\cite{1_bit_Adam}) and LAMB (1-bit LAMB ~\cite{1_bit_LAMB}), further scaling up training algorithms in the distributed setting.

To recap, while the analyses of gradient quatization have largely been restricted to unbiased compression schemes ~\cite{QSGD,NUQSGD,TernGrad, Natural_Compression_for_Distributed_Deep_Learning}, biased schemes which perform extreme compression practically perform much better often without any loss in convergence or accuracy~\cite{1_bit_SGD,EF_Sign_SGD,Scalable_distributed_DNN_training_using_commodity_GPU_cloud_computing,deep_gradient_compression,1_bit_Adam,1_bit_LAMB}.  
\subsubsection{Gradient Sparsification}
Gradient sparsification is an orthogonal approach to quatization methods, which reduces the communication bandwidth by sending only the important gradients. 
Since zeroing small gradients damages convergence, small gradients are accumulated over time locally until they become large enough to be transmitted. Thus, we send the large gradients immediately but eventually send all of the gradients.
\citet{Scalable_distributed_DNN_training_using_commodity_GPU_cloud_computing} proposed threshold quantization by considering only gradient elements whose absolute values exceed a threshold. 
A fixed threshold $\tau$ is chosen in advance. Gradient updates greater than $\tau$ are encoded with the value 1, and those less than $-\tau$ with the value of 0. Updates of magnitude less than $\tau$ are not sent at all, reducing the volume of data sent.  The reconstructed value is $\tau$ and $-\tau$ respectively, and error feedback is used as normal. 
However, the threshold is hard to choose in practice and, moreover, it can change over time during optimization. As a resolve, Top-k sparsification selects the top-k gradients in terms of absolute values at each iteration~\cite{StichCJ18, AlistarhH0KKR18} (see Fig.~\ref{fig: topk} for an example). \citet{adaptive_gradient_quantization} choose an adaptive threshold so as to keep a constant proportion of gradients each iteration.
\citet{Sparse_Communication_for_Distributed_Gradient_Descent} sparsify gradient updates by removing the R\%
smallest gradients by absolute value, dubbing this Gradient Dropping.
This approach is slightly different from ~\cite{adaptive_gradient_quantization} as it uses a single threshold based on absolute value, instead of dropping the positive and negative gradients separately. Concurrently, \citet{AdaComp} localize selection of gradient residues and automatically tunes the compression rate depending on local activity. \citet{deep_gradient_compression} further push the compression ratio by employing momentum correction, local gradient clipping, momentum factor masking, warm-up training on top of the gradient sparsification while maintaining model performance. Table~\ref{tab: Gradient Compression} summarises the gradient quantization and sparsification methods.
\begin{table}[t]
    \centering
    \begin{tabular}{ccc}
    \hline
    Method&Taxonomy&Reference\\
    \hline
         \multirow{2}{*}{Quant.} &  Unbiased& QSGD~\cite{QSGD}, NQSGD~\cite{NUQSGD}, TernGrad~\cite{TernGrad},  Natural~\cite{Natural_Compression_for_Distributed_Deep_Learning}\\
         ~ &  Biased& 
         \makecell*[c]{{1-bit SGD ~\cite{1_bit_SGD} /Adam~\cite{1_bit_Adam}/LAMB~\cite{1_bit_LAMB}}\\{EF-SignSGD~\cite{EF_Sign_SGD}}} \\
         \multirow{2}{*}{Spars.}
         & Random& Random-k~\cite{WangniWLZ18}\\
         ~.& Deterministic& \makecell*[c]{
         Fixed threshold~\cite{Scalable_distributed_DNN_training_using_commodity_GPU_cloud_computing} \\
         Top-K~\cite{StichCJ18, AlistarhH0KKR18},
         Adaptive threshold~\cite{adaptive_gradient_quantization, Sparse_Communication_for_Distributed_Gradient_Descent, AdaComp, deep_gradient_compression}}  \\
         \hline
    \end{tabular}
    \caption{Summary of Gradient Compression Methods}
    \label{tab: Gradient Compression}
\end{table}
% \RestyleAlgo{ruled}
% \SetKwInput{KwInit}{Init}
% \begin{algorithm}[t]
% \caption{Gradient Sparcification on node $k$}\label{alg: gradient_sparcification}
% \KwIn{Dataset $X$,minibatch size $b$ per node, the number of nodes $N$, optimization function \textit{SGD}, init parameters $w={w[0],\cdots,w[M]}$}
% \For{$t=0,1,\cdots$}{
% $G_t^k\leftarrow0$\;
% \For{$i=1,\cdots,B$}{
% Sample data $x$ from $X$\;
% $G_t^k \leftarrow G_t^k+\frac{1}{Nb}\nabla f(x;w_t)$
% }
% \For{$j=0,1,\cdots,M$}{
% Select threshold:$thr\leftarrow s\%$ of $| G_t^k[j]|$\;
% $Mask \leftarrow | G_t^k[j]| > thr$\;
% $\widetilde{G}_t^k[j]\leftarrow {G}_t^k[j] \odot Mask$\;
% ${G}_t^k[j]\leftarrow {G}_t^k[j] \odot \neg Mask$\;
% }
% All-Reduce $G_t^k: G_t \leftarrow \sum_{k=1}^N encode(\widetilde{G_t^k})$\;
% $w_{t+1}\leftarrow \textit{SGD}(w_t,G_t)$
% }
% \end{algorithm}

\subsection{Reducing Communication Frequency}
A parallel line of work reduces the communication cost by reducing the frequency of communication. For instance, local SGD saves the communication cost by allowing each worker to perform more than one batch update on local data and exchange the updated weights rather than the gradients among workers.

We consider a distributed SGD framework with $K$ worker nodes where all workers communicate with others via a central server or via direct inter-worker communication. In local SGD, each worker $k\in [K]$ performs $H$ sequential mini-batch SGD updates locally, and then the local models are synchronized by averaging weights among workers. Thus, the overall update rule at the $k$-th worker is given by
\begin{equation}
    \begin{split}
    w_{(t)+h+1}^k &:= w_{(t)+h}^k -\eta_{(t)}[\frac{1}{B_{loc}}\sum_{i\in {I^k_{(t)+h}}}\nabla f_i(w_{(t)+h}^k)]\\        
    w_{(t+1)}^k&:=\frac{1}{K}\sum_{k=1}^K w_{(t)+H}^k
    \end{split}
\end{equation}
where $w_{(t)+h}^k$ denotes the local model on worker $k$ with batch size $B_{loc}$ after $t$ global synchronization and $h$ local SGD updates.
Mini-batch SGD is a special case of local SGD, with $H=1$, that is, the local models are synchronized after every iteration. 
% The full algorithm of local SGD is given in Algorithm~\ref{alg: Local SGD}.
 The convergence results for convex and non-convex objectives are provided in ~\cite{Local_SGD_Converges_Fast_and_Communicates_Little,local_sgd_non_convex_convergence}.
However, while local updates reduce the communication frequency by performing global synchronization periodically instead of at per iteration, the discrepancies between local models can result in an inferior error-convergence. A larger value of $H$ (i.e., the number of sequential local SGD updates), which means less frequent averaging, saves communication delay and reduces the run-time per iteration. But on the other hand, a larger $H$ leads to slower convergence w.r.t. the number of iterations. The trade-off in between still need more exploration.
% \RestyleAlgo{ruled}
% \SetKwInput{KwInit}{Init}
% \begin{algorithm}[t]
% \caption{Local SGD}\label{alg: Local SGD}
% \KwInit{$x_0^k = x_0$ for worker $k \in [K]$}
% \For{$t=0,1,\cdots, T-1$}{
% parallel \For{$k \in [K]$}{
% Sample $i_t^k$ uniformly in $[n]$\;
% \If{$t+1\in I_T$}{$x_{t+1}^k\leftarrow \frac{1}{K}\sum_{k=1}^K (x_t^k-\eta_t\nabla f_{i_t^k}(x_t^k))$\Comment*[r]{global synchronization}}
% \Else{$x_{t+1}^k\leftarrow x_t^k - \eta_t\nabla f_{i_t^k}(x_t^k)$ \Comment*[r]{local update}}
% }
% }
% \end{algorithm}

In addtion to being communication efficient, \citet{Use_Local_SGD} find local SGD also exhibits good generalization behaviour. They argue that local SGD is a way to inject and control stochastic noise to the whole training procedure, and thus proposed post-local SGD as large batch training alternative for better generalization.

\section{Memory}\label{sec: Memory}
Larger models usually require more computation and memory resources to train. The amount of memory required to train these models can be several orders of magnitude larger than the amount of memory available on a single GPU. In this section, we will see how some popular techniques successfully reduce the memory requirements of training NNs without compromising model performance. Section~\ref{subsec: Mix-Precision Training} introduces how mix-precision training~\cite{mix_precision_training} lowers the burden on memory using fewer bits to preserve the weights and gradients during training. Section~\ref{subsec: Memory Efficient Adaptive Optimization} introduces two memory-efficient adaptive optimizers, Adafactor~\cite{Adafactor} and SM3~\cite{SM3}. And as orthogonal to the above methods, ZeRO~\cite{ZeRO} do not change the model optimization method or affect model convergence, but instead reduces the memory cost by removing the redundancy in data-parallel (Section~\ref{subsec: zero}). 

% \begin{table}[htbp]
%     \centering
%   \begin{tabular}{ccc}
%   \hline
%   \multicolumn{2}{c}{method}&Auxilary Memory\\
%   \hline
%   Gradient&SGD&0\\\hline
%   Momentum&\makecell*[c]{heavy ball\\Nestrove} &\multirow{5}{*}{N}\\
%   Adaptive&\makecell*[c]{Adagrad\\Adadelta\\RMSprop}&~\\\hline
%   Combined&\makecell*[c]{Adam\\Nadam\\AMSGrad}&2N\\
%   \hline
% \end{tabular}
%     \caption{Gradient-Based Optimization}
%     \label{tab: Gradient-Based Optimization}
% \end{table}

\subsection{Mix-Precision Training}\label{subsec: Mix-Precision Training}

Modern DL training systems use single-precision (FP32) format, which takes 32 bits of memory. However, lower-precision (FP16) takes 16 bits of memory instead. Modern accelerators like Google TPUs and NVIDIA GPUs can run operations faster in the FP16 format, as they have specialized hardware to run 16-bit computations and 16-bit dtypes can be read from memory faster. These lower-precision provides numerous benefits. First, they require less memory, enabling the training and deployment of larger NNs. Second, they lowers the burden on memory since fewer bits are required to preserve the same number of values than the FP32 format, thereby speeding up data transfer operations. Third, they speed up the mathematical computation since low-precision calculation is less time-consuming, especially on GPUs with Tensor Core support for that precision. 
However, low precision training also introduces a trade-off of the number of bits used versus the statistical accuracy: the fewer bits used, the lower accuracy. 
\begin{figure}[t]
    \centering
    \includegraphics[scale=0.4]{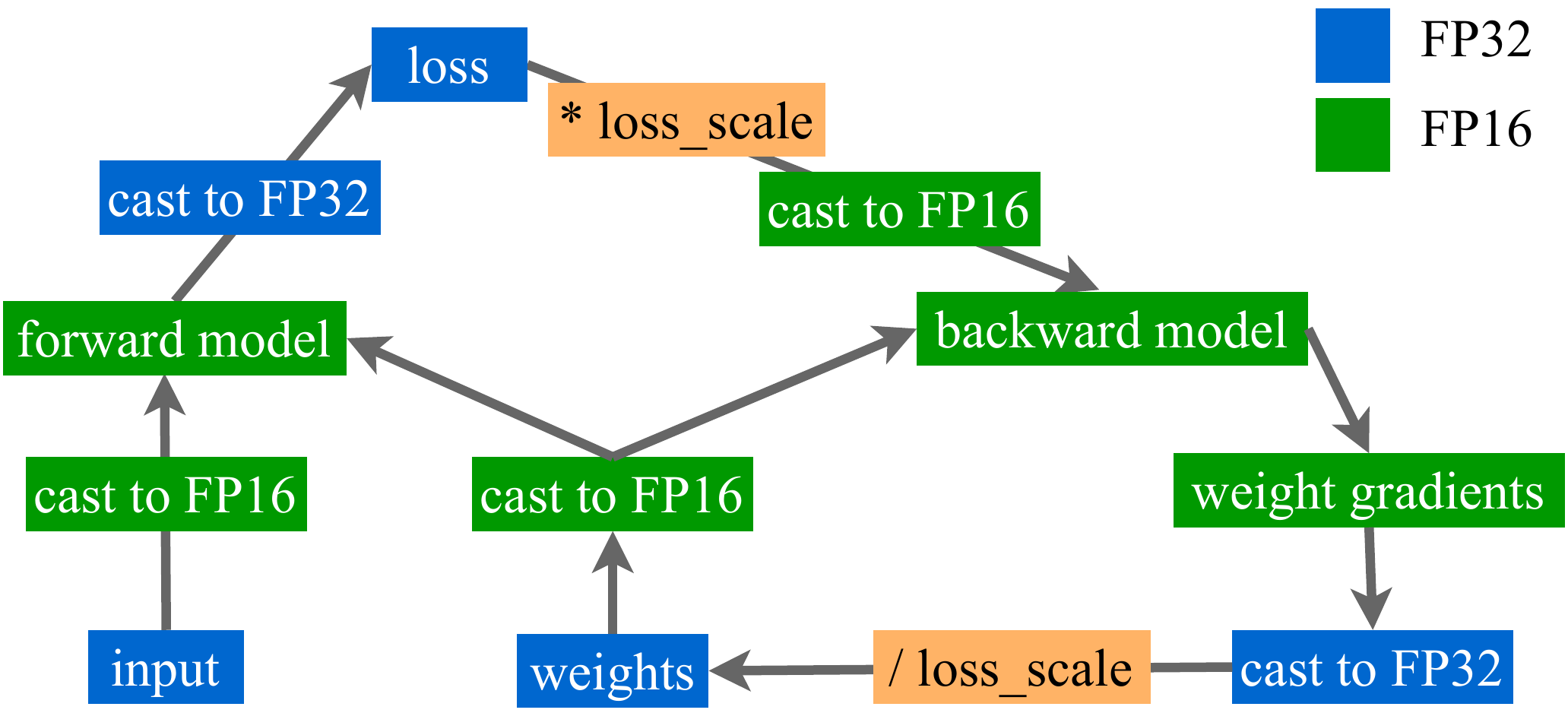}
    \caption{Workflow of Mix Precision Training}
    \label{fig: mix_precision}
\end{figure}
Mixed precision training is a very simple and practical technique, jointly published by Baidu and Google in \cite{mix_precision_training}, which almost halves the memory footprint while maintaining the model accuracy. This is achieved by identifying the steps that require full precision and using FP32 for only those steps while using FP16 everywhere else.   
We explain the workflow of mix precision training in Figure~\ref{fig: mix_precision}. Due to the differences in representable ranges, simply converting the model to FP16 can cause gradient underflow or overflow problems. We can avoid these challenges with the following four steps: (1) Conversion into FP16. In step one, we cast the inputs from FP32 to FP16 for compute intensive operations. (2) Use FP32 to compute the loss. Because FP16 might cause underflow or overflow issues, we do loss calculation in FP32 in the backward pass and cast gradients back to FP16, which means that the weights
and the gradients are still in FP16. (3) FP32 master weights. In the backward pass, gradients are small compared to the parameters. If we try to update our parameters with the gradients that are much smaller than the parameters, then we might lose those parameter updates. To compensate we will maintain the master copy of weights in FP32. This means that, to the end of the backward pass, FP16 gradients will be cast into FP32 and thereby applied to FP32 weights. In the forward pass, we will cast the weights into FP16 so that the gradient computations remain in FP16. So effectively we have a master copy of all the parameters which are weights and biases stored in FP32 but all the computational operations will see the casted version which is in FP16. (4) Loss (Gradient) scaling. The last step is to do loss scaling to avoid a gradient underflow problem. Before computing gradients from the FP32 loss, we scale a loss by multiplying it with a loss scale factor. By doing so, gradients are pushed to larger values and we can safely represent them in FP16. Later when updating the weights we
can re-scale the gradients by dividing them with the same loss scale factor.

Their article~\cite{mix_precision_training} is not the first to propose the use of lower precision for training, but its influence is far-reaching, and many current programs are designed based on this work. \citet{DBLP:journals/corr/abs-1807-11205} apply mixed-precision training to large-batch strategies such as LARS. Using LARS with mixed-precision training,  ResNet-50 with the mini-batch size of 64K, could maintain the top-1 accuracy as 76.2\%.

\subsection{Memory Efficient Adaptive Optimization}\label{subsec: Memory Efficient Adaptive Optimization}
Some stochastic optimization methods (e.g., RMSProp, Adam~\cite{kingma2017adam}, Adadelta~\cite{Adadelta}), keep first and second moment estimates of the per-parameter gradients to scale the gradients which triples the required memory. As models continue to grow, the memory overhead will pose more limitation on the quality of the trained model. Motivated by these challenges, memory efficient adaptive optimization methods are proposed to retain the benefits of standard per-parameter adaptivity while significantly reduce memory overhead. For instance, Adafactor~\cite{Adafactor} was proposed as a way to reduce the memory costs of AdaGrad~\cite{AdaGrad}, primarily for training large language models, and SM3~\cite{SM3} saves memory by sharing moments of similar magnitude.
% Therefore, many memory optimizations have been proposed in order to fit up to billions of parameters. For example, the Adam optimizer keeps first and second moment estimates of the per-parameter gradients which triples the required memory. ~\cite{MoE} sets $\beta_1=0$ to eliminate the need for keeping a first-moment estimator, and replace the second-moment estimators with a factored approximation to reduce the required memory. Similarly, ~\cite{Adafactor} maintains only the per-row and per-column sums of the exponential moving averages of squared past gradients, and uses them to estimate the per-parameter second moments.

\subsubsection{Adafactor}
Adafactor ~\cite{Adafactor} is a space-efficient adaptive optimization which achieves a drastic reduction in auxiliary memory usage without hurting the performance (compared to that obtained using full accumulators). One of the key contribution is the use of factored second momentum estimation.

Consider a matrix-shaped parameter subset $X$ with second moment estimate $V$. They want to identify a low-rank representation of $V$ as a product of two factors $R$ and $S$, i.e., $V\approx RS$ which is compatible with exponential moving averaging. This would allow us to store just the low-rank factors across iteration, cutting down a memory usage. More formally, if factorization $F: V\mapsto (R,S)$, we want $F(\eta V_{t-1}+(1-\eta)G_t^2)=\eta F(V_{t-1})+(1-\eta)F(G_t^2)$. In particular, by using techniques from non-negative matrix factorization using I-divergence, the low-rank approximation can be converted into following optimization problem:
\begin{equation}
\begin{split}
    \mathop{minimize}_{R \in \mathbb{R}^{n\times k}, S \in \mathbb{R}^{k\times m}} \sum_{i=1}^n\sum_{j=1}^{m}d(V_{ij},[RS]_{ij})\\
    subject \,to\quad R_{ij}>0,  S_{ij}>0.
\end{split}
\end{equation}
In particular, for the case of rank one factors, i.e., $k=1$, the solution set of the optimization problem can be characterized as the set of all pairs $(R,S)$, whose product is equal to the expression below:
\begin{equation}
    \{(R,S):RS=\underbrace{V1_m}_{row\atop sums}\underbrace{1_n^TV}_{colum\atop sums}/\underbrace{1_n^TV1_m}_{sum\,of \atop all\,entries}\}.
\end{equation}
The right hand side can be broken down into the vector of row sums and column sums, and the denominator is the sum of all entries. In addition to the factored second moment estimation, other key changes in Adafactor include $\eta_2$ varies with time, update cliping, relative step size and no momentum. 
% \begin{equation}
%     \begin{split}
%         R_i = \sum_{j=1}^{m}V_{ij},\quad S_j = \frac{\sum_{i=1}^{n}V_{ij}}{\sum_{i=1}^{n}\sum_{j=1}^{m}V_{ij}}
%     \end{split}
% \end{equation}

\subsubsection{SM3 Algorithm}
 Adaptive gradient methods, such as AdaGrad~\cite{AdaGrad}, have proved to be particularly useful in training sparse models. Crucially, however, Adagrad must maintain auxiliary sequence of accumulators (i.e., the diagonal preconditioner) $H_t$ (also in Eq.[\ref{eq: adagrad}]) :
 \begin{equation}
    H_{t,ii} = \sum_{s\leq t} g_{s,ii}^2
\end{equation}
and thus needs $\Omega(n)$ additional space $n$ is the number of parameters. SM3 ~\cite{SM3} provides a memory-efficient methods with comparable convergence characteristics which refrains from maintaining the full vectors.
SM3 is short for save memory by sharing moments of similar magnitude. This is because they observe that the diagonal preconditioners $H_t$ accumulated by AdaGrad are actually similar in rows and columns, and by sharing moments cross rows and columns, the memory requirements therefore drop from $\Theta(mn)$ to merely $\Theta(m + n)$.
\begin{equation}
    \begin{split}
        \widehat{H_{t+1},ij} &= min(R_{t,i}, C_{t,j})+g_{t+1,ij}^2\\
        R_{t,i} &= \mathop{max}\limits_j(\widehat{H_{t+1},ij})\\
        C_{t,j} &= \mathop{max}\limits_i(\widehat{H_{t+1},ij})
    \end{split}
\end{equation}
SM3 can be viewed as a diagonal version of Shampoo (see Table~\ref{tab: Memory}). We refer readers about the implementation details to ~\cite{SM3}.
% More formally, the algorithm employs a cover of the parameters: a collection of k nonempty sets $\{S_r\}_{r=1}^k$, such that $S_r \subseteq [d]$ and $\cup_r S_r = [d]$.

\subsection{ZeRO}\label{subsec: zero}
\begin{figure}[t]
    \centering
    \includegraphics[width=0.75\linewidth]{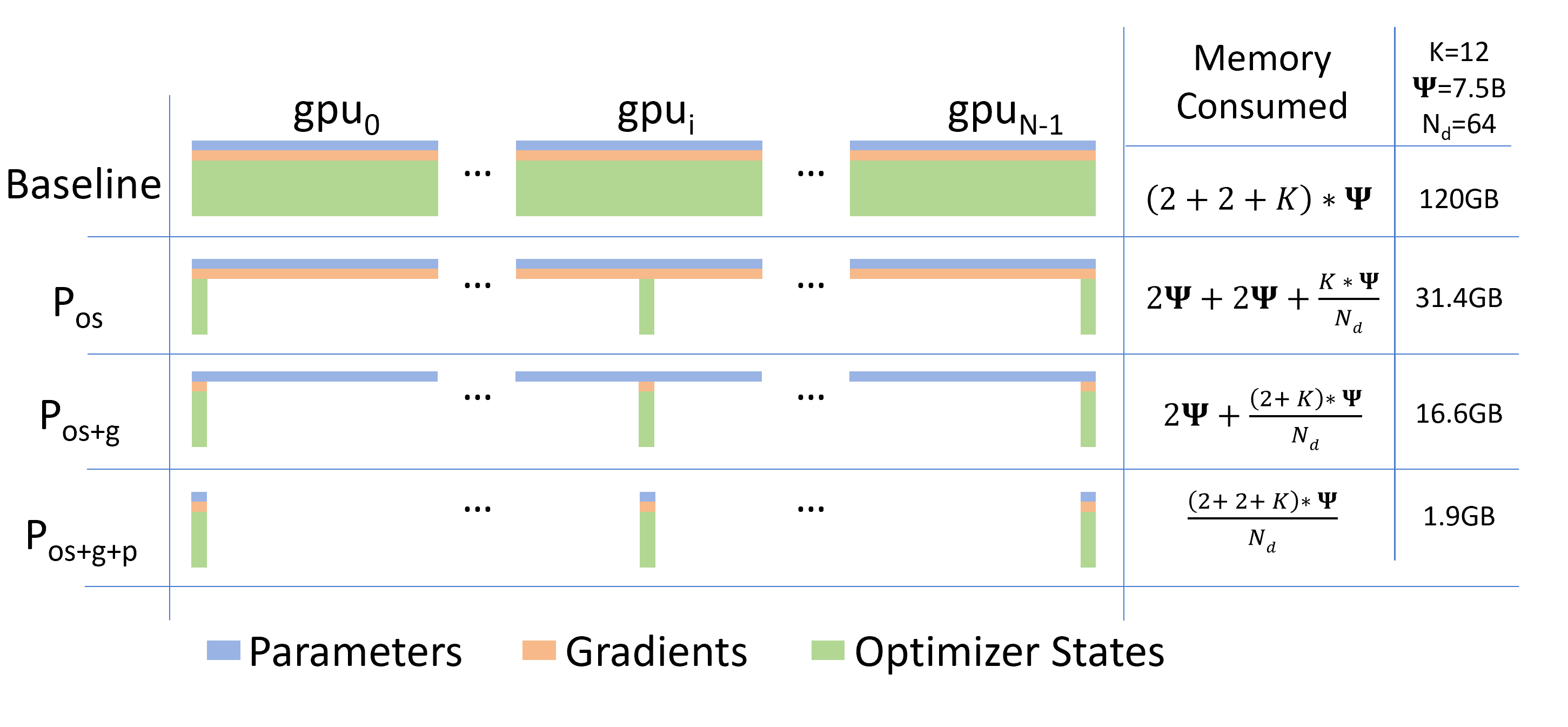}
    \caption{Comparing the per-device memory consumption of model states, with three stages of ZeRO-DP optimizations. Src:\cite{ZeRO}.}
    \label{fig: zero}
\end{figure}
The Zero Redundancy Optimizer~\cite{ZeRO} (abbreviated as ZeRO) is a novel memory optimization technology for large-scale distributed DL. Contrary to Adafactor ~\cite{Adafactor} and SM3~\cite{SM3} which reduce memory consumption of adaptive optimization methods by maintaining coarser-grained statistics of model parameters and gradients, ZeRO do not change the model optimization method or affect the model convergence. 

We show how ZeRO works in Fig.[\ref{fig: zero}]. The first row shows the memory map while training a model in data parallel. The first row shown in blue represents the memory consumed by the parameters; the second row in orange shows the memory consumed by the gradients; and the big chunk in green shows the memory consumed by the optimizer states (e.g., this could be momentum and variance for Adam). So the key thing is that these optimizer states gradients and parameters (which we collectively call the model states) are replicated across all the different GPUs in distributed data parallel training. The way ZeRO works is by removing this redundancy across these GPUs.  Since there are three different types of model states, there are three different phases of ZeRO, each of them removing the redundancy for one of these states  by simply partitioning these model states across GPUs instead of replicating them.

In addition to reducing the memory from model states, ZeRO also has a bunch of more optimizations that allows reduction in memory from other components (see Table~\ref{tab: ZeRO}). For example, just like how model states are replicated across multiple GPUs in data parallel training, ZeRO removes the redundancy in activation memory by partitioning the activations across these model parallel GPUs.  We can also offload these activation memories to CPU if we don't have enough memory to train extremely large models.  The next optimization that ZeRO can do is to convert fragmented memory to defragmented memory on the fly during training.  During training if the memory is fragmented, we might still run out of memory even though there might be enough fragmented memory that can satisfy the request if they were contiguous.  In ZeRO the memory defragmentation will on the fly defragment these memory fragments so that all the memory is contiguous and you are able to satisfy these larger memory requests. So with all these different memory optimizations, ZeRO is able to train models with up to 200 billion parameters up to 10 times faster than the SOTA.

% ZeRO can train DL models with 100 billion parameters on the current generation of GPU clusters at three to five times the throughput of the current best system. It also presents a clear path to training models with trillions of parameters, demonstrating an unprecedented leap in DL system technology. 
\begin{table}[t]
    \centering
    \begin{tabular}{cc}
    \hline
         Memory Consumption&Optimization  \\\hline
         Model State Memory&Partition optimizer state, gradient, and parameters\\ 
         Activation Memory&Partition activations; Offload to GPU\\
         Fragmented Memory&Proactively manage memory w.r.t tensor lifetime\\
         \hline
    \end{tabular}
    \caption{Different Memory Optimizations in ZeRO}
    \label{tab: ZeRO}
\end{table}

\section{Conclusion}\label{sec: Conclusion}
\label{sec:conclusion}
Given larger datasets and bigger models consistently yielding significant improvements in accuracy, large-scale deep learning has become an inevitable trend. As datasets increase in size and DNNs in complexity, the computational intensity, communication cost and memory demands of deep learning increase proportionally. Considerable efforts have been devoted to accelerating the training speed. In this article, we give an overview of large-scale deep learning optimization. The goal in general is two-fold: model accuracy and model efficiency. As for the model accuracy, we investigate algorithms that are most commonly used for optimizing, spanning from the gradient descent variants to the (large-batch) adaptive methods, and from first-order to second-order methods. Further, we elaborate the debatable topic of generalization gap arises in large-batch training. As for the model efficiency, we summarise the SOTA techniques in addressing the expensive cost of communication overhead and memory footprint. 
% Model accuracy and model efficiency is not far from independent. Without  an acceptable budget of communication and memory, large-scale training can not be implemented in practice, let alone achieving desired model accuracy. 
We hope this article can provide a clean sketch for those who are interested in training large-scale training.

\bibliographystyle{ACM-Reference-Format}
\bibliography{sample-acmsmall}


\begin{thebibliography}{126}

%%% the Web address.

\ifx \showCODEN    \undefined \def \showCODEN     #1{\unskip}     \fi
\ifx \showDOI      \undefined \def \showDOI       #1{#1}\fi
\ifx \showISBNx    \undefined \def \showISBNx     #1{\unskip}     \fi
\ifx \showISBNxiii \undefined \def \showISBNxiii  #1{\unskip}     \fi
\ifx \showISSN     \undefined \def \showISSN      #1{\unskip}     \fi
\ifx \showLCCN     \undefined \def \showLCCN      #1{\unskip}     \fi
\ifx \shownote     \undefined \def \shownote      #1{#1}          \fi
\ifx \showarticletitle \undefined \def \showarticletitle #1{#1}   \fi
\ifx \showURL      \undefined \def \showURL       {\relax}        \fi
% invisible to TeX
\providecommand\bibfield[2]{#2}
\providecommand\bibinfo[2]{#2}
\providecommand\natexlab[1]{#1}
\providecommand\showeprint[2][]{arXiv:#2}

\bibitem[\protect\citeauthoryear{Abadi, Barham, Chen, Chen, Davis, Dean, Devin,
  Ghemawat, Irving, Isard, Kudlur, Levenberg, Monga, Moore, Murray, Steiner,
  Tucker, Vasudevan, Warden, Wicke, Yu, and Zhang}{Abadi et~al\mbox{.}}{2016}]%
        {tensorflow}
\bibfield{author}{\bibinfo{person}{Mart{\'i}n Abadi}, \bibinfo{person}{Paul
  Barham}, \bibinfo{person}{Jianmin Chen}, \bibinfo{person}{Z. Chen},
  \bibinfo{person}{Andy Davis}, \bibinfo{person}{Jeffrey Dean},
  \bibinfo{person}{Matthieu Devin}, \bibinfo{person}{Sanjay Ghemawat},
  \bibinfo{person}{Geoffrey Irving}, \bibinfo{person}{Michael Isard},
  \bibinfo{person}{Manjunath Kudlur}, \bibinfo{person}{Josh Levenberg},
  \bibinfo{person}{Rajat Monga}, \bibinfo{person}{Sherry Moore},
  \bibinfo{person}{Derek~Gordon Murray}, \bibinfo{person}{Benoit Steiner},
  \bibinfo{person}{Paul~A. Tucker}, \bibinfo{person}{Vijay Vasudevan},
  \bibinfo{person}{Pete Warden}, \bibinfo{person}{Martin Wicke},
  \bibinfo{person}{Yuan Yu}, {and} \bibinfo{person}{Xiaoqian Zhang}.}
  \bibinfo{year}{2016}\natexlab{}.
\newblock \showarticletitle{TensorFlow: A system for large-scale machine
  learning}. In \bibinfo{booktitle}{\emph{OSDI}}.
\newblock


\bibitem[\protect\citeauthoryear{Agarwal, Bullins, and Hazan}{Agarwal
  et~al\mbox{.}}{2016a}]%
        {agarwal2016second}
\bibfield{author}{\bibinfo{person}{Naman Agarwal}, \bibinfo{person}{Brian
  Bullins}, {and} \bibinfo{person}{Elad Hazan}.}
  \bibinfo{year}{2016}\natexlab{a}.
\newblock \showarticletitle{Second-order stochastic optimization in linear
  time}.
\newblock \bibinfo{journal}{\emph{stat}}  \bibinfo{volume}{1050}
  (\bibinfo{year}{2016}), \bibinfo{pages}{15}.
\newblock


\bibitem[\protect\citeauthoryear{Agarwal, Zhu, Bullins, Hazan, and Ma}{Agarwal
  et~al\mbox{.}}{2016b}]%
  {Finding_Approximate_Local_Minima_for_Nonconvex_Optimization_in_Linear_Time}
\bibfield{author}{\bibinfo{person}{Naman Agarwal},
  \bibinfo{person}{Zeyuan~Allen Zhu}, \bibinfo{person}{Brian Bullins},
  \bibinfo{person}{Elad Hazan}, {and} \bibinfo{person}{Tengyu Ma}.}
  \bibinfo{year}{2016}\natexlab{b}.
\newblock \showarticletitle{Finding Approximate Local Minima for Nonconvex
  Optimization in Linear Time}.
\newblock \bibinfo{journal}{\emph{CoRR}}  \bibinfo{volume}{abs/1611.01146}
  (\bibinfo{year}{2016}).
\newblock


\bibitem[\protect\citeauthoryear{Aji and Heafield}{Aji and Heafield}{2017}]%
        {Sparse_Communication_for_Distributed_Gradient_Descent}
\bibfield{author}{\bibinfo{person}{Alham~Fikri Aji} {and}
  \bibinfo{person}{Kenneth Heafield}.} \bibinfo{year}{2017}\natexlab{}.
\newblock \showarticletitle{Sparse Communication for Distributed Gradient
  Descent}. In \bibinfo{booktitle}{\emph{Proceedings of the 2017 Conference on
  Empirical Methods in Natural Language Processing, {EMNLP} 2017, Copenhagen,
  Denmark, September 9-11, 2017}}. \bibinfo{pages}{440--445}.
\newblock


\bibitem[\protect\citeauthoryear{Alistarh, Grubic, Li, Tomioka, and
  Vojnovic}{Alistarh et~al\mbox{.}}{2017}]%
        {QSGD}
\bibfield{author}{\bibinfo{person}{Dan Alistarh}, \bibinfo{person}{Demjan
  Grubic}, \bibinfo{person}{Jerry Li}, \bibinfo{person}{Ryota Tomioka}, {and}
  \bibinfo{person}{Milan Vojnovic}.} \bibinfo{year}{2017}\natexlab{}.
\newblock \showarticletitle{QSGD: Communication-Efficient SGD via Gradient
  Quantization and Encoding}. In \bibinfo{booktitle}{\emph{NIPS}}.
  \bibinfo{pages}{1709--1720}.
\newblock


\bibitem[\protect\citeauthoryear{Alistarh, Hoefler, Johansson, Konstantinov,
  Khirirat, and Renggli}{Alistarh et~al\mbox{.}}{2018}]%
        {AlistarhH0KKR18}
\bibfield{author}{\bibinfo{person}{Dan Alistarh}, \bibinfo{person}{Torsten
  Hoefler}, \bibinfo{person}{Mikael Johansson}, \bibinfo{person}{Nikola
  Konstantinov}, \bibinfo{person}{Sarit Khirirat}, {and}
  \bibinfo{person}{C{\'{e}}dric Renggli}.} \bibinfo{year}{2018}\natexlab{}.
\newblock \showarticletitle{The Convergence of Sparsified Gradient Methods}. In
  \bibinfo{booktitle}{\emph{Advances in Neural Information Processing Systems
  31: Annual Conference on Neural Information Processing Systems 2018, NeurIPS
  2018, December 3-8, 2018, Montr{\'{e}}al, Canada}}.
  \bibinfo{pages}{5977--5987}.
\newblock


\bibitem[\protect\citeauthoryear{Amari}{Amari}{1998}]%
        {natural_gradient}
\bibfield{author}{\bibinfo{person}{Shun{-}ichi Amari}.}
  \bibinfo{year}{1998}\natexlab{}.
\newblock \showarticletitle{Natural Gradient Works Efficiently in Learning}.
\newblock \bibinfo{journal}{\emph{Neural Comput.}} \bibinfo{volume}{10},
  \bibinfo{number}{2} (\bibinfo{year}{1998}), \bibinfo{pages}{251--276}.
\newblock


\bibitem[\protect\citeauthoryear{Amari, Park, and Fukumizu}{Amari
  et~al\mbox{.}}{2000}]%
  {Adaptive_Method_of_Realizing_Natural_Gradient_Learning_for_Multilayer_Perceptrons}
\bibfield{author}{\bibinfo{person}{Shun{-}ichi Amari},
  \bibinfo{person}{Hyeyoung Park}, {and} \bibinfo{person}{Kenji Fukumizu}.}
  \bibinfo{year}{2000}\natexlab{}.
\newblock \showarticletitle{Adaptive Method of Realizing Natural Gradient
  Learning for Multilayer Perceptrons}.
\newblock \bibinfo{journal}{\emph{Neural Comput.}} \bibinfo{volume}{12},
  \bibinfo{number}{6} (\bibinfo{year}{2000}), \bibinfo{pages}{1399--1409}.
\newblock


\bibitem[\protect\citeauthoryear{Anil, Gupta, Koren, Regan, and Singer}{Anil
  et~al\mbox{.}}{2020}]%
        {scalable_shampoo}
\bibfield{author}{\bibinfo{person}{Rohan Anil}, \bibinfo{person}{Vineet Gupta},
  \bibinfo{person}{Tomer Koren}, \bibinfo{person}{Kevin Regan}, {and}
  \bibinfo{person}{Yoram Singer}.} \bibinfo{year}{2020}\natexlab{}.
\newblock \showarticletitle{Scalable second order optimization for deep
  learning}.
\newblock \bibinfo{journal}{\emph{arXiv preprint arXiv:2002.09018}}
  (\bibinfo{year}{2020}).
\newblock


\bibitem[\protect\citeauthoryear{Anil, Gupta, Koren, and Singer}{Anil
  et~al\mbox{.}}{2019}]%
        {SM3}
\bibfield{author}{\bibinfo{person}{Rohan Anil}, \bibinfo{person}{Vineet Gupta},
  \bibinfo{person}{Tomer Koren}, {and} \bibinfo{person}{Yoram Singer}.}
  \bibinfo{year}{2019}\natexlab{}.
\newblock \showarticletitle{Memory Efficient Adaptive Optimization}. In
  \bibinfo{booktitle}{\emph{NeurIPS}}.
\newblock


\bibitem[\protect\citeauthoryear{Battiti}{Battiti}{1992}]%
        {6796060}
\bibfield{author}{\bibinfo{person}{Roberto Battiti}.}
  \bibinfo{year}{1992}\natexlab{}.
\newblock \showarticletitle{First- and Second-Order Methods for Learning:
  Between Steepest Descent and Newton's Method}.
\newblock \bibinfo{journal}{\emph{Neural Computation}} \bibinfo{volume}{4},
  \bibinfo{number}{2} (\bibinfo{year}{1992}), \bibinfo{pages}{141--166}.
\newblock
\urldef\tempurl%
\url{https://doi.org/10.1162/neco.1992.4.2.141}
\showDOI{\tempurl}


\bibitem[\protect\citeauthoryear{Ben{-}Nun and Hoefler}{Ben{-}Nun and
  Hoefler}{2019}]%
  {Demystifying_Parallel_and_Distributed_Deep_Learning_An_In_depth_Concurrency_Analysis}
\bibfield{author}{\bibinfo{person}{Tal Ben{-}Nun} {and}
  \bibinfo{person}{Torsten Hoefler}.} \bibinfo{year}{2019}\natexlab{}.
\newblock \showarticletitle{Demystifying Parallel and Distributed Deep
  Learning: An In-depth Concurrency Analysis}.
\newblock \bibinfo{journal}{\emph{{ACM} Comput. Surv.}} \bibinfo{volume}{52},
  \bibinfo{number}{4} (\bibinfo{year}{2019}), \bibinfo{pages}{65:1--65:43}.
\newblock


\bibitem[\protect\citeauthoryear{Bernstein, Wang, Azizzadenesheli, and
  Anandkumar}{Bernstein et~al\mbox{.}}{2018}]%
        {1_bit_SGD_convergence}
\bibfield{author}{\bibinfo{person}{Jeremy Bernstein}, \bibinfo{person}{Yu-Xiang
  Wang}, \bibinfo{person}{Kamyar Azizzadenesheli}, {and}
  \bibinfo{person}{Animashree Anandkumar}.} \bibinfo{year}{2018}\natexlab{}.
\newblock \showarticletitle{signSGD: Compressed optimisation for non-convex
  problems}. In \bibinfo{booktitle}{\emph{International Conference on Machine
  Learning}}. PMLR, \bibinfo{pages}{560--569}.
\newblock


\bibitem[\protect\citeauthoryear{Betzel, Khatamifard, Suresh, Lilja, Sartori,
  and Karpuzcu}{Betzel et~al\mbox{.}}{2018}]%
  {survey_Approximate_Communication_Techniques_for_Reducing_Communication_Bottlenecks_in_Large_Scale_Parallel_Systems}
\bibfield{author}{\bibinfo{person}{Filipe Betzel}, \bibinfo{person}{S.~Karen
  Khatamifard}, \bibinfo{person}{Harini Suresh}, \bibinfo{person}{David~J.
  Lilja}, \bibinfo{person}{John Sartori}, {and} \bibinfo{person}{Ulya~R.
  Karpuzcu}.} \bibinfo{year}{2018}\natexlab{}.
\newblock \showarticletitle{Approximate Communication: Techniques for Reducing
  Communication Bottlenecks in Large-Scale Parallel Systems}.
\newblock \bibinfo{journal}{\emph{{ACM} Comput. Surv.}} \bibinfo{volume}{51},
  \bibinfo{number}{1} (\bibinfo{year}{2018}), \bibinfo{pages}{1:1--1:32}.
\newblock


\bibitem[\protect\citeauthoryear{Bollapragada, Byrd, and Nocedal}{Bollapragada
  et~al\mbox{.}}{2016}]%
        {bollapragada2016exact}
\bibfield{author}{\bibinfo{person}{Raghu Bollapragada},
  \bibinfo{person}{Richard Byrd}, {and} \bibinfo{person}{Jorge Nocedal}.}
  \bibinfo{year}{2016}\natexlab{}.
\newblock \bibinfo{title}{Exact and Inexact Subsampled Newton Methods for
  Optimization}.
\newblock
\newblock
\showeprint[arxiv]{1609.08502}~[math.OC]


\bibitem[\protect\citeauthoryear{Bordes, Bottou, and Gallinari}{Bordes
  et~al\mbox{.}}{2009}]%
        {SGD_QN}
\bibfield{author}{\bibinfo{person}{Antoine Bordes}, \bibinfo{person}{L{\'{e}}on
  Bottou}, {and} \bibinfo{person}{Patrick Gallinari}.}
  \bibinfo{year}{2009}\natexlab{}.
\newblock \showarticletitle{{SGD-QN:} Careful Quasi-Newton Stochastic Gradient
  Descent}.
\newblock \bibinfo{journal}{\emph{J. Mach. Learn. Res.}}  \bibinfo{volume}{10}
  (\bibinfo{year}{2009}), \bibinfo{pages}{1737--1754}.
\newblock


\bibitem[\protect\citeauthoryear{Botev, Ritter, and Barber}{Botev
  et~al\mbox{.}}{2017}]%
        {Diagonal_Gauss_Newton}
\bibfield{author}{\bibinfo{person}{Aleksandar Botev}, \bibinfo{person}{Hippolyt
  Ritter}, {and} \bibinfo{person}{David Barber}.}
  \bibinfo{year}{2017}\natexlab{}.
\newblock \showarticletitle{Practical gauss-newton optimisation for deep
  learning}. In \bibinfo{booktitle}{\emph{International Conference on Machine
  Learning}}. PMLR, \bibinfo{pages}{557--565}.
\newblock


\bibitem[\protect\citeauthoryear{Bottou and Bousquet}{Bottou and
  Bousquet}{2007}]%
        {The_Tradeoffs_of_Large_Scale_Learning}
\bibfield{author}{\bibinfo{person}{L{\'{e}}on Bottou} {and}
  \bibinfo{person}{Olivier Bousquet}.} \bibinfo{year}{2007}\natexlab{}.
\newblock \showarticletitle{The Tradeoffs of Large Scale Learning}. In
  \bibinfo{booktitle}{\emph{Proceedings of the 20th International Conference on
  Neural Information Processing Systems}}. \bibinfo{pages}{161–168}.
\newblock


\bibitem[\protect\citeauthoryear{Bottou, Curtis, and Nocedal}{Bottou
  et~al\mbox{.}}{2018}]%
        {Optimization_Methods_for_Large_Scale_Machine_Learning}
\bibfield{author}{\bibinfo{person}{L{\'{e}}on Bottou},
  \bibinfo{person}{Frank~E. Curtis}, {and} \bibinfo{person}{Jorge Nocedal}.}
  \bibinfo{year}{2018}\natexlab{}.
\newblock \showarticletitle{Optimization Methods for Large-Scale Machine
  Learning}.
\newblock \bibinfo{journal}{\emph{{SIAM} Rev.}} \bibinfo{volume}{60},
  \bibinfo{number}{2} (\bibinfo{year}{2018}), \bibinfo{pages}{223--311}.
\newblock


\bibitem[\protect\citeauthoryear{Byrd, Chin, Nocedal, and Wu}{Byrd
  et~al\mbox{.}}{2012}]%
        {Sample_size_selection_in_optimization_methods_for_machine_learning}
\bibfield{author}{\bibinfo{person}{Richard~H. Byrd},
  \bibinfo{person}{Gillian~M. Chin}, \bibinfo{person}{Jorge Nocedal}, {and}
  \bibinfo{person}{Yuchen Wu}.} \bibinfo{year}{2012}\natexlab{}.
\newblock \showarticletitle{Sample size selection in optimization methods for
  machine learning}.
\newblock \bibinfo{journal}{\emph{Math. Program.}} \bibinfo{volume}{134},
  \bibinfo{number}{1} (\bibinfo{year}{2012}), \bibinfo{pages}{127--155}.
\newblock


\bibitem[\protect\citeauthoryear{Carmon, Duchi, Hinder, and Sidford}{Carmon
  et~al\mbox{.}}{2018}]%
        {Accelerated_Methods_for_NonConvex_Optimization}
\bibfield{author}{\bibinfo{person}{Yair Carmon}, \bibinfo{person}{John~C.
  Duchi}, \bibinfo{person}{Oliver Hinder}, {and} \bibinfo{person}{Aaron
  Sidford}.} \bibinfo{year}{2018}\natexlab{}.
\newblock \showarticletitle{Accelerated Methods for NonConvex Optimization}.
\newblock \bibinfo{journal}{\emph{{SIAM} J. Optim.}} \bibinfo{volume}{28},
  \bibinfo{number}{2} (\bibinfo{year}{2018}), \bibinfo{pages}{1751--1772}.
\newblock


\bibitem[\protect\citeauthoryear{Chaudhari, Choromanska, Soatto, LeCun,
  Baldassi, Borgs, Chayes, Sagun, and Zecchina}{Chaudhari
  et~al\mbox{.}}{2017}]%
        {entropy_SGD}
\bibfield{author}{\bibinfo{person}{Pratik Chaudhari}, \bibinfo{person}{Anna
  Choromanska}, \bibinfo{person}{Stefano Soatto}, \bibinfo{person}{Yann LeCun},
  \bibinfo{person}{Carlo Baldassi}, \bibinfo{person}{Christian Borgs},
  \bibinfo{person}{Jennifer~T. Chayes}, \bibinfo{person}{Levent Sagun}, {and}
  \bibinfo{person}{Riccardo Zecchina}.} \bibinfo{year}{2017}\natexlab{}.
\newblock \showarticletitle{Entropy-SGD: Biasing Gradient Descent Into Wide
  Valleys}. In \bibinfo{booktitle}{\emph{5th International Conference on
  Learning Representations, {ICLR} 2017, Toulon, France, April 24-26, 2017,
  Conference Track Proceedings}}.
\newblock


\bibitem[\protect\citeauthoryear{Chen, Choi, Brand, Agrawal, Zhang, and
  Gopalakrishnan}{Chen et~al\mbox{.}}{2018}]%
        {AdaComp}
\bibfield{author}{\bibinfo{person}{Chia-Yu Chen}, \bibinfo{person}{Jungwook
  Choi}, \bibinfo{person}{Daniel Brand}, \bibinfo{person}{Ankur Agrawal},
  \bibinfo{person}{Wei Zhang}, {and} \bibinfo{person}{Kailash Gopalakrishnan}.}
  \bibinfo{year}{2018}\natexlab{}.
\newblock \showarticletitle{Adacomp: Adaptive residual gradient compression for
  data-parallel distributed training}. In \bibinfo{booktitle}{\emph{Proceedings
  of the AAAI Conference on Artificial Intelligence}},
  Vol.~\bibinfo{volume}{32}.
\newblock


\bibitem[\protect\citeauthoryear{Chen, Kornblith, Norouzi, and Hinton}{Chen
  et~al\mbox{.}}{2020}]%
        {ChenK0H20}
\bibfield{author}{\bibinfo{person}{Ting Chen}, \bibinfo{person}{Simon
  Kornblith}, \bibinfo{person}{Mohammad Norouzi}, {and}
  \bibinfo{person}{Geoffrey~E. Hinton}.} \bibinfo{year}{2020}\natexlab{}.
\newblock \showarticletitle{A Simple Framework for Contrastive Learning of
  Visual Representations}. In \bibinfo{booktitle}{\emph{Proceedings of the 37th
  International Conference on Machine Learning, {ICML} 2020, 13-18 July 2020,
  Virtual Event}} \emph{(\bibinfo{series}{Proceedings of Machine Learning
  Research}, Vol.~\bibinfo{volume}{119})}. \bibinfo{publisher}{{PMLR}},
  \bibinfo{pages}{1597--1607}.
\newblock


\bibitem[\protect\citeauthoryear{Choi, Shallue, Nado, Lee, Maddison, and
  Dahl}{Choi et~al\mbox{.}}{2019}]%
        {On_Empirical_Comparisons_of_Optimizers_for_Deep_Learning}
\bibfield{author}{\bibinfo{person}{Dami Choi}, \bibinfo{person}{Christopher~J.
  Shallue}, \bibinfo{person}{Zachary Nado}, \bibinfo{person}{Jaehoon Lee},
  \bibinfo{person}{Chris~J. Maddison}, {and} \bibinfo{person}{George~E. Dahl}.}
  \bibinfo{year}{2019}\natexlab{}.
\newblock \showarticletitle{On Empirical Comparisons of Optimizers for Deep
  Learning}.
\newblock \bibinfo{journal}{\emph{CoRR}}  \bibinfo{volume}{abs/1910.05446}
  (\bibinfo{year}{2019}).
\newblock


\bibitem[\protect\citeauthoryear{Conn, Gould, and Toint}{Conn
  et~al\mbox{.}}{2000}]%
        {Trust_Region_Methods}
\bibfield{author}{\bibinfo{person}{Andrew~R. Conn}, \bibinfo{person}{Nicholas
  I.~M. Gould}, {and} \bibinfo{person}{Philippe~L. Toint}.}
  \bibinfo{year}{2000}\natexlab{}.
\newblock \bibinfo{booktitle}{\emph{Trust Region Methods}}.
\newblock \bibinfo{publisher}{{SIAM}}.
\newblock


\bibitem[\protect\citeauthoryear{Devarakonda, Naumov, and Garland}{Devarakonda
  et~al\mbox{.}}{2017}]%
        {AdaBatch}
\bibfield{author}{\bibinfo{person}{Aditya Devarakonda}, \bibinfo{person}{Maxim
  Naumov}, {and} \bibinfo{person}{Michael Garland}.}
  \bibinfo{year}{2017}\natexlab{}.
\newblock \showarticletitle{AdaBatch: Adaptive Batch Sizes for Training Deep
  Neural Networks}.
\newblock \bibinfo{journal}{\emph{ArXiv}}  \bibinfo{volume}{abs/1712.02029}
  (\bibinfo{year}{2017}).
\newblock


\bibitem[\protect\citeauthoryear{Devlin, Chang, Lee, and Toutanova}{Devlin
  et~al\mbox{.}}{2019}]%
        {bert}
\bibfield{author}{\bibinfo{person}{Jacob Devlin}, \bibinfo{person}{Ming-Wei
  Chang}, \bibinfo{person}{Kenton Lee}, {and} \bibinfo{person}{Kristina
  Toutanova}.} \bibinfo{year}{2019}\natexlab{}.
\newblock \showarticletitle{BERT: Pre-training of Deep Bidirectional
  Transformers for Language Understanding}. In
  \bibinfo{booktitle}{\emph{NAACL-HLT (1)}}.
\newblock


\bibitem[\protect\citeauthoryear{Dinh, Pascanu, Bengio, and Bengio}{Dinh
  et~al\mbox{.}}{2017}]%
        {Sharp_Minima_Can_Generalize}
\bibfield{author}{\bibinfo{person}{Laurent Dinh}, \bibinfo{person}{Razvan
  Pascanu}, \bibinfo{person}{Samy Bengio}, {and} \bibinfo{person}{Yoshua
  Bengio}.} \bibinfo{year}{2017}\natexlab{}.
\newblock \showarticletitle{Sharp Minima Can Generalize For Deep Nets}. In
  \bibinfo{booktitle}{\emph{International Conference on Machine Learning}}.
  PMLR, \bibinfo{pages}{1019--1028}.
\newblock


\bibitem[\protect\citeauthoryear{Dozat}{Dozat}{2016}]%
        {NAdam}
\bibfield{author}{\bibinfo{person}{Timothy Dozat}.}
  \bibinfo{year}{2016}\natexlab{}.
\newblock \showarticletitle{Incorporating nesterov momentum into adam}.
\newblock  (\bibinfo{year}{2016}).
\newblock


\bibitem[\protect\citeauthoryear{Dryden, Moon, Jacobs, and Van~Essen}{Dryden
  et~al\mbox{.}}{2016}]%
        {adaptive_gradient_quantization}
\bibfield{author}{\bibinfo{person}{Nikoli Dryden}, \bibinfo{person}{Tim Moon},
  \bibinfo{person}{Sam~Ade Jacobs}, {and} \bibinfo{person}{Brian Van~Essen}.}
  \bibinfo{year}{2016}\natexlab{}.
\newblock \showarticletitle{Communication quantization for data-parallel
  training of deep neural networks}. In \bibinfo{booktitle}{\emph{2016 2nd
  Workshop on Machine Learning in HPC Environments (MLHPC)}}. IEEE,
  \bibinfo{pages}{1--8}.
\newblock


\bibitem[\protect\citeauthoryear{Duchi, Hazan, and Singer}{Duchi
  et~al\mbox{.}}{2011}]%
        {AdaGrad}
\bibfield{author}{\bibinfo{person}{John Duchi}, \bibinfo{person}{Elad Hazan},
  {and} \bibinfo{person}{Yoram Singer}.} \bibinfo{year}{2011}\natexlab{}.
\newblock \showarticletitle{Adaptive subgradient methods for online learning
  and stochastic optimization.}
\newblock \bibinfo{journal}{\emph{Journal of machine learning research}}
  \bibinfo{volume}{12}, \bibinfo{number}{7} (\bibinfo{year}{2011}).
\newblock


\bibitem[\protect\citeauthoryear{Foret, Kleiner, Mobahi, and Neyshabur}{Foret
  et~al\mbox{.}}{2021}]%
        {foret2021sharpnessaware}
\bibfield{author}{\bibinfo{person}{Pierre Foret}, \bibinfo{person}{Ariel
  Kleiner}, \bibinfo{person}{Hossein Mobahi}, {and} \bibinfo{person}{Behnam
  Neyshabur}.} \bibinfo{year}{2021}\natexlab{}.
\newblock \bibinfo{title}{Sharpness-Aware Minimization for Efficiently
  Improving Generalization}.
\newblock
\newblock
\showeprint[arxiv]{2010.01412}~[cs.LG]


\bibitem[\protect\citeauthoryear{Ge, Huang, Jin, and Yuan}{Ge
  et~al\mbox{.}}{2015}]%
        {GeHJY15}
\bibfield{author}{\bibinfo{person}{Rong Ge}, \bibinfo{person}{Furong Huang},
  \bibinfo{person}{Chi Jin}, {and} \bibinfo{person}{Yang Yuan}.}
  \bibinfo{year}{2015}\natexlab{}.
\newblock \showarticletitle{Escaping from saddle points—online stochastic
  gradient for tensor decomposition}. In \bibinfo{booktitle}{\emph{Conference
  on learning theory}}. PMLR, \bibinfo{pages}{797--842}.
\newblock


\bibitem[\protect\citeauthoryear{Girshick}{Girshick}{2015}]%
        {Fast_R_CNN}
\bibfield{author}{\bibinfo{person}{Ross Girshick}.}
  \bibinfo{year}{2015}\natexlab{}.
\newblock \showarticletitle{Fast r-cnn}. In
  \bibinfo{booktitle}{\emph{Proceedings of the IEEE international conference on
  computer vision}}. \bibinfo{pages}{1440--1448}.
\newblock


\bibitem[\protect\citeauthoryear{Goldfarb, Ren, and Bahamou}{Goldfarb
  et~al\mbox{.}}{2020}]%
        {quasi_newton}
\bibfield{author}{\bibinfo{person}{Donald Goldfarb}, \bibinfo{person}{Yi Ren},
  {and} \bibinfo{person}{Achraf Bahamou}.} \bibinfo{year}{2020}\natexlab{}.
\newblock \showarticletitle{Practical Quasi-Newton Methods for Training Deep
  Neural Networks}. In \bibinfo{booktitle}{\emph{Advances in Neural Information
  Processing Systems 33: Annual Conference on Neural Information Processing
  Systems 2020, NeurIPS 2020}}.
\newblock


\bibitem[\protect\citeauthoryear{Goyal, Dollár, Girshick, Noordhuis,
  Wesolowski, Kyrola, Tulloch, Jia, and He}{Goyal et~al\mbox{.}}{2018}]%
        {Training_ImageNet_in_1_Hour}
\bibfield{author}{\bibinfo{person}{Priya Goyal}, \bibinfo{person}{Piotr
  Dollár}, \bibinfo{person}{Ross Girshick}, \bibinfo{person}{Pieter
  Noordhuis}, \bibinfo{person}{Lukasz Wesolowski}, \bibinfo{person}{Aapo
  Kyrola}, \bibinfo{person}{Andrew Tulloch}, \bibinfo{person}{Yangqing Jia},
  {and} \bibinfo{person}{Kaiming He}.} \bibinfo{year}{2018}\natexlab{}.
\newblock \bibinfo{title}{Accurate, Large Minibatch SGD: Training ImageNet in 1
  Hour}.
\newblock
\newblock
\showeprint[arxiv]{1706.02677}~[cs.CV]


\bibitem[\protect\citeauthoryear{Grill, Strub, Altch{\'e}, Tallec, Richemond,
  Buchatskaya, Doersch, Pires, Guo, Azar, et~al\mbox{.}}{Grill
  et~al\mbox{.}}{2020}]%
        {GrillSATRBDPGAP20}
\bibfield{author}{\bibinfo{person}{Jean-Bastien Grill},
  \bibinfo{person}{Florian Strub}, \bibinfo{person}{Florent Altch{\'e}},
  \bibinfo{person}{Corentin Tallec}, \bibinfo{person}{Pierre Richemond},
  \bibinfo{person}{Elena Buchatskaya}, \bibinfo{person}{Carl Doersch},
  \bibinfo{person}{Bernardo Pires}, \bibinfo{person}{Zhaohan Guo},
  \bibinfo{person}{Mohammad Azar}, {et~al\mbox{.}}}
  \bibinfo{year}{2020}\natexlab{}.
\newblock \showarticletitle{Bootstrap Your Own Latent: A new approach to
  self-supervised learning}. In \bibinfo{booktitle}{\emph{Neural Information
  Processing Systems}}.
\newblock


\bibitem[\protect\citeauthoryear{Grosse and Martens}{Grosse and
  Martens}{2016}]%
        {KFAC_CNN}
\bibfield{author}{\bibinfo{person}{Roger~B Grosse} {and} \bibinfo{person}{James
  Martens}.} \bibinfo{year}{2016}\natexlab{}.
\newblock \showarticletitle{A Kronecker-factored approximate Fisher matrix for
  convolution layers}. In \bibinfo{booktitle}{\emph{ICML}},
  Vol.~\bibinfo{volume}{48}. \bibinfo{pages}{573--582}.
\newblock


\bibitem[\protect\citeauthoryear{Gupta, Koren, and Singer}{Gupta
  et~al\mbox{.}}{2018}]%
        {shampoo}
\bibfield{author}{\bibinfo{person}{Vineet Gupta}, \bibinfo{person}{Tomer
  Koren}, {and} \bibinfo{person}{Yoram Singer}.}
  \bibinfo{year}{2018}\natexlab{}.
\newblock \showarticletitle{Shampoo: Preconditioned Stochastic Tensor
  Optimization}. In \bibinfo{booktitle}{\emph{ICML}},
  Vol.~\bibinfo{volume}{80}. \bibinfo{pages}{1837--1845}.
\newblock


\bibitem[\protect\citeauthoryear{He, Gkioxari, Doll{\'{a}}r, and Girshick}{He
  et~al\mbox{.}}{2020}]%
        {Mask_R_CNN}
\bibfield{author}{\bibinfo{person}{Kaiming He}, \bibinfo{person}{Georgia
  Gkioxari}, \bibinfo{person}{Piotr Doll{\'{a}}r}, {and}
  \bibinfo{person}{Ross~B. Girshick}.} \bibinfo{year}{2020}\natexlab{}.
\newblock \showarticletitle{Mask {R-CNN}}.
\newblock \bibinfo{journal}{\emph{{IEEE} Trans. Pattern Anal. Mach. Intell.}}
  \bibinfo{volume}{42}, \bibinfo{number}{2} (\bibinfo{year}{2020}),
  \bibinfo{pages}{386--397}.
\newblock


\bibitem[\protect\citeauthoryear{He, Zhang, Ren, and Sun}{He
  et~al\mbox{.}}{2016}]%
        {image_classification}
\bibfield{author}{\bibinfo{person}{Kaiming He}, \bibinfo{person}{Xiangyu
  Zhang}, \bibinfo{person}{Shaoqing Ren}, {and} \bibinfo{person}{Jian Sun}.}
  \bibinfo{year}{2016}\natexlab{}.
\newblock \showarticletitle{Deep residual learning for image recognition}. In
  \bibinfo{booktitle}{\emph{Proceedings of the IEEE conference on computer
  vision and pattern recognition}}. \bibinfo{pages}{770--778}.
\newblock


\bibitem[\protect\citeauthoryear{He, Liao, Zhang, Nie, Hu, and Chua}{He
  et~al\mbox{.}}{2017}]%
        {Neural_Collaborative_Filtering}
\bibfield{author}{\bibinfo{person}{Xiangnan He}, \bibinfo{person}{Lizi Liao},
  \bibinfo{person}{Hanwang Zhang}, \bibinfo{person}{Liqiang Nie},
  \bibinfo{person}{Xia Hu}, {and} \bibinfo{person}{Tat-Seng Chua}.}
  \bibinfo{year}{2017}\natexlab{}.
\newblock \showarticletitle{Neural Collaborative Filtering}. In
  \bibinfo{booktitle}{\emph{WWW}}. \bibinfo{pages}{173--182}.
\newblock


\bibitem[\protect\citeauthoryear{Hochreiter and Schmidhuber}{Hochreiter and
  Schmidhuber}{1997}]%
        {hochreiter1997flat}
\bibfield{author}{\bibinfo{person}{Sepp Hochreiter} {and}
  \bibinfo{person}{J{\"u}rgen Schmidhuber}.} \bibinfo{year}{1997}\natexlab{}.
\newblock \showarticletitle{Flat minima}.
\newblock \bibinfo{journal}{\emph{Neural computation}} \bibinfo{volume}{9},
  \bibinfo{number}{1} (\bibinfo{year}{1997}), \bibinfo{pages}{1--42}.
\newblock


\bibitem[\protect\citeauthoryear{Hoffer, Hubara, and Soudry}{Hoffer
  et~al\mbox{.}}{2017}]%
        {Train_longer_generalize_better}
\bibfield{author}{\bibinfo{person}{Elad Hoffer}, \bibinfo{person}{Itay Hubara},
  {and} \bibinfo{person}{Daniel Soudry}.} \bibinfo{year}{2017}\natexlab{}.
\newblock \showarticletitle{Train longer, generalize better: closing the
  generalization gap in large batch training of neural networks}. In
  \bibinfo{booktitle}{\emph{Proceedings of the 31st International Conference on
  Neural Information Processing Systems}}. \bibinfo{pages}{1729--1739}.
\newblock


\bibitem[\protect\citeauthoryear{Horvath, Ho, Horvath, Sahu, Canini, and
  Richt{\'{a}}rik}{Horvath et~al\mbox{.}}{2019}]%
        {Natural_Compression_for_Distributed_Deep_Learning}
\bibfield{author}{\bibinfo{person}{Samuel Horvath}, \bibinfo{person}{Chen{-}Yu
  Ho}, \bibinfo{person}{Ludovit Horvath}, \bibinfo{person}{Atal~Narayan Sahu},
  \bibinfo{person}{Marco Canini}, {and} \bibinfo{person}{Peter
  Richt{\'{a}}rik}.} \bibinfo{year}{2019}\natexlab{}.
\newblock \showarticletitle{Natural Compression for Distributed Deep Learning}.
\newblock \bibinfo{journal}{\emph{CoRR}}  \bibinfo{volume}{abs/1905.10988}
  (\bibinfo{year}{2019}).
\newblock


\bibitem[\protect\citeauthoryear{Huang, Liu, van~der Maaten, and
  Weinberger}{Huang et~al\mbox{.}}{2017}]%
        {image_classification_densely_connected_convolutional_networks}
\bibfield{author}{\bibinfo{person}{Gao Huang}, \bibinfo{person}{Zhuang Liu},
  \bibinfo{person}{Laurens van~der Maaten}, {and} \bibinfo{person}{Kilian~Q
  Weinberger}.} \bibinfo{year}{2017}\natexlab{}.
\newblock \showarticletitle{Densely Connected Convolutional Networks}. In
  \bibinfo{booktitle}{\emph{CVPR}}. \bibinfo{pages}{2261--2269}.
\newblock


\bibitem[\protect\citeauthoryear{Jastrz{\k{e}}bski, Kenton, Arpit, Ballas,
  Fischer, Bengio, and Storkey}{Jastrz{\k{e}}bski et~al\mbox{.}}{2017}]%
        {three_factor_influencing_minima_in_SGD}
\bibfield{author}{\bibinfo{person}{Stanis{\l}aw Jastrz{\k{e}}bski},
  \bibinfo{person}{Zachary Kenton}, \bibinfo{person}{Devansh Arpit},
  \bibinfo{person}{Nicolas Ballas}, \bibinfo{person}{Asja Fischer},
  \bibinfo{person}{Yoshua Bengio}, {and} \bibinfo{person}{Amos Storkey}.}
  \bibinfo{year}{2017}\natexlab{}.
\newblock \showarticletitle{Three factors influencing minima in sgd}.
\newblock \bibinfo{journal}{\emph{arXiv preprint arXiv:1711.04623}}
  (\bibinfo{year}{2017}).
\newblock


\bibitem[\protect\citeauthoryear{Jia, Song, He, Wang, Rong, Zhou, Xie, Guo,
  Yang, Yu, Chen, Hu, Shi, and Chu}{Jia et~al\mbox{.}}{2018}]%
        {DBLP:journals/corr/abs-1807-11205}
\bibfield{author}{\bibinfo{person}{Xianyan Jia}, \bibinfo{person}{Shutao Song},
  \bibinfo{person}{W. He}, \bibinfo{person}{Yangzihao Wang},
  \bibinfo{person}{Haidong Rong}, \bibinfo{person}{Feihu Zhou},
  \bibinfo{person}{Liqiang Xie}, \bibinfo{person}{Zhenyu Guo},
  \bibinfo{person}{Yuanzhou Yang}, \bibinfo{person}{Liwei Yu},
  \bibinfo{person}{Tiegang Chen}, \bibinfo{person}{Guangxiao Hu},
  \bibinfo{person}{Shaohuai Shi}, {and} \bibinfo{person}{Xiaowen Chu}.}
  \bibinfo{year}{2018}\natexlab{}.
\newblock \showarticletitle{Highly Scalable Deep Learning Training System with
  Mixed-Precision: Training ImageNet in Four Minutes}.
\newblock \bibinfo{journal}{\emph{ArXiv}}  \bibinfo{volume}{abs/1807.11205}
  (\bibinfo{year}{2018}).
\newblock


\bibitem[\protect\citeauthoryear{Kamp, Adilova, Sicking, H{\"u}ger, Schlicht,
  Wirtz, and Wrobel}{Kamp et~al\mbox{.}}{2018}]%
        {Efficient_Decentralized_Deep_Learning_by_Dynamic_Model_Averaging}
\bibfield{author}{\bibinfo{person}{Michael Kamp}, \bibinfo{person}{Linara
  Adilova}, \bibinfo{person}{Joachim Sicking}, \bibinfo{person}{Fabian
  H{\"u}ger}, \bibinfo{person}{Peter Schlicht}, \bibinfo{person}{Tim Wirtz},
  {and} \bibinfo{person}{Stefan Wrobel}.} \bibinfo{year}{2018}\natexlab{}.
\newblock \showarticletitle{Efficient Decentralized Deep Learning by Dynamic
  Model Averaging}. In \bibinfo{booktitle}{\emph{ECML/PKDD (1)}}.
\newblock


\bibitem[\protect\citeauthoryear{Karimireddy, Rebjock, Stich, and
  Jaggi}{Karimireddy et~al\mbox{.}}{2019}]%
        {EF_Sign_SGD}
\bibfield{author}{\bibinfo{person}{Sai~Praneeth Karimireddy},
  \bibinfo{person}{Quentin Rebjock}, \bibinfo{person}{Sebastian Stich}, {and}
  \bibinfo{person}{Martin Jaggi}.} \bibinfo{year}{2019}\natexlab{}.
\newblock \showarticletitle{Error feedback fixes signsgd and other gradient
  compression schemes}. In \bibinfo{booktitle}{\emph{International Conference
  on Machine Learning}}. PMLR, \bibinfo{pages}{3252--3261}.
\newblock


\bibitem[\protect\citeauthoryear{Keskar, Mudigere, Nocedal, Smelyanskiy, and
  Tang}{Keskar et~al\mbox{.}}{2017}]%
        {generalization_gap_and_sharp_minima}
\bibfield{author}{\bibinfo{person}{Nitish~Shirish Keskar},
  \bibinfo{person}{Dheevatsa Mudigere}, \bibinfo{person}{Jorge Nocedal},
  \bibinfo{person}{Mikhail Smelyanskiy}, {and} \bibinfo{person}{Ping Tak~Peter
  Tang}.} \bibinfo{year}{2017}\natexlab{}.
\newblock \bibinfo{title}{On Large-Batch Training for Deep Learning:
  Generalization Gap and Sharp Minima}.
\newblock
\newblock
\showeprint[arxiv]{1609.04836}~[cs.LG]


\bibitem[\protect\citeauthoryear{Kingma and Ba}{Kingma and Ba}{2017}]%
        {kingma2017adam}
\bibfield{author}{\bibinfo{person}{Diederik~P. Kingma} {and}
  \bibinfo{person}{Jimmy Ba}.} \bibinfo{year}{2017}\natexlab{}.
\newblock \bibinfo{title}{Adam: A Method for Stochastic Optimization}.
\newblock
\newblock
\showeprint[arxiv]{1412.6980}~[cs.LG]


\bibitem[\protect\citeauthoryear{Krizhevsky}{Krizhevsky}{2014}]%
        {one_weird_trick}
\bibfield{author}{\bibinfo{person}{Alex Krizhevsky}.}
  \bibinfo{year}{2014}\natexlab{}.
\newblock \showarticletitle{One weird trick for parallelizing convolutional
  neural networks}.
\newblock \bibinfo{journal}{\emph{arXiv preprint arXiv:1404.5997}}
  (\bibinfo{year}{2014}).
\newblock


\bibitem[\protect\citeauthoryear{Krizhevsky, Sutskever, and Hinton}{Krizhevsky
  et~al\mbox{.}}{2012}]%
        {KrizhevskySH12}
\bibfield{author}{\bibinfo{person}{Alex Krizhevsky}, \bibinfo{person}{Ilya
  Sutskever}, {and} \bibinfo{person}{Geoffrey~E Hinton}.}
  \bibinfo{year}{2012}\natexlab{}.
\newblock \showarticletitle{Imagenet classification with deep convolutional
  neural networks}.
\newblock \bibinfo{journal}{\emph{Advances in neural information processing
  systems}}  \bibinfo{volume}{25} (\bibinfo{year}{2012}),
  \bibinfo{pages}{1097--1105}.
\newblock


\bibitem[\protect\citeauthoryear{Krizhevsky, Sutskever, and Hinton}{Krizhevsky
  et~al\mbox{.}}{2017}]%
        {KrizhevskySH17}
\bibfield{author}{\bibinfo{person}{Alex Krizhevsky}, \bibinfo{person}{Ilya
  Sutskever}, {and} \bibinfo{person}{Geoffrey~E. Hinton}.}
  \bibinfo{year}{2017}\natexlab{}.
\newblock \showarticletitle{ImageNet classification with deep convolutional
  neural networks}.
\newblock \bibinfo{journal}{\emph{Commun. {ACM}}} \bibinfo{volume}{60},
  \bibinfo{number}{6} (\bibinfo{year}{2017}), \bibinfo{pages}{84--90}.
\newblock


\bibitem[\protect\citeauthoryear{Lan, Chen, Goodman, Gimpel, Sharma, and
  Soricut}{Lan et~al\mbox{.}}{2019}]%
        {ALBERT}
\bibfield{author}{\bibinfo{person}{Zhenzhong Lan}, \bibinfo{person}{Mingda
  Chen}, \bibinfo{person}{Sebastian Goodman}, \bibinfo{person}{Kevin Gimpel},
  \bibinfo{person}{Piyush Sharma}, {and} \bibinfo{person}{Radu Soricut}.}
  \bibinfo{year}{2019}\natexlab{}.
\newblock \showarticletitle{Albert: A lite bert for self-supervised learning of
  language representations}.
\newblock \bibinfo{journal}{\emph{arXiv preprint arXiv:1909.11942}}
  (\bibinfo{year}{2019}).
\newblock


\bibitem[\protect\citeauthoryear{Li, Awan, Tang, Rajbhandari, and He}{Li
  et~al\mbox{.}}{2021}]%
        {1_bit_LAMB}
\bibfield{author}{\bibinfo{person}{Conglong Li}, \bibinfo{person}{Ammar~Ahmad
  Awan}, \bibinfo{person}{Hanlin Tang}, \bibinfo{person}{Samyam Rajbhandari},
  {and} \bibinfo{person}{Yuxiong He}.} \bibinfo{year}{2021}\natexlab{}.
\newblock \showarticletitle{1-bit LAMB: Communication Efficient Large-Scale
  Large-Batch Training with LAMB's Convergence Speed}.
\newblock \bibinfo{journal}{\emph{ArXiv}}  \bibinfo{volume}{abs/2104.06069}
  (\bibinfo{year}{2021}).
\newblock


\bibitem[\protect\citeauthoryear{Li, Xu, Taylor, Studer, and Goldstein}{Li
  et~al\mbox{.}}{2018}]%
        {Visualizing_the_Loss_Landscape_of_Neural_Nets}
\bibfield{author}{\bibinfo{person}{Hao Li}, \bibinfo{person}{Zheng Xu},
  \bibinfo{person}{Gavin Taylor}, \bibinfo{person}{Christoph Studer}, {and}
  \bibinfo{person}{Tom Goldstein}.} \bibinfo{year}{2018}\natexlab{}.
\newblock \showarticletitle{Visualizing the loss landscape of neural nets}. In
  \bibinfo{booktitle}{\emph{Proceedings of the 32nd International Conference on
  Neural Information Processing Systems}}. \bibinfo{pages}{6391--6401}.
\newblock


\bibitem[\protect\citeauthoryear{Li, Andersen, Smola, and Yu}{Li
  et~al\mbox{.}}{2014a}]%
  {Communication_Efficient_Distributed_Machine_Learning_with_the_Parameter_Server}
\bibfield{author}{\bibinfo{person}{Mu Li}, \bibinfo{person}{David~G Andersen},
  \bibinfo{person}{Alexander~J Smola}, {and} \bibinfo{person}{Kai Yu}.}
  \bibinfo{year}{2014}\natexlab{a}.
\newblock \showarticletitle{Communication efficient distributed machine
  learning with the parameter server}.
\newblock \bibinfo{journal}{\emph{Advances in Neural Information Processing
  Systems}}  \bibinfo{volume}{27} (\bibinfo{year}{2014}),
  \bibinfo{pages}{19--27}.
\newblock


\bibitem[\protect\citeauthoryear{Li, Zhang, Chen, and Smola}{Li
  et~al\mbox{.}}{2014b}]%
        {Efficient_mini-batch_training_for_stochastic_optimization}
\bibfield{author}{\bibinfo{person}{Mu Li}, \bibinfo{person}{Tong Zhang},
  \bibinfo{person}{Yuqiang Chen}, {and} \bibinfo{person}{Alexander~J Smola}.}
  \bibinfo{year}{2014}\natexlab{b}.
\newblock \showarticletitle{Efficient mini-batch training for stochastic
  optimization}. In \bibinfo{booktitle}{\emph{Proceedings of the 20th ACM
  SIGKDD international conference on Knowledge discovery and data mining}}.
  \bibinfo{pages}{661--670}.
\newblock


\bibitem[\protect\citeauthoryear{Lin, Stich, Patel, and Jaggi}{Lin
  et~al\mbox{.}}{2020}]%
        {Use_Local_SGD}
\bibfield{author}{\bibinfo{person}{Tao Lin}, \bibinfo{person}{Sebastian~U.
  Stich}, \bibinfo{person}{Kumar~Kshitij Patel}, {and} \bibinfo{person}{Martin
  Jaggi}.} \bibinfo{year}{2020}\natexlab{}.
\newblock \bibinfo{title}{Don't Use Large Mini-Batches, Use Local SGD}.
\newblock
\newblock
\showeprint[arxiv]{1808.07217}~[cs.LG]


\bibitem[\protect\citeauthoryear{Lin, Han, Mao, Wang, and Dally}{Lin
  et~al\mbox{.}}{2018}]%
        {deep_gradient_compression}
\bibfield{author}{\bibinfo{person}{Yujun Lin}, \bibinfo{person}{Song Han},
  \bibinfo{person}{Huizi Mao}, \bibinfo{person}{Yu Wang}, {and}
  \bibinfo{person}{Bill Dally}.} \bibinfo{year}{2018}\natexlab{}.
\newblock \showarticletitle{Deep Gradient Compression: Reducing the
  Communication Bandwidth for Distributed Training}. In
  \bibinfo{booktitle}{\emph{International Conference on Learning
  Representations}}.
\newblock


\bibitem[\protect\citeauthoryear{Liu, Jiang, He, Chen, Liu, Gao, and Han}{Liu
  et~al\mbox{.}}{2019}]%
        {RAdam}
\bibfield{author}{\bibinfo{person}{Liyuan Liu}, \bibinfo{person}{Haoming
  Jiang}, \bibinfo{person}{Pengcheng He}, \bibinfo{person}{Weizhu Chen},
  \bibinfo{person}{Xiaodong Liu}, \bibinfo{person}{Jianfeng Gao}, {and}
  \bibinfo{person}{Jiawei Han}.} \bibinfo{year}{2019}\natexlab{}.
\newblock \showarticletitle{On the Variance of the Adaptive Learning Rate and
  Beyond}. In \bibinfo{booktitle}{\emph{International Conference on Learning
  Representations}}.
\newblock


\bibitem[\protect\citeauthoryear{Long, Shelhamer, and Darrell}{Long
  et~al\mbox{.}}{2015}]%
        {Fully_convolutional_networks_for_semantic_segmentation}
\bibfield{author}{\bibinfo{person}{Jonathan Long}, \bibinfo{person}{Evan
  Shelhamer}, {and} \bibinfo{person}{Trevor Darrell}.}
  \bibinfo{year}{2015}\natexlab{}.
\newblock \showarticletitle{Fully convolutional networks for semantic
  segmentation}. In \bibinfo{booktitle}{\emph{Proceedings of the IEEE
  conference on computer vision and pattern recognition}}.
  \bibinfo{pages}{3431--3440}.
\newblock


\bibitem[\protect\citeauthoryear{Lou, Xue, Zheng, and You}{Lou
  et~al\mbox{.}}{2021}]%
        {lou2021sparse}
\bibfield{author}{\bibinfo{person}{Yuxuan Lou}, \bibinfo{person}{Fuzhao Xue},
  \bibinfo{person}{Zangwei Zheng}, {and} \bibinfo{person}{Yang You}.}
  \bibinfo{year}{2021}\natexlab{}.
\newblock \showarticletitle{Sparse-MLP: A Fully-MLP Architecture with
  Conditional Computation}.
\newblock \bibinfo{journal}{\emph{arXiv preprint arXiv:2109.02008}}
  (\bibinfo{year}{2021}).
\newblock


\bibitem[\protect\citeauthoryear{Martens}{Martens}{2010}]%
        {Hessian_free}
\bibfield{author}{\bibinfo{person}{James Martens}.}
  \bibinfo{year}{2010}\natexlab{}.
\newblock \showarticletitle{Deep learning via Hessian-free optimization}. In
  \bibinfo{booktitle}{\emph{Proceedings of the 27th International Conference on
  Machine Learning (ICML-10), June 21-24, 2010, Haifa, Israel}}.
  \bibinfo{pages}{735--742}.
\newblock


\bibitem[\protect\citeauthoryear{Martens}{Martens}{2020}]%
        {New_Insights_and_Perspectives_on_the_Natural_Gradient_Method}
\bibfield{author}{\bibinfo{person}{James Martens}.}
  \bibinfo{year}{2020}\natexlab{}.
\newblock \showarticletitle{New Insights and Perspectives on the Natural
  Gradient Method}.
\newblock \bibinfo{journal}{\emph{J. Mach. Learn. Res.}}  \bibinfo{volume}{21}
  (\bibinfo{year}{2020}), \bibinfo{pages}{146:1--146:76}.
\newblock


\bibitem[\protect\citeauthoryear{Martens and Grosse}{Martens and
  Grosse}{2015}]%
        {K-FAC}
\bibfield{author}{\bibinfo{person}{James Martens} {and} \bibinfo{person}{Roger
  Grosse}.} \bibinfo{year}{2015}\natexlab{}.
\newblock \showarticletitle{Optimizing neural networks with kronecker-factored
  approximate curvature}. In \bibinfo{booktitle}{\emph{International conference
  on machine learning}}. PMLR, \bibinfo{pages}{2408--2417}.
\newblock


\bibitem[\protect\citeauthoryear{Masters and Luschi}{Masters and
  Luschi}{2018}]%
        {abs-1804-07612}
\bibfield{author}{\bibinfo{person}{Dominic Masters} {and}
  \bibinfo{person}{Carlo Luschi}.} \bibinfo{year}{2018}\natexlab{}.
\newblock \showarticletitle{Revisiting Small Batch Training for Deep Neural
  Networks}.
\newblock \bibinfo{journal}{\emph{CoRR}}  \bibinfo{volume}{abs/1804.07612}
  (\bibinfo{year}{2018}).
\newblock


\bibitem[\protect\citeauthoryear{McDonald, Hall, and Mann}{McDonald
  et~al\mbox{.}}{2010}]%
        {Distributed_Training_Strategies_for_the_Structured_Perceptron}
\bibfield{author}{\bibinfo{person}{Ryan McDonald}, \bibinfo{person}{Keith
  Hall}, {and} \bibinfo{person}{Gideon Mann}.} \bibinfo{year}{2010}\natexlab{}.
\newblock \showarticletitle{Distributed training strategies for the structured
  perceptron}. In \bibinfo{booktitle}{\emph{Human language technologies: The
  2010 annual conference of the North American chapter of the association for
  computational linguistics}}. \bibinfo{pages}{456--464}.
\newblock


\bibitem[\protect\citeauthoryear{Micikevicius, Narang, Alben, Diamos, Elsen,
  Garc{\'i}a, Ginsburg, Houston, Kuchaiev, Venkatesh, and Wu}{Micikevicius
  et~al\mbox{.}}{2018}]%
        {mix_precision_training}
\bibfield{author}{\bibinfo{person}{Paulius Micikevicius},
  \bibinfo{person}{Sharan Narang}, \bibinfo{person}{Jonah Alben},
  \bibinfo{person}{Gregory~Frederick Diamos}, \bibinfo{person}{Erich Elsen},
  \bibinfo{person}{David Garc{\'i}a}, \bibinfo{person}{Boris Ginsburg},
  \bibinfo{person}{Michael Houston}, \bibinfo{person}{Oleksii Kuchaiev},
  \bibinfo{person}{Ganesh Venkatesh}, {and} \bibinfo{person}{Hao Wu}.}
  \bibinfo{year}{2018}\natexlab{}.
\newblock \showarticletitle{Mixed Precision Training}.
\newblock \bibinfo{journal}{\emph{ArXiv}}  \bibinfo{volume}{abs/1710.03740}
  (\bibinfo{year}{2018}).
\newblock


\bibitem[\protect\citeauthoryear{Nado, Gilmer, Shallue, Anil, and Dahl}{Nado
  et~al\mbox{.}}{2021}]%
        {opp_LARS_LAMB}
\bibfield{author}{\bibinfo{person}{Zachary Nado}, \bibinfo{person}{Justin
  Gilmer}, \bibinfo{person}{Christopher~J. Shallue}, \bibinfo{person}{Rohan
  Anil}, {and} \bibinfo{person}{George~E. Dahl}.}
  \bibinfo{year}{2021}\natexlab{}.
\newblock \showarticletitle{A Large Batch Optimizer Reality Check: Traditional,
  Generic Optimizers Suffice Across Batch Sizes}.
\newblock \bibinfo{journal}{\emph{ArXiv}}  \bibinfo{volume}{abs/2102.06356}
  (\bibinfo{year}{2021}).
\newblock


\bibitem[\protect\citeauthoryear{Nesterov}{Nesterov}{1983}]%
        {nesterov1983method}
\bibfield{author}{\bibinfo{person}{Yurii~E Nesterov}.}
  \bibinfo{year}{1983}\natexlab{}.
\newblock \showarticletitle{A method for solving the convex programming problem
  with convergence rate O (1/k\^{} 2)}. In \bibinfo{booktitle}{\emph{Dokl.
  akad. nauk Sssr}}, Vol.~\bibinfo{volume}{269}. \bibinfo{pages}{543--547}.
\newblock


\bibitem[\protect\citeauthoryear{Osawa, Tsuji, Ueno, Naruse, Yokota, and
  Matsuoka}{Osawa et~al\mbox{.}}{2019}]%
        {distributed_K_FAC}
\bibfield{author}{\bibinfo{person}{Kazuki Osawa}, \bibinfo{person}{Yohei
  Tsuji}, \bibinfo{person}{Yuichiro Ueno}, \bibinfo{person}{Akira Naruse},
  \bibinfo{person}{Rio Yokota}, {and} \bibinfo{person}{Satoshi Matsuoka}.}
  \bibinfo{year}{2019}\natexlab{}.
\newblock \showarticletitle{Large-scale distributed second-order optimization
  using kronecker-factored approximate curvature for deep convolutional neural
  networks}. In \bibinfo{booktitle}{\emph{Proceedings of the IEEE/CVF
  Conference on Computer Vision and Pattern Recognition}}.
  \bibinfo{pages}{12359--12367}.
\newblock


\bibitem[\protect\citeauthoryear{{Park}, {Chan}, {Zhang}, {Chiu}, {Zoph},
  {Cubuk}, and {Le}}{{Park} et~al\mbox{.}}{2019}]%
        {park2019specaugment}
\bibfield{author}{\bibinfo{person}{Daniel~S. {Park}}, \bibinfo{person}{William
  {Chan}}, \bibinfo{person}{Yu {Zhang}}, \bibinfo{person}{Chung-Cheng {Chiu}},
  \bibinfo{person}{Barret {Zoph}}, \bibinfo{person}{Ekin~Dogus {Cubuk}}, {and}
  \bibinfo{person}{Quoc~V. {Le}}.} \bibinfo{year}{2019}\natexlab{}.
\newblock \showarticletitle{SpecAugment: A Simple Data Augmentation Method for
  Automatic Speech Recognition}. In \bibinfo{booktitle}{\emph{Interspeech
  2019}}. \bibinfo{pages}{2613--2617}.
\newblock


\bibitem[\protect\citeauthoryear{Paszke, Gross, Massa, Lerer, Bradbury, Chanan,
  Killeen, Lin, Gimelshein, Antiga, et~al\mbox{.}}{Paszke
  et~al\mbox{.}}{2019}]%
        {pytorch}
\bibfield{author}{\bibinfo{person}{Adam Paszke}, \bibinfo{person}{Sam Gross},
  \bibinfo{person}{Francisco Massa}, \bibinfo{person}{Adam Lerer},
  \bibinfo{person}{James Bradbury}, \bibinfo{person}{Gregory Chanan},
  \bibinfo{person}{Trevor Killeen}, \bibinfo{person}{Zeming Lin},
  \bibinfo{person}{Natalia Gimelshein}, \bibinfo{person}{Luca Antiga},
  {et~al\mbox{.}}} \bibinfo{year}{2019}\natexlab{}.
\newblock \showarticletitle{Pytorch: An imperative style, high-performance deep
  learning library}.
\newblock \bibinfo{journal}{\emph{Advances in neural information processing
  systems}}  \bibinfo{volume}{32} (\bibinfo{year}{2019}),
  \bibinfo{pages}{8026--8037}.
\newblock


\bibitem[\protect\citeauthoryear{Pauloski, Zhang, Huang, Xu, and
  Foster}{Pauloski et~al\mbox{.}}{2020}]%
        {feature_decomposition_K_FAC}
\bibfield{author}{\bibinfo{person}{J~Gregory Pauloski}, \bibinfo{person}{Zhao
  Zhang}, \bibinfo{person}{Lei Huang}, \bibinfo{person}{Weijia Xu}, {and}
  \bibinfo{person}{Ian~T Foster}.} \bibinfo{year}{2020}\natexlab{}.
\newblock \showarticletitle{Convolutional neural network training with
  distributed K-FAC}. In \bibinfo{booktitle}{\emph{SC20: International
  Conference for High Performance Computing, Networking, Storage and
  Analysis}}. IEEE, \bibinfo{pages}{1--12}.
\newblock


\bibitem[\protect\citeauthoryear{Pouyanfar, Sadiq, Yan, Tian, Tao, Reyes, Shyu,
  Chen, and Iyengar}{Pouyanfar et~al\mbox{.}}{2019}]%
        {A_Survey_on_Deep_Learning_Algorithms_Techniques_and_Applications}
\bibfield{author}{\bibinfo{person}{Samira Pouyanfar}, \bibinfo{person}{Saad
  Sadiq}, \bibinfo{person}{Yilin Yan}, \bibinfo{person}{Haiman Tian},
  \bibinfo{person}{Yudong Tao}, \bibinfo{person}{Maria E.~Presa Reyes},
  \bibinfo{person}{Mei{-}Ling Shyu}, \bibinfo{person}{Shu{-}Ching Chen}, {and}
  \bibinfo{person}{S.~S. Iyengar}.} \bibinfo{year}{2019}\natexlab{}.
\newblock \showarticletitle{A Survey on Deep Learning: Algorithms, Techniques,
  and Applications}.
\newblock \bibinfo{journal}{\emph{{ACM} Comput. Surv.}} \bibinfo{volume}{51},
  \bibinfo{number}{5} (\bibinfo{year}{2019}), \bibinfo{pages}{92:1--92:36}.
\newblock


\bibitem[\protect\citeauthoryear{Qian}{Qian}{1999}]%
        {journals/nn/Qian99}
\bibfield{author}{\bibinfo{person}{Ning Qian}.}
  \bibinfo{year}{1999}\natexlab{}.
\newblock \showarticletitle{On the momentum term in gradient descent learning
  algorithms}.
\newblock \bibinfo{journal}{\emph{Neural networks}} \bibinfo{volume}{12},
  \bibinfo{number}{1} (\bibinfo{year}{1999}), \bibinfo{pages}{145--151}.
\newblock


\bibitem[\protect\citeauthoryear{Rajbhandari, Rasley, Ruwase, and
  He}{Rajbhandari et~al\mbox{.}}{2020}]%
        {ZeRO}
\bibfield{author}{\bibinfo{person}{Samyam Rajbhandari}, \bibinfo{person}{Jeff
  Rasley}, \bibinfo{person}{Olatunji Ruwase}, {and} \bibinfo{person}{Yuxiong
  He}.} \bibinfo{year}{2020}\natexlab{}.
\newblock \showarticletitle{Zero: Memory optimizations toward training trillion
  parameter models}. In \bibinfo{booktitle}{\emph{SC20: International
  Conference for High Performance Computing, Networking, Storage and
  Analysis}}. IEEE, \bibinfo{pages}{1--16}.
\newblock


\bibitem[\protect\citeauthoryear{Ramezani{-}Kebrya, Faghri, Markov, Aksenov,
  Alistarh, and Roy}{Ramezani{-}Kebrya et~al\mbox{.}}{2021}]%
        {NUQSGD}
\bibfield{author}{\bibinfo{person}{Ali Ramezani{-}Kebrya},
  \bibinfo{person}{Fartash Faghri}, \bibinfo{person}{Ilia Markov},
  \bibinfo{person}{Vitaly Aksenov}, \bibinfo{person}{Dan Alistarh}, {and}
  \bibinfo{person}{Daniel~M. Roy}.} \bibinfo{year}{2021}\natexlab{}.
\newblock \showarticletitle{{NUQSGD:} Provably Communication-efficient
  Data-parallel {SGD} via Nonuniform Quantization}.
\newblock \bibinfo{journal}{\emph{CoRR}}  \bibinfo{volume}{abs/2104.13818}
  (\bibinfo{year}{2021}).
\newblock


\bibitem[\protect\citeauthoryear{Rasley, Rajbhandari, Ruwase, and He}{Rasley
  et~al\mbox{.}}{2020}]%
        {DeepSpeed}
\bibfield{author}{\bibinfo{person}{Jeff Rasley}, \bibinfo{person}{Samyam
  Rajbhandari}, \bibinfo{person}{Olatunji Ruwase}, {and}
  \bibinfo{person}{Yuxiong He}.} \bibinfo{year}{2020}\natexlab{}.
\newblock \showarticletitle{Deepspeed: System optimizations enable training
  deep learning models with over 100 billion parameters}. In
  \bibinfo{booktitle}{\emph{Proceedings of the 26th ACM SIGKDD International
  Conference on Knowledge Discovery \& Data Mining}}.
  \bibinfo{pages}{3505--3506}.
\newblock


\bibitem[\protect\citeauthoryear{Reddi, Kale, and Kumar}{Reddi
  et~al\mbox{.}}{2019}]%
        {reddi2019convergence}
\bibfield{author}{\bibinfo{person}{Sashank~J Reddi}, \bibinfo{person}{Satyen
  Kale}, {and} \bibinfo{person}{Sanjiv Kumar}.}
  \bibinfo{year}{2019}\natexlab{}.
\newblock \showarticletitle{On the convergence of adam and beyond}.
\newblock \bibinfo{journal}{\emph{arXiv preprint arXiv:1904.09237}}
  (\bibinfo{year}{2019}).
\newblock


\bibitem[\protect\citeauthoryear{Ren, He, Girshick, and Sun}{Ren
  et~al\mbox{.}}{2015}]%
  {Faster_R_CNN_Towards_Real_Time_Object_Detection_with_Region_Proposal_Networks}
\bibfield{author}{\bibinfo{person}{Shaoqing Ren}, \bibinfo{person}{Kaiming He},
  \bibinfo{person}{Ross Girshick}, {and} \bibinfo{person}{Jian Sun}.}
  \bibinfo{year}{2015}\natexlab{}.
\newblock \showarticletitle{Faster r-cnn: Towards real-time object detection
  with region proposal networks}.
\newblock \bibinfo{journal}{\emph{Advances in neural information processing
  systems}}  \bibinfo{volume}{28} (\bibinfo{year}{2015}),
  \bibinfo{pages}{91--99}.
\newblock


\bibitem[\protect\citeauthoryear{Ren and Goldfarb}{Ren and Goldfarb}{2021}]%
        {Kronecker_factored_Quasi_Newton_Methods_for_CNNs}
\bibfield{author}{\bibinfo{person}{Yi Ren} {and} \bibinfo{person}{Donald
  Goldfarb}.} \bibinfo{year}{2021}\natexlab{}.
\newblock \showarticletitle{Kronecker-factored Quasi-Newton Methods for
  Convolutional Neural Networks}.
\newblock \bibinfo{journal}{\emph{CoRR}}  \bibinfo{volume}{abs/2102.06737}
  (\bibinfo{year}{2021}).
\newblock


\bibitem[\protect\citeauthoryear{Ruder}{Ruder}{2016}]%
        {overview_of_gradient_descent_optimization_algorithms}
\bibfield{author}{\bibinfo{person}{Sebastian Ruder}.}
  \bibinfo{year}{2016}\natexlab{}.
\newblock \showarticletitle{An overview of gradient descent optimization
  algorithms}.
\newblock \bibinfo{journal}{\emph{CoRR}}  \bibinfo{volume}{abs/1609.04747}
  (\bibinfo{year}{2016}).
\newblock


\bibitem[\protect\citeauthoryear{Schneider, Balles, and Hennig}{Schneider
  et~al\mbox{.}}{2019}]%
        {DeepOBS}
\bibfield{author}{\bibinfo{person}{Frank Schneider}, \bibinfo{person}{Lukas
  Balles}, {and} \bibinfo{person}{Philipp Hennig}.}
  \bibinfo{year}{2019}\natexlab{}.
\newblock \showarticletitle{DeepOBS: {A} Deep Learning Optimizer Benchmark
  Suite}. In \bibinfo{booktitle}{\emph{7th International Conference on Learning
  Representations, {ICLR} 2019, New Orleans, LA, USA, May 6-9, 2019}}.
\newblock


\bibitem[\protect\citeauthoryear{Schraudolph, Yu, and G{\"u}nter}{Schraudolph
  et~al\mbox{.}}{2007}]%
        {L-BFGS}
\bibfield{author}{\bibinfo{person}{Nicol~N Schraudolph}, \bibinfo{person}{Jin
  Yu}, {and} \bibinfo{person}{Simon G{\"u}nter}.}
  \bibinfo{year}{2007}\natexlab{}.
\newblock \showarticletitle{A stochastic quasi-Newton method for online convex
  optimization}. In \bibinfo{booktitle}{\emph{Artificial intelligence and
  statistics}}. PMLR, \bibinfo{pages}{436--443}.
\newblock


\bibitem[\protect\citeauthoryear{Seide, Fu, Droppo, Li, and Yu}{Seide
  et~al\mbox{.}}{2014}]%
        {1_bit_SGD}
\bibfield{author}{\bibinfo{person}{Frank Seide}, \bibinfo{person}{Hao Fu},
  \bibinfo{person}{Jasha Droppo}, \bibinfo{person}{Gang Li}, {and}
  \bibinfo{person}{Dong Yu}.} \bibinfo{year}{2014}\natexlab{}.
\newblock \showarticletitle{1-bit stochastic gradient descent and its
  application to data-parallel distributed training of speech dnns}. In
  \bibinfo{booktitle}{\emph{Fifteenth Annual Conference of the International
  Speech Communication Association}}. Citeseer.
\newblock


\bibitem[\protect\citeauthoryear{Shalev{-}Shwartz and
  Ben{-}David}{Shalev{-}Shwartz and Ben{-}David}{2014}]%
        {Understanding_Machine_Learning_From_Theory_to_Algorithms}
\bibfield{author}{\bibinfo{person}{Shai Shalev{-}Shwartz} {and}
  \bibinfo{person}{Shai Ben{-}David}.} \bibinfo{year}{2014}\natexlab{}.
\newblock \bibinfo{booktitle}{\emph{Understanding Machine Learning - From
  Theory to Algorithms}}.
\newblock \bibinfo{publisher}{Cambridge University Press}.
\newblock
\showISBNx{978-1-10-705713-5}
\urldef\tempurl%
\url{http://www.cambridge.org/de/academic/subjects/computer-science/pattern-recognition-and-machine-learning/understanding-machine-learning-theory-algorithms}
\showURL{%
\tempurl}


\bibitem[\protect\citeauthoryear{Shallue, Lee, Antognini, Sohl{-}Dickstein,
  Frostig, and Dahl}{Shallue et~al\mbox{.}}{2019}]%
        {two_regime}
\bibfield{author}{\bibinfo{person}{Christopher~J. Shallue},
  \bibinfo{person}{Jaehoon Lee}, \bibinfo{person}{Joseph~M. Antognini},
  \bibinfo{person}{Jascha Sohl{-}Dickstein}, \bibinfo{person}{Roy Frostig},
  {and} \bibinfo{person}{George~E. Dahl}.} \bibinfo{year}{2019}\natexlab{}.
\newblock \showarticletitle{Measuring the Effects of Data Parallelism on Neural
  Network Training}.
\newblock \bibinfo{journal}{\emph{J. Mach. Learn. Res.}}  \bibinfo{volume}{20}
  (\bibinfo{year}{2019}), \bibinfo{pages}{112:1--112:49}.
\newblock


\bibitem[\protect\citeauthoryear{Shazeer and Stern}{Shazeer and Stern}{2018}]%
        {Adafactor}
\bibfield{author}{\bibinfo{person}{Noam Shazeer} {and}
  \bibinfo{person}{Mitchell Stern}.} \bibinfo{year}{2018}\natexlab{}.
\newblock \showarticletitle{Adafactor: Adaptive learning rates with sublinear
  memory cost}. In \bibinfo{booktitle}{\emph{International Conference on
  Machine Learning}}. PMLR, \bibinfo{pages}{4596--4604}.
\newblock


\bibitem[\protect\citeauthoryear{Simonyan and Zisserman}{Simonyan and
  Zisserman}{2014}]%
        {SimonyanZ14a}
\bibfield{author}{\bibinfo{person}{Karen Simonyan} {and}
  \bibinfo{person}{Andrew Zisserman}.} \bibinfo{year}{2014}\natexlab{}.
\newblock \showarticletitle{Very deep convolutional networks for large-scale
  image recognition}.
\newblock \bibinfo{journal}{\emph{arXiv preprint arXiv:1409.1556}}
  (\bibinfo{year}{2014}).
\newblock


\bibitem[\protect\citeauthoryear{Smith, Elsen, and De}{Smith
  et~al\mbox{.}}{2020}]%
        {smith2020generalization}
\bibfield{author}{\bibinfo{person}{Samuel Smith}, \bibinfo{person}{Erich
  Elsen}, {and} \bibinfo{person}{Soham De}.} \bibinfo{year}{2020}\natexlab{}.
\newblock \showarticletitle{On the Generalization Benefit of Noise in
  Stochastic Gradient Descent}. In \bibinfo{booktitle}{\emph{International
  Conference on Machine Learning}}. PMLR, \bibinfo{pages}{9058--9067}.
\newblock


\bibitem[\protect\citeauthoryear{Smith, Kindermans, Ying, and Le}{Smith
  et~al\mbox{.}}{2018}]%
        {increase_batch_size}
\bibfield{author}{\bibinfo{person}{Samuel~L Smith}, \bibinfo{person}{Pieter-Jan
  Kindermans}, \bibinfo{person}{Chris Ying}, {and} \bibinfo{person}{Quoc~V
  Le}.} \bibinfo{year}{2018}\natexlab{}.
\newblock \showarticletitle{Don't Decay the Learning Rate, Increase the Batch
  Size}. In \bibinfo{booktitle}{\emph{International Conference on Learning
  Representations}}.
\newblock


\bibitem[\protect\citeauthoryear{Smith and Le}{Smith and Le}{2018}]%
  {A_Bayesian_Perspective_on_Generalization_and_Stochastic_Gradient_Descent}
\bibfield{author}{\bibinfo{person}{Samuel~L Smith} {and}
  \bibinfo{person}{Quoc~V Le}.} \bibinfo{year}{2018}\natexlab{}.
\newblock \showarticletitle{A Bayesian Perspective on Generalization and
  Stochastic Gradient Descent}. In \bibinfo{booktitle}{\emph{International
  Conference on Learning Representations}}.
\newblock


\bibitem[\protect\citeauthoryear{Stich}{Stich}{2019}]%
        {Local_SGD_Converges_Fast_and_Communicates_Little}
\bibfield{author}{\bibinfo{person}{Sebastian~U. Stich}.}
  \bibinfo{year}{2019}\natexlab{}.
\newblock \showarticletitle{Local {SGD} Converges Fast and Communicates
  Little}. In \bibinfo{booktitle}{\emph{7th International Conference on
  Learning Representations, {ICLR} 2019, New Orleans, LA, USA, May 6-9, 2019}}.
\newblock


\bibitem[\protect\citeauthoryear{Stich, Cordonnier, and Jaggi}{Stich
  et~al\mbox{.}}{2018}]%
        {StichCJ18}
\bibfield{author}{\bibinfo{person}{Sebastian~U Stich},
  \bibinfo{person}{Jean-Baptiste Cordonnier}, {and} \bibinfo{person}{Martin
  Jaggi}.} \bibinfo{year}{2018}\natexlab{}.
\newblock \showarticletitle{Sparsified SGD with Memory}.
\newblock \bibinfo{journal}{\emph{Advances in Neural Information Processing
  Systems}}  \bibinfo{volume}{31} (\bibinfo{year}{2018}),
  \bibinfo{pages}{4447--4458}.
\newblock


\bibitem[\protect\citeauthoryear{Strom}{Strom}{2015}]%
        {Scalable_distributed_DNN_training_using_commodity_GPU_cloud_computing}
\bibfield{author}{\bibinfo{person}{Nikko Strom}.}
  \bibinfo{year}{2015}\natexlab{}.
\newblock \showarticletitle{Scalable distributed {DNN} training using commodity
  {GPU} cloud computing}. In \bibinfo{booktitle}{\emph{{INTERSPEECH} 2015, 16th
  Annual Conference of the International Speech Communication Association,
  Dresden, Germany, September 6-10, 2015}}. \bibinfo{pages}{1488--1492}.
\newblock


\bibitem[\protect\citeauthoryear{Sun}{Sun}{2019}]%
        {survey_Optimization_for_deep_learning_theory_and_algorithms}
\bibfield{author}{\bibinfo{person}{Ruoyu Sun}.}
  \bibinfo{year}{2019}\natexlab{}.
\newblock \showarticletitle{Optimization for deep learning: theory and
  algorithms}.
\newblock \bibinfo{journal}{\emph{CoRR}}  \bibinfo{volume}{abs/1912.08957}
  (\bibinfo{year}{2019}).
\newblock


\bibitem[\protect\citeauthoryear{Sutskever, Martens, Dahl, and
  Hinton}{Sutskever et~al\mbox{.}}{2013a}]%
        {On_the_importance_of_initialization_and_momentum_in_DL}
\bibfield{author}{\bibinfo{person}{Ilya Sutskever}, \bibinfo{person}{James
  Martens}, \bibinfo{person}{George~E. Dahl}, {and}
  \bibinfo{person}{Geoffrey~E. Hinton}.} \bibinfo{year}{2013}\natexlab{a}.
\newblock \showarticletitle{On the importance of initialization and momentum in
  deep learning}. In \bibinfo{booktitle}{\emph{Proceedings of the 30th
  International Conference on Machine Learning, {ICML} 2013, Atlanta, GA, USA,
  16-21 June 2013}} \emph{(\bibinfo{series}{{JMLR} Workshop and Conference
  Proceedings}, Vol.~\bibinfo{volume}{28})}. \bibinfo{pages}{1139--1147}.
\newblock


\bibitem[\protect\citeauthoryear{Sutskever, Martens, Dahl, and
  Hinton}{Sutskever et~al\mbox{.}}{2013b}]%
        {On_the_importance_of_initialization_and_momentum_in_deep_learning}
\bibfield{author}{\bibinfo{person}{Ilya Sutskever}, \bibinfo{person}{James
  Martens}, \bibinfo{person}{George~E. Dahl}, {and}
  \bibinfo{person}{Geoffrey~E. Hinton}.} \bibinfo{year}{2013}\natexlab{b}.
\newblock \showarticletitle{On the importance of initialization and momentum in
  deep learning}. In \bibinfo{booktitle}{\emph{Proceedings of the 30th
  International Conference on Machine Learning, {ICML} 2013, Atlanta, GA, USA,
  16-21 June 2013}} \emph{(\bibinfo{series}{{JMLR} Workshop and Conference
  Proceedings}, Vol.~\bibinfo{volume}{28})}. \bibinfo{pages}{1139--1147}.
\newblock


\bibitem[\protect\citeauthoryear{Szegedy, Liu, Jia, Sermanet, Reed, Anguelov,
  Erhan, Vanhoucke, and Rabinovich}{Szegedy et~al\mbox{.}}{2015}]%
        {SzegedyLJSRAEVR15}
\bibfield{author}{\bibinfo{person}{Christian Szegedy}, \bibinfo{person}{Wei
  Liu}, \bibinfo{person}{Yangqing Jia}, \bibinfo{person}{Pierre Sermanet},
  \bibinfo{person}{Scott~E. Reed}, \bibinfo{person}{Dragomir Anguelov},
  \bibinfo{person}{Dumitru Erhan}, \bibinfo{person}{Vincent Vanhoucke}, {and}
  \bibinfo{person}{Andrew Rabinovich}.} \bibinfo{year}{2015}\natexlab{}.
\newblock \showarticletitle{Going deeper with convolutions}. In
  \bibinfo{booktitle}{\emph{{IEEE} Conference on Computer Vision and Pattern
  Recognition, {CVPR} 2015, Boston, MA, USA, June 7-12, 2015}}.
  \bibinfo{publisher}{{IEEE} Computer Society}, \bibinfo{pages}{1--9}.
\newblock


\bibitem[\protect\citeauthoryear{Tang, Gan, Awan, Rajbhandari, Li, Lian, Liu,
  Zhang, and He}{Tang et~al\mbox{.}}{2021}]%
        {1_bit_Adam}
\bibfield{author}{\bibinfo{person}{Hanlin Tang}, \bibinfo{person}{Shaoduo Gan},
  \bibinfo{person}{Ammar~Ahmad Awan}, \bibinfo{person}{Samyam Rajbhandari},
  \bibinfo{person}{Conglong Li}, \bibinfo{person}{Xiangru Lian},
  \bibinfo{person}{Ji Liu}, \bibinfo{person}{Ce Zhang}, {and}
  \bibinfo{person}{Yuxiong He}.} \bibinfo{year}{2021}\natexlab{}.
\newblock \showarticletitle{1-bit Adam: Communication Efficient Large-Scale
  Training with Adam's Convergence Speed}. In \bibinfo{booktitle}{\emph{ICML}}.
\newblock


\bibitem[\protect\citeauthoryear{Tang, Shi, Chu, Wang, and Li}{Tang
  et~al\mbox{.}}{2020}]%
        {survey_Communication_Efficient_Distributed_DL}
\bibfield{author}{\bibinfo{person}{Zhenheng Tang}, \bibinfo{person}{Shaohuai
  Shi}, \bibinfo{person}{Xiaowen Chu}, \bibinfo{person}{Wei Wang}, {and}
  \bibinfo{person}{Bo Li}.} \bibinfo{year}{2020}\natexlab{}.
\newblock \showarticletitle{Communication-Efficient Distributed Deep Learning:
  {A} Comprehensive Survey}.
\newblock \bibinfo{journal}{\emph{CoRR}}  \bibinfo{volume}{abs/2003.06307}
  (\bibinfo{year}{2020}).
\newblock


\bibitem[\protect\citeauthoryear{Tieleman and Hinton}{Tieleman and
  Hinton}{2012}]%
        {RMSProp}
\bibfield{author}{\bibinfo{person}{T. Tieleman} {and} \bibinfo{person}{G.
  Hinton}.} \bibinfo{year}{2012}\natexlab{}.
\newblock \bibinfo{title}{{Lecture 6.5---RmsProp: Divide the gradient by a
  running average of its recent magnitude}}.
\newblock \bibinfo{howpublished}{COURSERA: Neural Networks for Machine
  Learning}.
\newblock


\bibitem[\protect\citeauthoryear{Vaswani, Shazeer, Parmar, Uszkoreit, Jones,
  Gomez, Kaiser, and Polosukhin}{Vaswani et~al\mbox{.}}{2017}]%
        {Attention_is_All_you_Need}
\bibfield{author}{\bibinfo{person}{Ashish Vaswani}, \bibinfo{person}{Noam
  Shazeer}, \bibinfo{person}{Niki Parmar}, \bibinfo{person}{Jakob Uszkoreit},
  \bibinfo{person}{Llion Jones}, \bibinfo{person}{Aidan~N. Gomez},
  \bibinfo{person}{Lukasz Kaiser}, {and} \bibinfo{person}{Illia Polosukhin}.}
  \bibinfo{year}{2017}\natexlab{}.
\newblock \showarticletitle{Attention is All you Need}. In
  \bibinfo{booktitle}{\emph{Advances in Neural Information Processing Systems
  30: Annual Conference on Neural Information Processing Systems 2017, December
  4-9, 2017, Long Beach, CA, {USA}}},
  \bibfield{editor}{\bibinfo{person}{Isabelle Guyon}, \bibinfo{person}{Ulrike
  von Luxburg}, \bibinfo{person}{Samy Bengio}, \bibinfo{person}{Hanna~M.
  Wallach}, \bibinfo{person}{Rob Fergus}, \bibinfo{person}{S.~V.~N.
  Vishwanathan}, {and} \bibinfo{person}{Roman Garnett}} (Eds.).
  \bibinfo{pages}{5998--6008}.
\newblock


\bibitem[\protect\citeauthoryear{Verma, Qassim, and Feinzimer}{Verma
  et~al\mbox{.}}{2017}]%
        {DBLP:conf/uemcom/VermaQF17}
\bibfield{author}{\bibinfo{person}{Abhishek Verma}, \bibinfo{person}{Hussam
  Qassim}, {and} \bibinfo{person}{David Feinzimer}.}
  \bibinfo{year}{2017}\natexlab{}.
\newblock \showarticletitle{Residual squeeze CNDS deep learning CNN model for
  very large scale places image recognition}.
\newblock \bibinfo{journal}{\emph{2017 IEEE 8th Annual Ubiquitous Computing,
  Electronics and Mobile Communication Conference (UEMCON)}}
  (\bibinfo{year}{2017}), \bibinfo{pages}{463--469}.
\newblock


\bibitem[\protect\citeauthoryear{Wang, Fu, He, Hao, and Wu}{Wang
  et~al\mbox{.}}{2020}]%
        {wang2020survey}
\bibfield{author}{\bibinfo{person}{Meng Wang}, \bibinfo{person}{Weijie Fu},
  \bibinfo{person}{Xiangnan He}, \bibinfo{person}{Shijie Hao}, {and}
  \bibinfo{person}{Xindong Wu}.} \bibinfo{year}{2020}\natexlab{}.
\newblock \showarticletitle{A survey on large-scale machine learning}.
\newblock \bibinfo{journal}{\emph{IEEE Transactions on Knowledge and Data
  Engineering}} (\bibinfo{year}{2020}).
\newblock


\bibitem[\protect\citeauthoryear{Wangni, Wang, Liu, and Zhang}{Wangni
  et~al\mbox{.}}{2018}]%
        {WangniWLZ18}
\bibfield{author}{\bibinfo{person}{Jianqiao Wangni}, \bibinfo{person}{Jialei
  Wang}, \bibinfo{person}{Ji Liu}, {and} \bibinfo{person}{Tong Zhang}.}
  \bibinfo{year}{2018}\natexlab{}.
\newblock \showarticletitle{Gradient Sparsification for Communication-Efficient
  Distributed Optimization}. In \bibinfo{booktitle}{\emph{Advances in Neural
  Information Processing Systems 31: Annual Conference on Neural Information
  Processing Systems 2018, NeurIPS 2018, December 3-8, 2018, Montr{\'{e}}al,
  Canada}}, \bibfield{editor}{\bibinfo{person}{Samy Bengio},
  \bibinfo{person}{Hanna~M. Wallach}, \bibinfo{person}{Hugo Larochelle},
  \bibinfo{person}{Kristen Grauman}, \bibinfo{person}{Nicol{\`{o}}
  Cesa{-}Bianchi}, {and} \bibinfo{person}{Roman Garnett}} (Eds.).
  \bibinfo{pages}{1306--1316}.
\newblock


\bibitem[\protect\citeauthoryear{Wen, Xu, Yan, Wu, Wang, Chen, and Li}{Wen
  et~al\mbox{.}}{2017}]%
        {TernGrad}
\bibfield{author}{\bibinfo{person}{Wei Wen}, \bibinfo{person}{Cong Xu},
  \bibinfo{person}{Feng Yan}, \bibinfo{person}{Chunpeng Wu},
  \bibinfo{person}{Yandan Wang}, \bibinfo{person}{Yiran Chen}, {and}
  \bibinfo{person}{Hai Li}.} \bibinfo{year}{2017}\natexlab{}.
\newblock \showarticletitle{TernGrad: Ternary Gradients to Reduce Communication
  in Distributed Deep Learning}. In \bibinfo{booktitle}{\emph{Advances in
  Neural Information Processing Systems 30: Annual Conference on Neural
  Information Processing Systems 2017, December 4-9, 2017, Long Beach, CA,
  {USA}}}. \bibinfo{pages}{1509--1519}.
\newblock


\bibitem[\protect\citeauthoryear{Wilson, Roelofs, Stern, Srebro, and
  Recht}{Wilson et~al\mbox{.}}{2017}]%
        {The_Marginal_Value_of_Adaptive_Gradient_Methods_in_Machine_Learning}
\bibfield{author}{\bibinfo{person}{Ashia~C. Wilson}, \bibinfo{person}{Rebecca
  Roelofs}, \bibinfo{person}{Mitchell Stern}, \bibinfo{person}{Nati Srebro},
  {and} \bibinfo{person}{Benjamin Recht}.} \bibinfo{year}{2017}\natexlab{}.
\newblock \showarticletitle{The Marginal Value of Adaptive Gradient Methods in
  Machine Learning}. In \bibinfo{booktitle}{\emph{Advances in Neural
  Information Processing Systems 30: Annual Conference on Neural Information
  Processing Systems 2017, December 4-9, 2017, Long Beach, CA, {USA}}}.
  \bibinfo{pages}{4148--4158}.
\newblock


\bibitem[\protect\citeauthoryear{Wu, Schuster, Chen, Le, Norouzi, Macherey,
  Krikun, Cao, Gao, Macherey, Klingner, Shah, Johnson, Liu, Kaiser, Gouws,
  Kato, Kudo, Kazawa, Stevens, Kurian, Patil, Wang, Young, Smith, Riesa,
  Rudnick, Vinyals, Corrado, Hughes, and Dean}{Wu et~al\mbox{.}}{2016}]%
        {machine_translation}
\bibfield{author}{\bibinfo{person}{Yonghui Wu}, \bibinfo{person}{Mike
  Schuster}, \bibinfo{person}{Zhifeng Chen}, \bibinfo{person}{Quoc~V. Le},
  \bibinfo{person}{Mohammad Norouzi}, \bibinfo{person}{Wolfgang Macherey},
  \bibinfo{person}{Maxim Krikun}, \bibinfo{person}{Yuan Cao},
  \bibinfo{person}{Qin Gao}, \bibinfo{person}{Klaus Macherey},
  \bibinfo{person}{Jeff Klingner}, \bibinfo{person}{Apurva Shah},
  \bibinfo{person}{Melvin Johnson}, \bibinfo{person}{Xiaobing Liu},
  \bibinfo{person}{Lukasz Kaiser}, \bibinfo{person}{Stephan Gouws},
  \bibinfo{person}{Yoshikiyo Kato}, \bibinfo{person}{Taku Kudo},
  \bibinfo{person}{Hideto Kazawa}, \bibinfo{person}{Keith Stevens},
  \bibinfo{person}{George Kurian}, \bibinfo{person}{Nishant Patil},
  \bibinfo{person}{Wei Wang}, \bibinfo{person}{Cliff Young},
  \bibinfo{person}{Jason Smith}, \bibinfo{person}{Jason Riesa},
  \bibinfo{person}{Alex Rudnick}, \bibinfo{person}{Oriol Vinyals},
  \bibinfo{person}{Greg Corrado}, \bibinfo{person}{Macduff Hughes}, {and}
  \bibinfo{person}{Jeffrey Dean}.} \bibinfo{year}{2016}\natexlab{}.
\newblock \showarticletitle{Google's Neural Machine Translation System:
  Bridging the Gap between Human and Machine Translation}.
\newblock \bibinfo{journal}{\emph{CoRR}}  \bibinfo{volume}{abs/1609.08144}
  (\bibinfo{year}{2016}).
\newblock


\bibitem[\protect\citeauthoryear{Xiong, Yang, He, Zheng, Zheng, Xing, Zhang,
  Lan, Wang, and Liu}{Xiong et~al\mbox{.}}{2020}]%
        {On_Layer_Normalization_in_the_Transformer_Architecture}
\bibfield{author}{\bibinfo{person}{Ruibin Xiong}, \bibinfo{person}{Yunchang
  Yang}, \bibinfo{person}{Di He}, \bibinfo{person}{Kai Zheng},
  \bibinfo{person}{Shuxin Zheng}, \bibinfo{person}{Chen Xing},
  \bibinfo{person}{Huishuai Zhang}, \bibinfo{person}{Yanyan Lan},
  \bibinfo{person}{Liwei Wang}, {and} \bibinfo{person}{Tie{-}Yan Liu}.}
  \bibinfo{year}{2020}\natexlab{}.
\newblock \showarticletitle{On Layer Normalization in the Transformer
  Architecture}. In \bibinfo{booktitle}{\emph{Proceedings of the 37th
  International Conference on Machine Learning, {ICML} 2020, 13-18 July 2020,
  Virtual Event}} \emph{(\bibinfo{series}{Proceedings of Machine Learning
  Research}, Vol.~\bibinfo{volume}{119})}. \bibinfo{pages}{10524--10533}.
\newblock


\bibitem[\protect\citeauthoryear{Xu, Roosta, and Mahoney}{Xu
  et~al\mbox{.}}{2020}]%
  {Second_order_Optimization_for_Non_convex_Machine_Learning_an_Empirical_Study}
\bibfield{author}{\bibinfo{person}{Peng Xu}, \bibinfo{person}{Fred Roosta},
  {and} \bibinfo{person}{Michael~W. Mahoney}.} \bibinfo{year}{2020}\natexlab{}.
\newblock \showarticletitle{Second-order Optimization for Non-convex Machine
  Learning: an Empirical Study}. In \bibinfo{booktitle}{\emph{Proceedings of
  the 2020 {SIAM} International Conference on Data Mining, {SDM} 2020,
  Cincinnati, Ohio, USA, May 7-9, 2020}}. \bibinfo{pages}{199--207}.
\newblock


\bibitem[\protect\citeauthoryear{Xue, Shi, Lou, Liu, and You}{Xue
  et~al\mbox{.}}{2021}]%
        {xue2021go}
\bibfield{author}{\bibinfo{person}{Fuzhao Xue}, \bibinfo{person}{Ziji Shi},
  \bibinfo{person}{Yuxuan Lou}, \bibinfo{person}{Yong Liu}, {and}
  \bibinfo{person}{Yang You}.} \bibinfo{year}{2021}\natexlab{}.
\newblock \showarticletitle{Go Wider Instead of Deeper}.
\newblock \bibinfo{journal}{\emph{arXiv preprint arXiv:2107.11817}}
  (\bibinfo{year}{2021}).
\newblock


\bibitem[\protect\citeauthoryear{Yamazaki, Kasagi, Tabuchi, Honda, Miwa,
  Fukumoto, Tabaru, Ike, and Nakashima}{Yamazaki et~al\mbox{.}}{2019}]%
        {DBLP:journals/corr/abs-1903-12650}
\bibfield{author}{\bibinfo{person}{Masafumi Yamazaki}, \bibinfo{person}{Akihiko
  Kasagi}, \bibinfo{person}{Akihiro Tabuchi}, \bibinfo{person}{Takumi Honda},
  \bibinfo{person}{Masahiro Miwa}, \bibinfo{person}{Naoto Fukumoto},
  \bibinfo{person}{Tsuguchika Tabaru}, \bibinfo{person}{Atsushi Ike}, {and}
  \bibinfo{person}{Kohta Nakashima}.} \bibinfo{year}{2019}\natexlab{}.
\newblock \showarticletitle{Yet Another Accelerated SGD: ResNet-50 Training on
  ImageNet in 74.7 seconds}.
\newblock \bibinfo{journal}{\emph{ArXiv}}  \bibinfo{volume}{abs/1903.12650}
  (\bibinfo{year}{2019}).
\newblock


\bibitem[\protect\citeauthoryear{Ying, Kumar, Chen, Wang, and Cheng}{Ying
  et~al\mbox{.}}{2018}]%
        {Image_Classification_at_Supercomputer_Scale}
\bibfield{author}{\bibinfo{person}{Chris Ying}, \bibinfo{person}{Sameer Kumar},
  \bibinfo{person}{Dehao Chen}, \bibinfo{person}{Tao Wang}, {and}
  \bibinfo{person}{Youlong Cheng}.} \bibinfo{year}{2018}\natexlab{}.
\newblock \showarticletitle{Image Classification at Supercomputer Scale}.
\newblock \bibinfo{journal}{\emph{CoRR}}  \bibinfo{volume}{abs/1811.06992}
  (\bibinfo{year}{2018}).
\newblock


\bibitem[\protect\citeauthoryear{You, Gitman, and Ginsburg}{You
  et~al\mbox{.}}{2017}]%
        {LARS}
\bibfield{author}{\bibinfo{person}{Yang You}, \bibinfo{person}{Igor Gitman},
  {and} \bibinfo{person}{Boris Ginsburg}.} \bibinfo{year}{2017}\natexlab{}.
\newblock \bibinfo{title}{Large Batch Training of Convolutional Networks}.
\newblock
\newblock
\showeprint[arxiv]{1708.03888}~[cs.CV]


\bibitem[\protect\citeauthoryear{You, Li, Reddi, Hseu, Kumar, Bhojanapalli,
  Song, Demmel, Keutzer, and Hsieh}{You et~al\mbox{.}}{2020}]%
        {LAMB}
\bibfield{author}{\bibinfo{person}{Yang You}, \bibinfo{person}{Jing Li},
  \bibinfo{person}{Sashank Reddi}, \bibinfo{person}{Jonathan Hseu},
  \bibinfo{person}{Sanjiv Kumar}, \bibinfo{person}{Srinadh Bhojanapalli},
  \bibinfo{person}{Xiaodan Song}, \bibinfo{person}{James Demmel},
  \bibinfo{person}{Kurt Keutzer}, {and} \bibinfo{person}{Cho-Jui Hsieh}.}
  \bibinfo{year}{2020}\natexlab{}.
\newblock \bibinfo{title}{Large Batch Optimization for Deep Learning: Training
  BERT in 76 minutes}.
\newblock
\newblock
\showeprint[arxiv]{1904.00962}~[cs.LG]


\bibitem[\protect\citeauthoryear{Zeiler}{Zeiler}{2012}]%
        {Adadelta}
\bibfield{author}{\bibinfo{person}{Matthew~D. Zeiler}.}
  \bibinfo{year}{2012}\natexlab{}.
\newblock \showarticletitle{ADADELTA: An Adaptive Learning Rate Method}.
\newblock \bibinfo{journal}{\emph{ArXiv}}  \bibinfo{volume}{abs/1212.5701}
  (\bibinfo{year}{2012}).
\newblock


\bibitem[\protect\citeauthoryear{Zhang, Li, Nado, Martens, Sachdeva, Dahl,
  Shallue, and Grosse}{Zhang et~al\mbox{.}}{2019}]%
        {zhang2019algorithmic}
\bibfield{author}{\bibinfo{person}{Guodong Zhang}, \bibinfo{person}{Lala Li},
  \bibinfo{person}{Zachary Nado}, \bibinfo{person}{James Martens},
  \bibinfo{person}{Sushant Sachdeva}, \bibinfo{person}{George Dahl},
  \bibinfo{person}{Chris Shallue}, {and} \bibinfo{person}{Roger~B Grosse}.}
  \bibinfo{year}{2019}\natexlab{}.
\newblock \showarticletitle{Which algorithmic choices matter at which batch
  sizes? insights from a noisy quadratic model}.
\newblock \bibinfo{journal}{\emph{Advances in neural information processing
  systems}}  \bibinfo{volume}{32} (\bibinfo{year}{2019}),
  \bibinfo{pages}{8196--8207}.
\newblock


\bibitem[\protect\citeauthoryear{Zhang, Sa, Mitliagkas, and R{\'{e}}}{Zhang
  et~al\mbox{.}}{2016}]%
        {Parallel_SGD_When_does_averaging_help}
\bibfield{author}{\bibinfo{person}{Jian Zhang}, \bibinfo{person}{Christopher~De
  Sa}, \bibinfo{person}{Ioannis Mitliagkas}, {and} \bibinfo{person}{Christopher
  R{\'{e}}}.} \bibinfo{year}{2016}\natexlab{}.
\newblock \showarticletitle{Parallel {SGD:} When does averaging help?}
\newblock \bibinfo{journal}{\emph{CoRR}}  \bibinfo{volume}{abs/1606.07365}
  (\bibinfo{year}{2016}).
\newblock


\bibitem[\protect\citeauthoryear{Zhou and Cong}{Zhou and Cong}{2018}]%
        {local_sgd_non_convex_convergence}
\bibfield{author}{\bibinfo{person}{Fan Zhou} {and} \bibinfo{person}{Guojing
  Cong}.} \bibinfo{year}{2018}\natexlab{}.
\newblock \showarticletitle{On the Convergence Properties of a K-step Averaging
  Stochastic Gradient Descent Algorithm for Nonconvex Optimization}. In
  \bibinfo{booktitle}{\emph{Proceedings of the Twenty-Seventh International
  Joint Conference on Artificial Intelligence, {IJCAI} 2018, July 13-19, 2018,
  Stockholm, Sweden}}. \bibinfo{pages}{3219--3227}.
\newblock


\bibitem[\protect\citeauthoryear{Zinkevich, Weimer, Smola, and Li}{Zinkevich
  et~al\mbox{.}}{2010}]%
        {Parallelized_SGD}
\bibfield{author}{\bibinfo{person}{Martin Zinkevich}, \bibinfo{person}{Markus
  Weimer}, \bibinfo{person}{Alexander~J. Smola}, {and} \bibinfo{person}{Lihong
  Li}.} \bibinfo{year}{2010}\natexlab{}.
\newblock \showarticletitle{Parallelized Stochastic Gradient Descent}. In
  \bibinfo{booktitle}{\emph{Advances in Neural Information Processing Systems
  23: 24th Annual Conference on Neural Information Processing Systems 2010.
  Proceedings of a meeting held 6-9 December 2010, Vancouver, British Columbia,
  Canada}}, \bibfield{editor}{\bibinfo{person}{John~D. Lafferty},
  \bibinfo{person}{Christopher K.~I. Williams}, \bibinfo{person}{John
  Shawe{-}Taylor}, \bibinfo{person}{Richard~S. Zemel}, {and}
  \bibinfo{person}{Aron Culotta}} (Eds.). \bibinfo{pages}{2595--2603}.
\newblock


\end{thebibliography}

% %%
% %% If your work has an appendix, this is the place to put it.
% \appendix









\end{document}